\documentclass[lettersize,journal]{IEEEtran}
\usepackage{graphicx}
\usepackage{amsmath}
\usepackage{times}
\usepackage{amssymb}
\usepackage{booktabs}
\usepackage{multirow}
\usepackage{multicol}
\usepackage{algorithm}
\usepackage{algorithmicx}
\usepackage{verbatim}
\usepackage{color}
\usepackage{cite}
\usepackage{subcaption}
\usepackage{caption}
\usepackage{tabularx}
\usepackage{xcolor}
\usepackage[numbers,sort&compress]{natbib}
\begin{document}

\title{Deep Learning for \\ 3D Point Cloud Enhancement: A Survey}
\author{Siwen~Quan, Junhao~Yu, Ziming~Nie, Muze~Wang, Sijia~Feng, Pei~An, and Jiaqi~Yang
\thanks{
This work is supported in part by the National Natural Science Foundation of China (No. 62372377 and 62006025).

Siwen Quan is with the School of Electronics and Control Engineering, Chang'an University, Xi'an 710064,	China. (Email: siwenquan@chd.edu.cn).

Junhao Yu and Sijia Feng are with the School of Software Engineering, Northwestern Polytechnical University, Xi'an 710064,	China. (Email: junhaoyu2003@163.com, fsj@mail.nwpu.edu.cn).

Ziming Nie, Muze Wang and Jiaqi Yang are with the School of Computer Science, Northwestern Polytechnical University, Xi'an 710064, China. (Email: \{nzm, iamwmz\}@mail.nwpu.edu.cn, jqyang@nwpu.edu.cn).

Pei An is with the School of Electronic
Information and Communications, Huazhong University of Science and Technology, and also with the Wuhan National Laboratory for Optoelectronics, Wuhan 430074, China. (Email: anpei96@hust.edu.cn). (\textit{Corresponding authors: Pei An and Jiaqi Yang})
}
}

\markboth{IEEE/CAA JOURNAL OF AUTOMATICA SINICA,~Vol.~X, No.~X, X~X}
{Shell \MakeLowercase{\textit{et al.}}: Bare Demo of IEEEtran.cls for IEEE Journals}

\maketitle

\begin{abstract}
Point cloud data now are popular data representations in a number of three-dimensional (3D) vision research realms. However, due to the limited performance of sensors and sensing noise, the raw data usually suffer from sparsity, noise, and incompleteness. This poses great challenges to down-stream point cloud processing tasks. In recent years, deep-learning-based point cloud enhancement methods, which aim to achieve dense, clean, and complete point clouds from low-quality raw point clouds using deep neural networks, are gaining tremendous research attention. This paper, for the first time to our knowledge, presents a comprehensive survey for deep-learning-based point cloud enhancement methods. It covers three main perspectives for  point cloud enhancement, i.e., (1) denoising to achieve clean data; (2) completion to recover unseen data; (3) upsampling to obtain dense data. Our survey presents a new taxonomy for recent state-of-the-art methods and systematic experimental results on standard benchmarks. In addition, we share our insightful observations, thoughts, and inspiring future research directions for point cloud enhancement with deep learning.
\end{abstract}


\section{Introduction}

\IEEEPARstart{T}{hree}-dimensional (3D) point clouds are essential data representations in 3D vision \cite{jas1}, such as robotics, autonomous driving, augmented reality, and metaverse \cite{jas2}. Due to the limited performance of  existing ranging sensors as well as sensing noise, the raw point cloud is often noisy, incomplete, and sparse \cite{jas3, jas4}. As such, point cloud enhancement is critical to generate high-quality 3D point clouds for the down-stream point-cloud-based tasks. Hence, this paper is dedicated to the point cloud enhancement process, which typically involves point cloud denosing, completion, and upsampling.


\begin{figure}[t]
	\centering
	\includegraphics[width=1.0\linewidth]{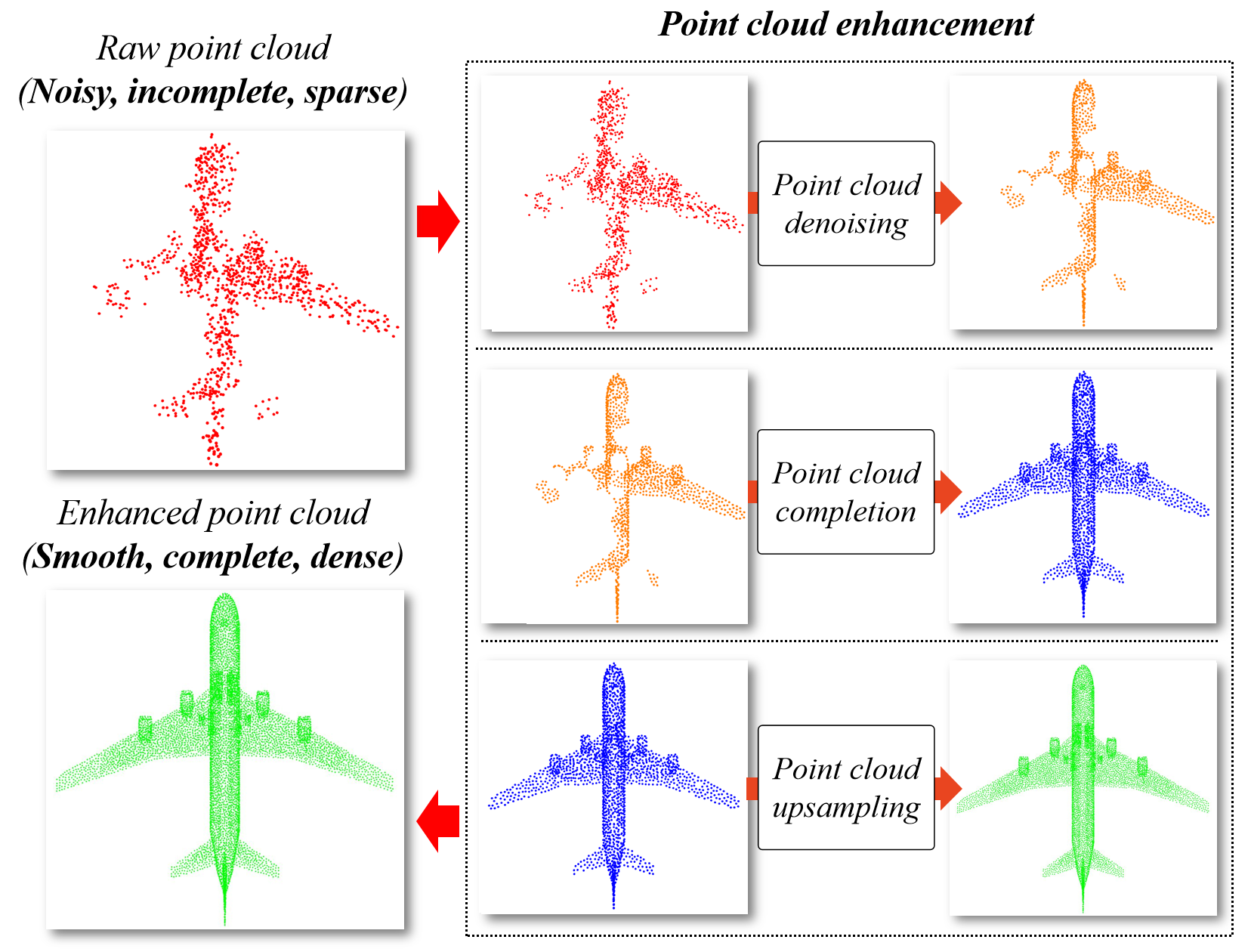}	
      \caption{Illustration of 3D point cloud enhancement. It aims to generate smooth, complete, and dense point clouds from low-quality raw point clouds.  In point cloud enhancement, point cloud denoising, completion, and upsampling are three main tasks. Specifically, denosing aims to eliminate or rectify noisy points; completion is to recover the missing part in point clouds; upsampling is to enhance the resolution of the point cloud.}
	\label{fig:into}
\end{figure}

The goal of 3D point cloud enhancement is to resample points upon given input point cloud to generate high-quality points, which  are expected to be  clean, complete, and dense. In recent years, deep-learning-based methods for point cloud enhancement spring up rapidly, thank to the advancements of several large-scale 3D vision datasets, such as ShapeNet \cite{chang2015shapenet}, KITTI \cite{geiger2012we} and ScanNet \cite{dai2017scannet}. They enable the neural networks to learn discriminative feature representations from raw point clouds and predict their high-quality data variations \cite{tmm_ref_3}. Owing to the strong representation ability of deep neural networks, the performance of deep-learning-based point cloud enhancement methods significantly outperforms traditional ones based on optimization and interpolation \cite{an_add_6}. Thus, performing point cloud enhancement with deep neural networks becomes a dominating solution. 
Unfortunately, existing 3D point cloud enhancement methods are only discussed partially in survey papers addressing other topics such as 3D point cloud learning \cite{an_add_7}, 3D object detection \cite{an_add_8}, and 3D scene completion \cite{an_add_9}.

\begin{figure*}[t]
    \centering
    \includegraphics[width=\textwidth]{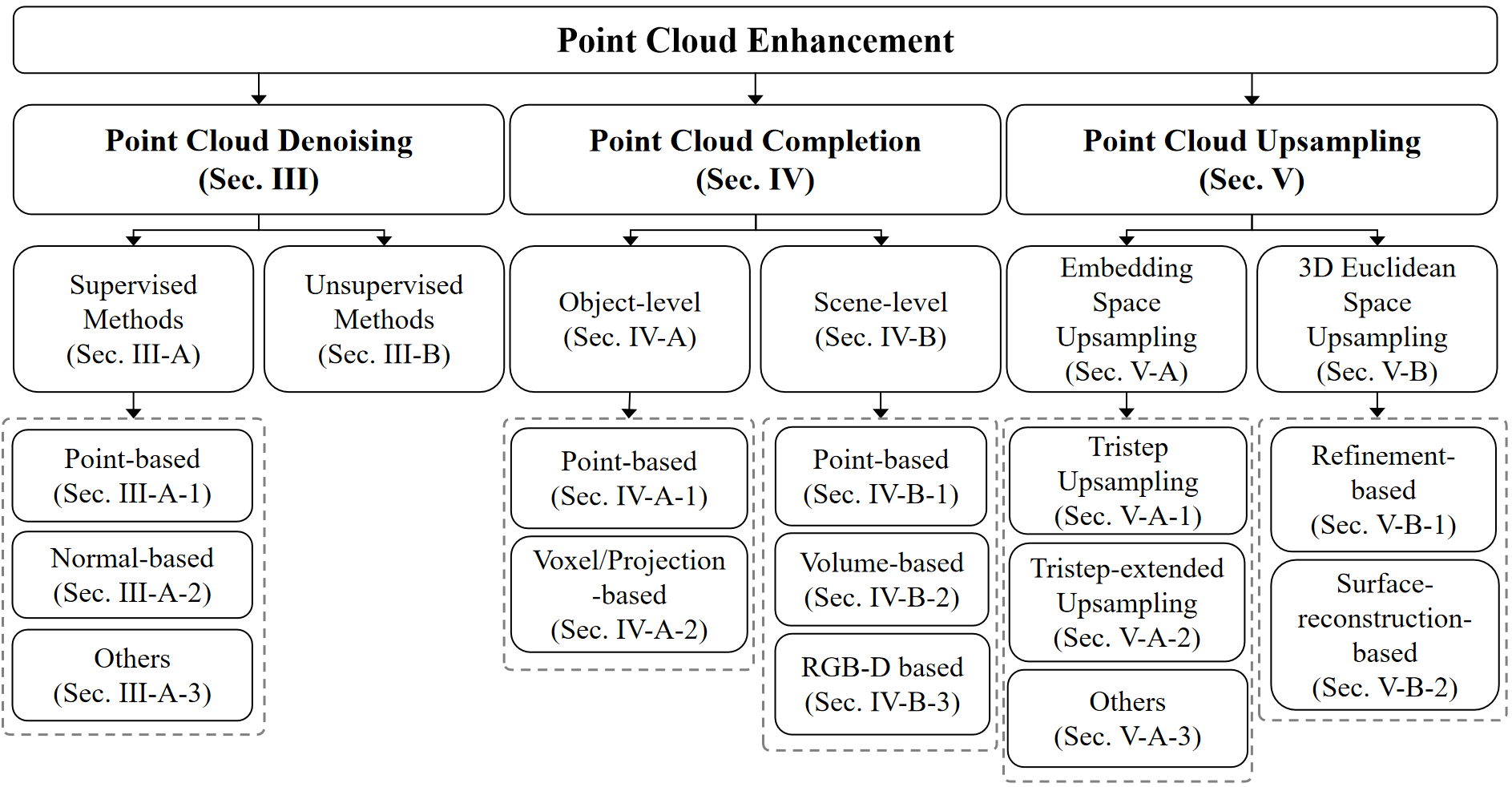}
    \caption{A taxonomy of 3D point cloud enhancement.}
    \label{resampling_taxonomy}
\end{figure*}


Motivated by this, we present a comprehensive survey on deep-learning-based 3D point cloud enhancement to fill the gap. We concentrate on three critical enhancement problems (Fig. \ref{fig:into}), including (i) denoising, (ii) completion, and (iii) upsampling. These tasks can enhance the smoothness, completeness, resolution of the raw point cloud, respectively. Our taxonomy of existing deep-learning-based 3D point cloud enhancement methods is shown in Fig.~\ref{resampling_taxonomy}. 

Compared with the published surveys related to point cloud enhancement \cite{review1,review2,review3,review4}, our survey has the following traits. First, our survey focuses on systematically analyzing point cloud enhancement in a global viewpoint, instead of illustrating methods in a sub-field of point cloud enhancement. For example, Zhou et al. \cite{review2} provided a broad overview of denoising methods, categorizing them into filter-based, optimization-based, and deep learning-based approaches. However, their review includes an extensive examination of classical methods and is not exclusively centered on deep learning. By contrast, our work is dedicated solely to deep learning techniques, offering a more detailed and specific discussion of these methods. Second, our literature review is more comprehensive. For instance, Kwon et al. \cite{review4} categorized point cloud upsampling techniques into surface consolidation and edge consolidation based on the target area of upsampling. Despite their valuable contributions, their review lacks comprehensiveness and covers a limited scope of the literature. Third, we propose a new taxonomy for deep learned point cloud enhancement methods. Fei et al. \cite{review3} classified existing architectures into point-based, view-based, convolution-based, graph-based, generative model-based, and transformer-based methods based on network structures. By contrast, our taxonomies for point cloud denoising, completion and upsampling methods are from different perspectives. Fourth, we present a more comprehensive experimental comparison by considering representative and the most recent methods in the realm. Lastly, we offer new insights to the community, contributing to a deeper understanding and fostering further advancements in this field.\par

To summarize, the major contributions of this work are as follows:
\begin{itemize}
\item {\textbf {Comprehensive review and systematic taxonomy.}} To the best of our knowledge, this is the first survey paper to comprehensively review recent deep learning methods for point cloud enhancement, offering a systematic taxonomy and a comprehensive literature coverage.
\item {\textbf {Benchmark overview and performance comparison.}} The experimental datasets and evaluation metrics for point cloud enhancement are systematically summarized. We also present thorough comparative results of representative state-of-the-art methods on standard benchmarks.
\item {\textbf {Outlook on future directions.}} The traits, merits, and demerits of existing methods have been highlighted. We present insightful discussions on current challenges and several future research directions to inspire further advancement in this field. 
\end{itemize}



The remainder of this paper is organized as follows. Sec. II details the background of 3D point cloud enhancement. Then, we survey the specific tasks in point cloud enhancement, i.e., point cloud denoising in Sec. III, point cloud completion in Sec. IV, and point cloud upsampling in Sec. V. Sec. VI presents current challenges and opportunities.  Finally, Sec. VII concludes the survey.

\section{Background}
We introduce the basic concepts, popular datasets, and standard evaluation metrics in the following.

\subsection{Basic Concepts} 
\textbf{3D Point Cloud}. A 3D point cloud $P$ is a set of 3D point  $ \{p_i | i = 1,2,\ldots,N\}, p_i \in \mathbf{R}^3$. It is a discrete sampling of the 3D continuous surface of objects or scenes, which is unorganized and permutation-insensitive. In this survey, we primary concentrate on point clouds without RGB and intensity attributes, so $p=(x,y,z)$ only contains the coordinate information.

\textbf{3D Point Cloud Enhancement}. Let $P_{raw}$ and $P_{resample}$ be the raw point cloud and resampled point cloud, respectively. The objective of 3D point cloud enhancement is to determine a proper transformation ${\cal T}: P_{raw} \rightarrow P_{resample}$. In particular,   $P_{resample}$ is expected to possess higher quality than $P_{raw}$, such as smoother, more complete, and denser. As such, point cloud enhancement is typically performed by independently or sequentially conducting point cloud denoising, completion, and upsampling.
\subsection{Datasets}
\begin{table*}[t]
  \centering
  \caption{Summary of standard benchmarks for 3D point cloud enhancement.}
  \vspace{-0.2cm}
  \resizebox{\textwidth}{!}{
    \begin{tabular}{ccccc}
      \toprule[1.0pt]
    Dataset & Year  & Samples & Type  & Task \\
    \midrule
    Kinect v1 \cite{wang2016mesh} & 2010  & 71 objects & real-world object & denoising \\
    NYUv2 \cite{silberman2012indoor} & 2012  & 1,449 RGB-D images & real-world indoor scene & completion \\
    KITTI~\cite{geiger2012we} & 2012  & 15K frames & real-world outdoor scene &  completion \& upsampling \\
    Kinect v2 \cite{wang2016mesh} & 2013  & 72 objects & real-world object & denoising \\
    Paris-rue-Madame \cite{serna2014paris} & 2014  & 642 objects & real-world urban object & denoising \\
    ShapeNet \cite{chang2015shapenet} & 2015  & 51,190 objects & synthetic object & completion \& upsampling \\
    NYUCAD \cite{firman2016structured} & 2016  & 90 desktop scenes & synthetic indoor scene & completion \\
    ScanNet \cite{dai2017scannet} & 2017  & 1,513 scans & real-world indoor scene & completion \\
    MatterPort3D \cite{chang2017matterport3d} & 2017  & 10,800 panoramic views & real-world indoor scene & completion \\
    SUNCG \cite{song2017semantic} & 2017  & 45622 scenes & synthetic indoor scene & completion \\
    PUNet \cite{yu2018pu} & 2018  & 20 meshes & synthetic object & denoising \\
    PCPNet \cite{guerrero2018pcpnet} & 2018  & 30 shapes & synthetic object & denoising \\
    Completion3D \cite{tchapmi2019topnet} & 2019  & 30,974 objects & synthetic object & completion \\
    SemanticKITTI \cite{behley2019semantickitti} & 2019  & 43552 scenes & real-world outdoor scene & completion \\
    PU-GAN~\cite{li2019pu} & 2019  & 1,024 objects & synthetic object & upsampling \\
    ScanObjectNN~\cite{uy2019revisiting} & 2019  & 2,902 objects & real-world object & upsampling \\
    Pointfilter \cite{zhang2020pointfilter} & 2020  & 11 CAD and 11 non-CAD models & synthetic object & denoising \\
    DMR \cite{luo2020differentiable} & 2020  & 60 meshes & synthetic object & denoising \\
    MVP \cite{pan2021vrcnet} & 2021  & over 100,000 objects & synthetic object & completion \\
    PU1K~\cite{qian2021pu} & 2021  & 1,147 objects & synthetic object & upsampling \\
    ScanSalon \cite{wu2023scoda} & 2023  & 7,644 scans & synthetic object & completion \\
    \bottomrule[1.0pt]
    \end{tabular}
  \label{resamplingtable}}
  \vspace{-0.3cm}
\end{table*}
There are various kinds of datasets utilized to train and evaluate the performance of point cloud enhancement methods. We summarize standard datasets in Table \ref{resamplingtable}.\par
\textbf{Point Cloud Denoising.} It typically experiments with two primary categories of datasets: synthetic and real-world. PUNet test dataset described by Yu et al. \cite{yu2018pu} is a representative synthetic dataset. It consists of point clouds sampled from 20 mesh shapes at two different resolutions, i.e., a sparse resolution with 10,000 points per shape and a dense resolution containing 50,000 points per shape. Each resolution is further subjected to three levels of Gaussian noise, equivalent to 1\%, 2\%, and 3\% of the bounding sphere's radius. Thus, it allows for the examination of noise impacts across varying densities. Real-world datasets, exemplified by the Kinect dataset from Wang et al. \cite{wang2016mesh}, encompass scanned data obtained using Kinect versions 1 and 2. These datasets are inherently noisy and present more complex challenges for denoising methods compared to synthetic data. The Kinect datasets also include reconstructed clean data, which serve as a reference for quantitative evaluations and benchmarks.

\textbf{Point Cloud Completion.} It leverages various datasets categorized into object and scene types. Object datasets primarily include ShapeNet \cite{chang2015shapenet}, Completion3D \cite{tchapmi2019topnet}, MVP \cite{pan2021vrcnet}, and ScanSalon \cite{wu2023scoda}. Scene-level datasets include NYUv2 \cite{silberman2012indoor}, KITTI \cite{geiger2013vision}, NYUCAD \cite{firman2016structured}, ScanNet \cite{dai2017scannet}, MatterPort3D \cite{chang2017matterport3d}, SUNCG \cite{song2017semantic}, and SemanticKITTI \cite{behley2019semantickitti}. Notably, NYUv2, KITTI, ScanNet, MatterPort3D, and SemanticKITTI are real-world datasets, while the others are synthetic ones. Compared to synthetic datasets, real-world datasets pose greater challenges and hold greater practical application value. \par

\textbf{Point Cloud Upsampling.} It typically utilizes two distinct categories of data: synthetic and real-scanned point clouds. Synthetic datasets are derived from CAD models. They are characterized by their uniformity and minimal noise. Prominent datasets in this category include PU-GAN \cite{li2019pu}, PU1K \cite{qian2021pu}, and ShapeNet \cite{chang2015shapenet}, which are extensively used to train and evaluate various upsampling methods. By contrast, real-scanned datasets such as KITTI \cite{geiger2012we} and ScanObjectNN \cite{uy2019revisiting} present greater challenges due to their inherent noise, non-uniformity, and sparsity. These attributes make real-scanned datasets particularly suitable for assessing the generalization ability of upsampling methods. \par
To ensure a fair comparison and comprehensive coverage of various methods, we selected datasets commonly utilized by numerous classic approaches. The rationale behind this choice is threefold. First, these datasets offer a large volume of data for evaluation. Second, they provide standard protocols. Third, they provide  comprehensive benchmark records for existing methods, enabling comprehensive comparison. Numerous papers \cite{rakotosaona2020pointcleannet,hermosilla2019total,luo2020differentiable,luo2021score,pistilli2020learning,yang2018foldingnet,AtlasNet2018papier,zhang2019crn,xie2020grnet,zhou2022seedformer,li2019rgbd,li2020attention,dong2023cvsformer,yifan2019patch,qian2021pu,qiu2022pu,qian2020pugeo,wang2023sspu,wu2023scoda} have conducted experiments on these datasets.
\subsection{Metrics}
\begin{table*}[t]
        \tiny
        \centering
        \caption{Summary of popular evaluation metrics for 3D point cloud enhancement. The symbol $\downarrow$ means a lower metric indicator indicates a better result, and as to $\uparrow$ is the opposite case.}
        \vspace{-0.2cm}
        \resizebox{\textwidth}{!}{
          \begin{tabular}{cccc}
        \toprule[0.5pt]
        Metric& Task& Objective & Remark\\
          \midrule
CD1 \& CD2 & upsampling \& denoising \& completion & $\downarrow$  & -                                        \\
DCD                  & completion \& upsampling              & $\downarrow$  & -                                        \\
HD                               & upsampling \& denoising \& completion & $\downarrow$  & -                                        \\
EMD                          & upsampling \& denoising \& completion & $\downarrow$  & the cardinalities of two point clouds should be identical   \\
P2F \& P2M & upsampling \& denoising \& completion & $\downarrow$  & the ground-truth mesh is required              \\
NUC               & upsampling                            & $\downarrow$  & -                                        \\
IoU                     & completion                            & $\uparrow$  & -                                        \\
mIoU                     & completion                            & $\uparrow$  & -                                        \\
  \bottomrule[0.5pt]
  \end{tabular}}
\label{metricstable}
\vspace{-0.3cm}
\end{table*}
To effectively evaluate the performance of various methods across different datasets, a range of evaluation metrics derived from distributional differences and machine learning practices are employed. 
The commonly used evaluation metrics for point cloud enhancement are reported in Table \ref{metricstable}.
\subsubsection{\textbf{Chamfer Distance}}
For Chamfer distance (CD), CD $l_{1}$ and CD $l_{2}$ are two representative CD metrics. They compute $l_{1}$ distance and $l_{2}$ distance between each point in one set and the nearest point to that point in the other set, respectively. The former is defined as:
\begin{equation}
\begin{aligned}
    d_{CD1}\left(P,Q\right)=&\frac{1}{\left|P\right|}\sum_{x\in P} \min_{y\in Q} \left\|x-y\right\|_{2}\\&+\frac{1}{\left|Q\right|}\sum_{y\in Q}\min_{x\in P}\left\|x-y\right\|_{2},
\end{aligned}
\end{equation}
where $P,Q$ denote two point clouds. The latter is defined as:
\begin{equation}
\begin{aligned}
    d_{CD2}\left(P,Q\right)=&\frac1{\left|P\right|}\sum_{x\in P}\min_{y\in Q}\left\|x-y\right\|_2^2\\&+\frac1{\left|Q\right|}\sum_{y\in Q}\min_{x\in P}\left\|x-y\right\|_2^2.
\end{aligned}
\end{equation}
\par Additionally, density-aware Chamfer Distance (DCD) \cite{wu2021density}, which is introduced for point cloud completion at first, is defined as:
\begin{multline}
d_{DCD}(P,Q) = \frac{1}{2} \left( \frac{1}{|P|} \sum_{x \in P} \left(1 - \frac{1}{n_{\hat{y}}} e^{-\alpha ||x - \hat{y}||_2} \right) \right. \\
\left. + \frac{1}{|Q|} \sum_{y \in Q} \left(1 - \frac{1}{n_{\hat{x}}} e^{-\alpha ||y - \hat{x}||_2} \right) \right),
\end{multline}
where $\hat{x}=\min_{x\in P}||y-x||_2$, $\hat{y}=\min_{y\in Q}||x-y||_2$,  and the symbol $\alpha$ represents a scalar quantity.
DCD is a derivative of CD that is capable of detecting differences in density distributions. DCD focuses on both the macroscopic organization and the specific geometric characteristics.

\subsubsection{\textbf{Hausdorff Distance}}
Hausdorff Distance (HD) is defined as:
\begin{equation}
\begin{aligned}
\mathsf{d}_{\mathrm{HD}}\left(P,Q\right)&=\max\left\{\max_{x\in P}\min_{y\in Q}\left\|x-y\right\|_{2},\right.\\&\left.\max_{y\in Q}\min_{x\in P}\left\|x-y\right\|_{2}\right\}.
\end{aligned}
\end{equation}
This metric evaluates the greatest of all the distances from a point in one point cloud to the closest point in the other point cloud, thus capturing the worst-case scenario.
\subsubsection{\textbf{Earth Mover's Distance}}
Earth Mover's Distance (EMD) is outlined as follows:
\begin{equation}
d_{\mathrm{EMD}}\left(P,Q\right)=\min_{T:P\to Q}\sum_{x\in P}\left\Vert x-T(x)\right\Vert_2.
\end{equation}
EMD, rooted in optimal transport theory, requires an equal number of points in both sets and searches for a bijection. $T:P \to Q$ that minimizes the overall distance, making it computationally intensive yet highly precise.
\subsubsection{\textbf{Point-to-Face and Point-to-Mesh Distances}}
The Point-to-Face (P2F) and Point-to-Mesh (P2M) distances are defined to measure discrepancies relative to the surface structure:
\begin{equation}
d_{\mathrm{P2F}}\left(P,Q\right)=\frac1{\left|P\right|}\sum_{x\in P}\left\|x-y\left(x\right)\right\|_2,
\end{equation}
where $y(x)=\arg\min_{\tilde{y}\in Q}\left\|x-\tilde{y}\right\|_2$, and
\begin{equation}
d_{\mathrm{P2M}}(P, M) = \min_{Q \in M} \| P - Q \|.
\end{equation}
These metrics specifically evaluate the distance from points to an underlying surface or mesh, offering insights into geometric fidelity.
\subsubsection{\textbf{Normalized Uniformity Coefficient}}
To address the uniformity of point sampling, the normalized uniformity coefficient (NUC) is employed. $D$ equal-size disks are put on object surface first and NUC computed over $K$ objects is defined by:
\begin{equation}
NUC=\sqrt{\frac1{K\times{D}}\sum_{k=1}^K\sum_{i=1}^D\left(\frac{n_i^k}{N^k\times{p}}-avg\right)^2},
\end{equation}
\begin{equation}
avg=\frac1{K\times{D}}\sum_{k=1}^K\sum_{i=1}^D\frac{n_i^k}{N^k\times{p}},
\end{equation}
where $n_i^k$ is the \textit{i}-th disk of of the \textit{k}-th object, $N^k$ is the total number of points on the $k$-th object, and $p$ is the percentage of the disk area over the total object surface area. This metric evaluates the distribution of points across designated areas on the object's surface, thus quantifying spatial homogeneity.\par

\begin{figure*}[t]
    \centering
    \includegraphics[width=\linewidth]{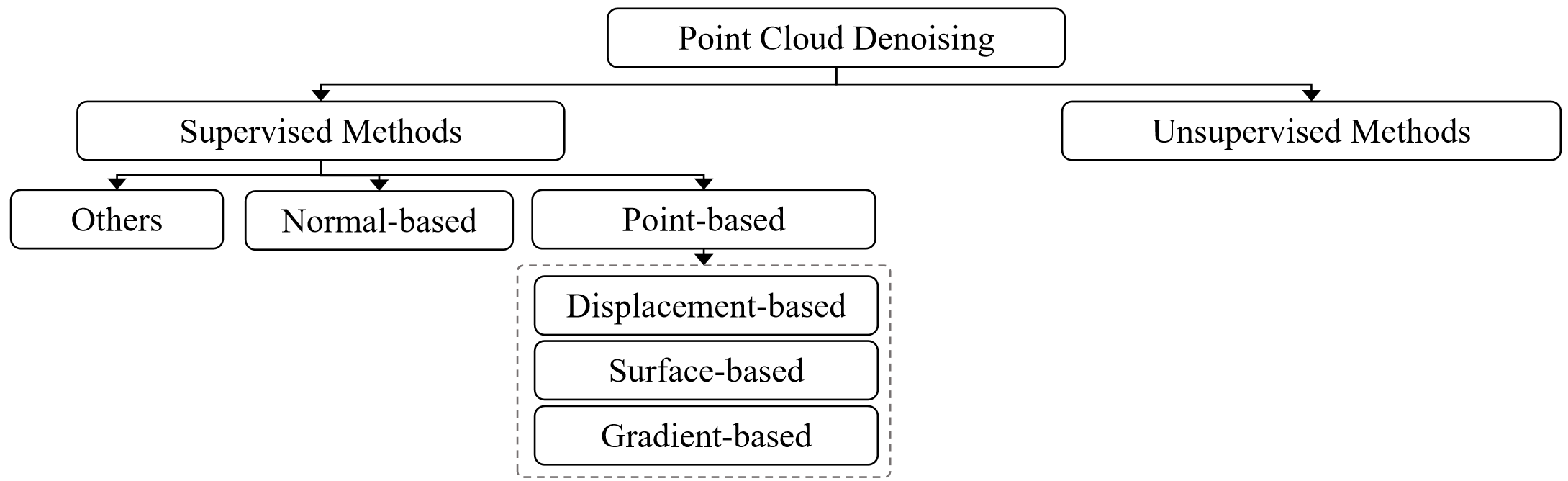}
    \caption{A detailed taxonomy of point cloud denoising methods.}
    \label{denoising1}
\end{figure*}

\subsubsection{\textbf{Mean Intersection of Union}}
The intersection of union (IoU) metric is defined by:
\begin{equation}
IoU = \frac{{TP}}{{TP + FP + FN}},
\end{equation}
where TP (true positives) represents the number of correctly predicted positive instances, FP (false positives) represents the number of negative instances incorrectly classified as positive, and FN (false negatives) represents the number of positive instances incorrectly classified as negative. 

Specifically, IoU is used to predict voxel labels compared to ground truth labels. For the semantic scene completion task, IoU is evaluated for each object class on both observed and occluded voxels. For the scene completion task, we treat all non-empty object classes as a single category and evaluate the IoU of the binary predictions on occluded voxels. Moreover, the mean of IoU (mIoU) is used to calculate the average IoU of all classes.

Standard machine learning metrics such as the Precision, Recall, and F1-score are also commonly used to evaluate the effectiveness of point cloud enhancement techniques.\par
It is important to note that the experimental results presented in our paper are derived from existing literature. In cases where the original studies did not include certain experiments, we have referenced results from subsequent highly-cited studies to ensure the reliability of the data. The results presented in the table are all under a consistent setting. If there is any deviation, we have specified in the table description which methods have different settings. Furthermore, to enhance reproducibility, we have included a table in each section indicating the availability of source-code of the discussed methods.

\section{3D Point Cloud Denoising}
Point cloud denoising aims to eliminate or significantly reduce noise from a noisy input point cloud, thereby closely approximating the true surface geometry.\par
\begin{figure*}
    \centering
    \includegraphics[width=\textwidth]{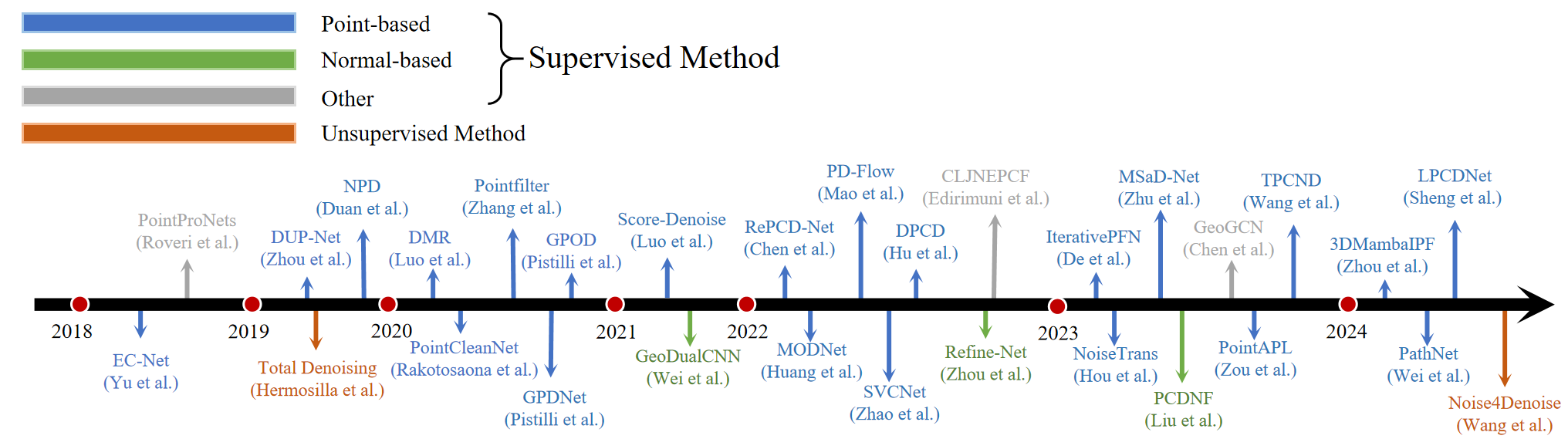}
    \caption{Chronological overview of representative point cloud denoising networks.}
    \label{denoising2}
\end{figure*}
\begin{figure*}[t]
    \centering
    \includegraphics[width=\textwidth]{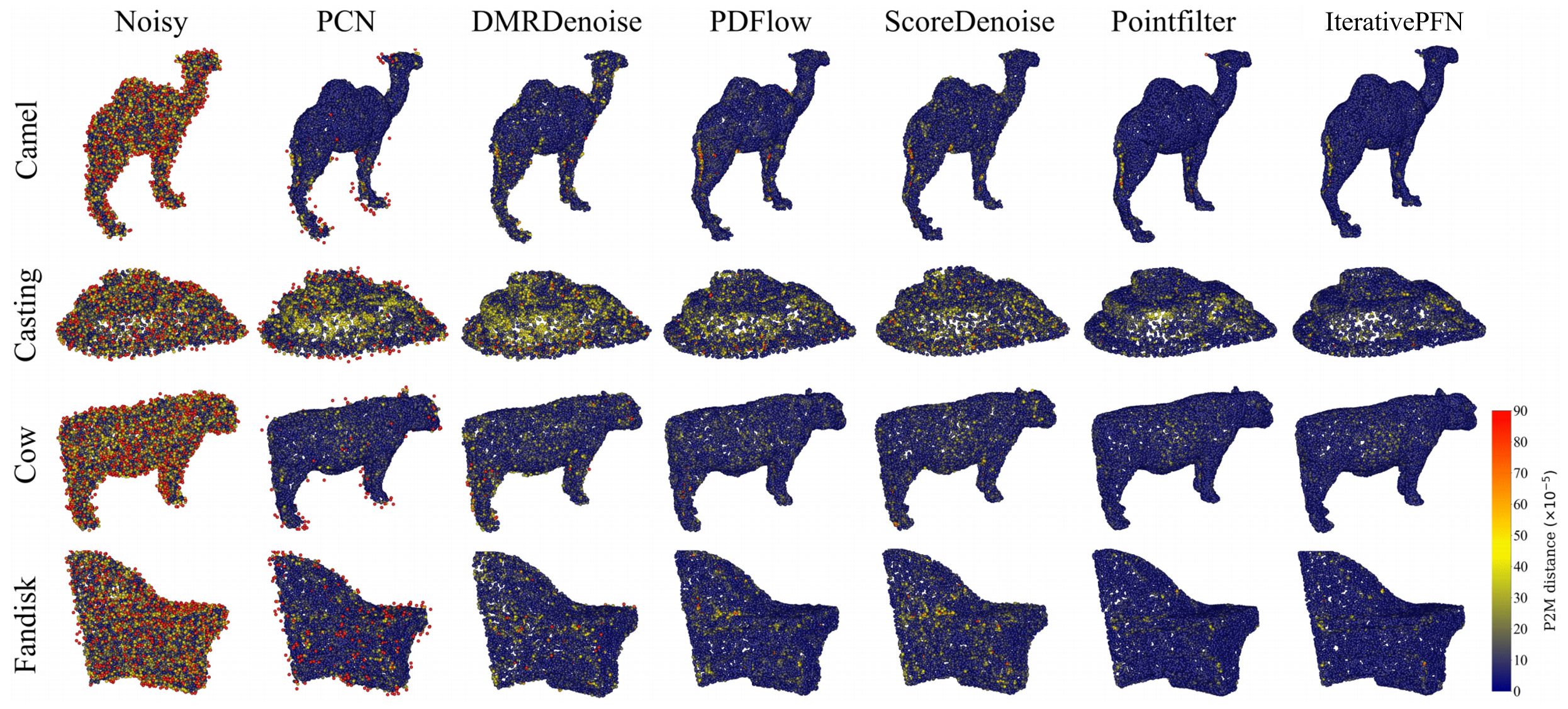}
    \caption{Visual results of point-wise P2M distance for 50K resolution shapes with Gaussian noise at 2\% of the bounding sphere radius. These results are derived from IterativePFN \cite{de2023iterativepfn}.}
    \label{denoisingvisual}
\end{figure*}

We classify existing deep-learning-based denoising methods into supervised and unsupervised categories, as illustrated in Fig. \ref{denoising1}. A chronological overview of point cloud denoising networks is illustrated in Fig. \ref{denoising2}. Point-based methods have emerged as the dominant approach for point cloud denoising tasks. By contrast, normal-based and unsupervised methods are relatively less common. Besides, it is a tendency to study unsupervised point cloud denoising. The experimental results of the point cloud denoising networks are presented in Table \ref{denoisetable}, and some visual results are depicted in Fig. \ref{denoisingvisual}.

\subsection{Supervised Methods}
Supervised point cloud denoising networks can be further categorized into four primary methodologies: point-based, normal-based, multi-task, and other methods. 
\subsubsection{\textbf{Point-based}} 
\begin{figure}[t]
    \centering
    \begin{subfigure}{\linewidth}
    \centering
    \includegraphics[width=\linewidth]{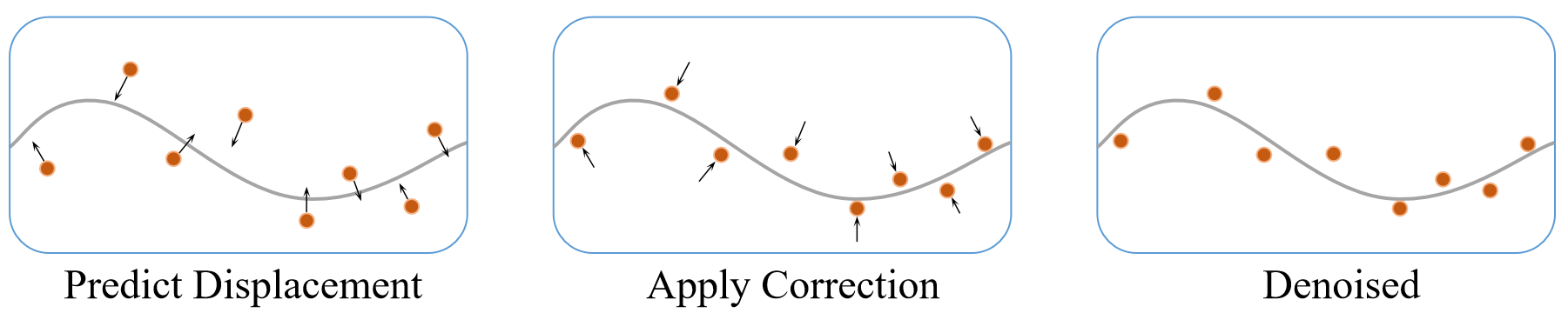}
    \caption{Displacement-based}
    \label{displacement}
    \end{subfigure}
    \centering
    \begin{subfigure}{\linewidth}
    \centering
    \includegraphics[width=\linewidth]{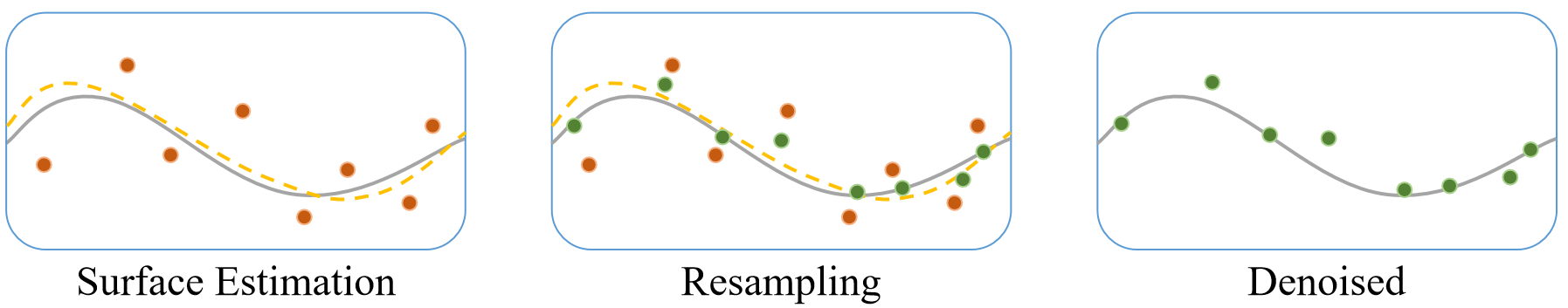}
    \caption{Surface-based}
    \label{surface}
    \end{subfigure}
    \centering
    \begin{subfigure}{\linewidth}
    \centering
    \includegraphics[width=\linewidth]{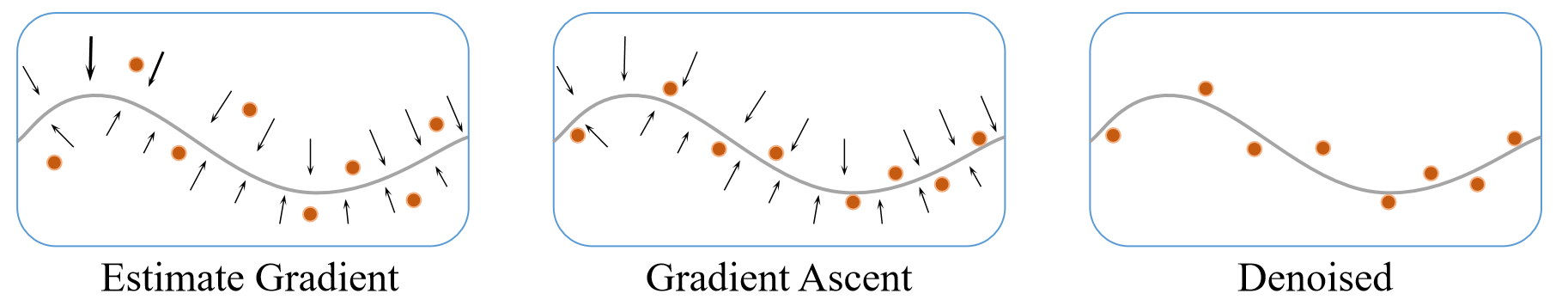}
    \caption{Gradient-based}
    \label{score}
    \end{subfigure}
    \caption{Illustration of the point-based methods. (a) Displacement-based methods. (b) Surface-based methods. (c) Gradient-based methods.}
    \label{pointbased}
\end{figure}

These methods focus solely on the spatial coordinates of points in the noisy point cloud. They primarily utilize the point positions, ignoring other possible attributes or contextual information. This category can be further divided into displacement-based, surface-based and the others.
\paragraph{\textbf{Displacement-based}} These methods design networks to estimate the spatial displacements of noisy points from a raw point cloud, in order to rectify noisy points, as is shown in Fig. \ref{displacement}.\par
Based on auto-encoder and PointNet \cite{qi2017pointnet}, Pointfilter \cite{zhang2020pointfilter} processes individual points and their neighbors, learning a latent representation to determine displacement vectors. Observing that point clouds often comprise self-similar local structures in different potentially distant locations, Huang et al. \cite{huang2020non} implemented a non-local learning unit equipped with a graph attention module. This approach progressively refines noise across multiple units while maintaining the integrity of the surface. GPDNet \cite{pistilli2020learning} utilizes dynamic graph convolution for effective denoising. \par
Due to the varying and intricate nature of noise in point clouds, many techniques struggle to maintain a balance between effective denoising and the preservation of fine details, particularly under high noise conditions. Additionally, point clouds produced by these methods often suffer from non-uniform distribution. To solve this problem, SVCNet \cite{zhao2022noise} proposes a `self-variation' point cloud by perturbing existing noise, maintaining the underlying clean signal but altering noise characteristics. This method filters out noise by identifying common features between two noisy datasets using an attention mechanism, and minimizes smoothing effects with an edge constraint module. RePCD-Net \cite{chen2022repcd} features a feature-aware recurrent architecture, incorporating a bi-directional RNN for adaptive feature extraction and a recurrent propagation layer to maintain geometric details across denoising stages. MODNet \cite{huang2022modnet} extracts features from varying scale patches and integrates geometric information across these scales with a multi-scale perception module, utilizing a multi-offset decoder for denoising displacements. This network dynamically adjusts the influence of each scale based on the geometric information gathered. PD-Flow \cite{mao2022pd} employs normalizing flows and noise disentanglement techniques, treating the noisy point cloud as a joint distribution of clean points and noise. The objective is to separate these components through a latent space encoding, noise filtering, and decoding to recover clean points. IterativePFN \cite{de2023iterativepfn} incorporates an iterative filtering process within the network architecture, featuring multiple IterationModules. These modules simulate successive filtering iterations. Some methods leverage transformer  architectures for feature encoding, such as NoiseTrans \cite{hou2023noisetrans}, TPCDN \cite{wang2023transformer}, and MSaD-Net \cite{zhu2023msad}. Meanwhile, MSaD-Net additionally introduces a channel attention module to further optimize the denoising process by focusing on the relationship between different feature channels of the point cloud data.\par
Drawing inspiration from the introduction of state space models for long sequence modeling in natural language processing, 3DMambaIPF \cite{zhou20243dmambaipf} integrates the Mamba module, is adept at efficiently managing long sequence data. This network partitions the input point cloud into manageable patches, which are then sequentially processed through MambaDenoise modules. Each module is equipped with Mamba encoders and decoders, meticulously designed to amplify positional features, thereby enhancing the overall data processing capability. PathNet \cite{wei2024pathnet} adopts a path-selective strategy based on reinforcement learning, dynamically choosing the optimal denoising path for each point to accurately align with its underlying surface. LPCDNet \cite{sheng2024denoising} is a lightweight network comprising three modules: a feature extraction module using trigonometric functions for capturing detailed local context, a non-parametric feature aggregation module for global pattern understanding through semantic similarities, and a decoder module that realigns noisy points to their correct positions. Mao et al. \cite{mao2024denoising} introduced a multi-level graph convolution network that captures geometric features across various scales, which are then integrated into an invertible neural network to achieve precise 3D noise disentanglement. By employing an invertible monotone operator, this approach effectively segregates noise from clean points within the latent features, projecting them onto distinct channels. Consequently, it enables the masking of noise channels and the restoration of clean points via inverse transformation. StraightPCF \cite{de2024straightpcf} filters noisy points along straight paths, thereby minimizing discretization errors and accelerating convergence to clean surfaces. This method leverages a velocity module to infer a constant flow velocity and a distance module to scale the trajectory, ensuring precise convergence.\par
\begin{figure*}[t]
    \centering
    \includegraphics[width=\linewidth]{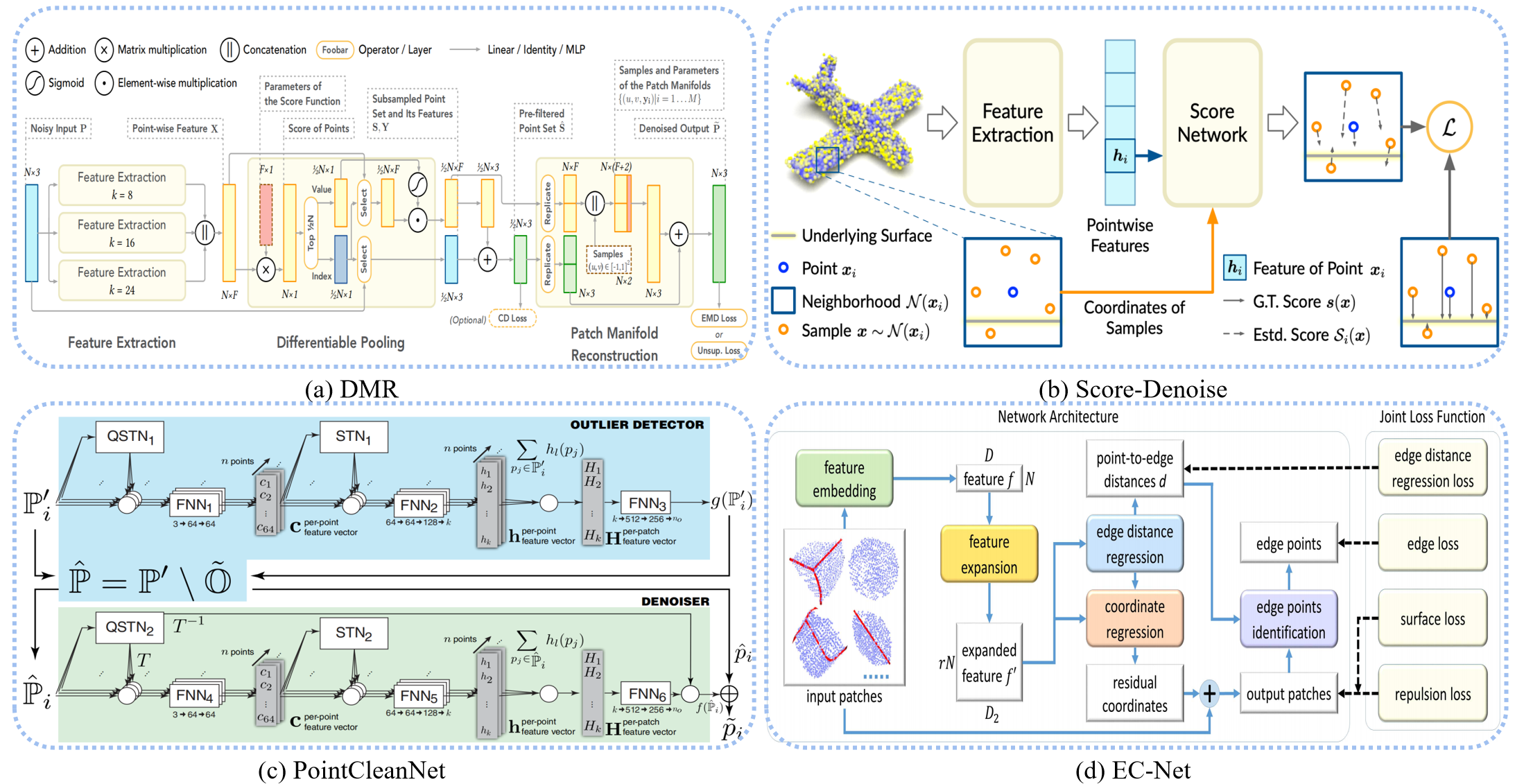}
    \caption{Pipelines of representative point cloud denoising methods.}
    \label{denoisenet}
\end{figure*}

In addition to denoising, several methods also perform tasks such as outlier removal and upsampling. PointCleanNet \cite{rakotosaona2020pointcleannet} is a pioneering learning architecture specifically designed for the direct cleaning of raw point clouds. It incorporates a quaternion spatial transformer network to orient patches of points into a canonical position. This is achieved with PointNet, which effectively distinguishes true surface points (inliers) from noise or artifacts (outliers), as illustrated in Fig. \ref{denoisenet}c. Subsequently, Pistilli et al. \cite{pistilli2020learningrobust} enhanced the model's robustness by leveraging a weighted cross-entropy loss, adeptly addressing the imbalance between the relatively few outliers and the predominant set of inliers. Outlier removal helps to eliminate data points that significantly deviate from the expected norm, thereby cleaning the data and improving its reliability.\par
There are also methods that perform point cloud denoising in cooperation with other tasks, such as resampling and upsampling, to enhance the quality of the point cloud. As the pioneering edge-aware consolidation network, EC-Net \cite{yu2018ec} is trained for heightened sensitivity towards edges, as depicted in Fig. \ref{denoisenet}d. It employs a regression mechanism to precisely recover point coordinates and their neighbors relative to edges. The network is distinctive for its edge-aware joint loss function, meticulously designed to minimize distances to 3D meshes and refine sharp edge representations. Concurrently, a notable work is DUP-Net \cite{zhou2019dup}, which combines denoising and upsampling capabilities within a single framework. This integration can simultaneously enhance point resolution and eliminate outliers, achieved via an adversarial strategy.
\begin{table}[htbp]
  \centering
    \caption{Code URL for point cloud denoising networks. The symbol `-' indicates that the code is unavailable.}
    \vspace{-0.2cm}
    \resizebox*{\linewidth}{!}{
    \begin{tabular}{|c|c|c|}
    \hline
    Year  & Method & URL \\
    \hline
    2019  & Total Denoising \cite{hermosilla2019total} & https://github.com/phermosilla/TotalDenoising \\
    \hline
    \multirow{4}[0]{*}{2020} & GPDNet \cite{pistilli2020learning}& - \\
    \cline{2-3}
          & Pointfilter \cite{zhang2020pointfilter}& https://github.com/dongbo-BUAA-VR/Pointfilter \\
    \cline{2-3}
          & DMR \cite{luo2020differentiable}& https://github.com/luost26/DMRDenoise \\
    \cline{2-3}
          & PointCleanNet \cite{rakotosaona2020pointcleannet}& https://github.com/mrakotosaon/pointcleannet \\
    \hline
    2021  & Score-denoise \cite{luo2021score}& https://github.com/luost26/score-denoise \\
    \hline
    \multirow{2}[0]{*}{2022} & PD-Flow \cite{mao2022pd}& https://github.com/unknownue/pdflow \\
    \cline{2-3}
          &  SVCNet \cite{zhao2022noise}& - \\
    \hline
    \multirow{2}[0]{*}{2023} & IterativePFN \cite{de2023iterativepfn}& https://github.com/ddsediri/IterativePFN \\
    \cline{2-3}
          & NoiseTrans \cite{hou2023noisetrans}& - \\
    \hline
    \multirow{5}[0]{*}{2024} & 3DMambaIPF \cite{zhou20243dmambaipf}& - \\
    \cline{2-3}
          & LPCDNet \cite{sheng2024denoising}& - \\
    \cline{2-3}
          & Noise4Denoise \cite{wangnoise4denoise}& - \\
    \cline{2-3}
          & PD-LTS \cite{mao2024denoising} & https://github.com/yanbiao1/PD-LTS \\
    \cline{2-3}
          & StraightPCF \cite{de2024straightpcf} & https://github.com/ddsediri/StraightPCF \\
    \hline
    \end{tabular}}
  \label{denoisingurl}%
\end{table}
\begin{table*}[t]
  \centering
  \caption{Comparative point cloud denoising results on the PU-Net dataset. The symbol `-' means the result is unavailable, and `-U' indicates unsupervised style. The CD and P2M values are multiplied by $10^{5}$.}
  \vspace{-0.2cm}
  \resizebox*{\textwidth}{!}{
    \begin{tabular}{|c|c|c|c|c|c|c|c|c|c|c|c|c|c|c|c|}
    \hline
    \multirow{4}[0]{*}{Year} & \multirow{4}[0]{*}{Method} & \multicolumn{14}{c|}{PU-Net Dataset} \\
    \cline{3-16}
          &  & \multicolumn{8}{c|}{10K points}                        & \multicolumn{6}{c|}{50K points} \\
    \cline{3-16}
          &  & \multicolumn{2}{c|}{1\% noise} & \multicolumn{2}{c|}{2\% noise} & \multicolumn{2}{c|}{2.5\% noise} & \multicolumn{2}{c|}{3\% noise} & \multicolumn{2}{c|}{1\% noise} & \multicolumn{2}{c|}{2\% noise} & \multicolumn{2}{c|}{2.5\% noise} \\
    \cline{3-16}
          &  & CD & P2M & CD & P2M & CD & P2M & CD & P2M & CD & P2M & CD & P2M & CD & P2M \\
    \hline
    2019  & Total Denoising \cite{hermosilla2019total} & 33.9  & 8.26  & 72.51 & 34.85 & -     & -     & -     & -     & 10.24 & 3.14  & 27.22 & 15.67 & -     & - \\
    \hline
    \multirow{5}[0]{*}{2020} & GPDNet \cite{pistilli2020learning} & 23.10  & 7.14  & 42.84 & 18.55 & 58.37 & 30.66 & -     & -     & 10.49 & 6.35  & 32.88 & 25.03 & 50.85 & 41.34 \\
    \cline{2-16}
          & Pointfilter \cite{zhang2020pointfilter}& 24.61 & 7.30   & 35.34 & 11.55 & 40.99 & 15.05 & -     & -     & 7.58  & 4.32  & 9.07  & 5.07  & 10.99 & 6.29 \\
          \cline{2-16}
          & DMR \cite{luo2020differentiable} & 47.12 & 21.96 & 50.85 & 25.23 & 52.77 & 26.69 & 58.92 & 28.46 & 12.05 & 7.62  & 14.43 & 9.70   & 16.96 & 11.9 \\
          \cline{2-16}
          & DMR -U \cite{luo2020differentiable}& 53.13 & 25.22 & 64.55 & 33.17 & -     & -     & 51.34 & 46.67 & 12.26 & 5.21  & 21.38 & 12.51 & -     & - \\
          \cline{2-16}
          & PointCleanNet \cite{rakotosaona2020pointcleannet}& 36.86 & 15.99 & 79.26 & 47.59 & 104.86 & 69.87 & 130.67 & 87.37 & 11.03 & 6.46  & 19.78 & 13.70  & 32.03 & 24.86 \\
          \hline
    \multirow{2}[0]{*}{2021} & Score-denoise \cite{luo2021score}& 25.22 & 7.54  & 36.83 & 13.8  & 42.32 & 19.04 & 47.08 & 19.42 & 7.16  & 4.00   & 12.89 & 8.33  & 14.45 & 9.58 \\
    \cline{2-16}
          & Score-U \cite{luo2021score}& 31.07 & 8.88  & 46.75 & 18.29 & -     & -     & 72.25 & 37.62 & 9.18  & 2.65  & 24.39 & 14.11 & -     & - \\
          \hline
    \multirow{2}[0]{*}{2022} & PD-Flow \cite{mao2022pd}& 21.26 & 6.74  & 32.46 & 13.24 & 36.27 & 17.02 & 44.5  & -     & 6.51  & 4.16  & 12.7  & 9.21  & 18.74 & 14.26 \\
    \cline{2-16}
          & SVCNet \cite{zhao2022noise} & 24.12 & 25.37 & 34.32 & 28.19 & - & - & 48.64 & 36.26 & -     & -     & -     & -     & -     & - \\
          \hline
    \multirow{2}[0]{*}{2023} & IterativePFN \cite{de2023iterativepfn} & 20.55 & 5.01  & 30.43 & 8.43  & 33.53 & 10.46 &       & -     & 6.05  & 3.02  & 8.03  & 4.36  & 10.15 & 5.88 \\
    \cline{2-16}
          & NoiseTrans \cite{hou2023noisetrans}& 22.88 & 3.60   & 32.51 & 8.43  & - & - & 40.70  & 14.70  & -     & -     & -     & -     & -     & - \\
          \hline
    \multirow{5}[0]{*}{2024} & 3DMambaIPF \cite{zhou20243dmambaipf}& 19.89 & 4.77  & 29.95 & 8.03  & 32.62 & 9.92  & -     & -     & 5.89  & 2.91  & 7.55  & 4.05  & 9.28  & 5.31 \\
    \cline{2-16}
          & LPCDNet \cite{sheng2024denoising}& 22.65 & 5.89  & 31.47 & 10.10  & -     & -     & -     & -     & 6.03  & 3.26  & 9.46  & 5.01  & -     & - \\
          \cline{2-16}
          & Noise4Denoise \cite{wangnoise4denoise}& 22.04 & 3.54  & 34.49 & 9.30   & -     & -     & -     & -     & 7.00     & 1.34  & 10.56 & 3.74  & -     & - \\
          \cline{2-16}
           &  PD-LTS \cite{mao2024denoising} & 17.81 & 4.65  & 24.41 & 7.58   & 26.99     & 9.67     & -     & -     & 4.70     & 2.95  & 6.46 & 4.25  & 8.63     & 5.81 \\
           \cline{2-16}
           & StraightPCF \cite{de2024straightpcf} & 18.70 & 2.39  & 26.44 & 6.04   & -     & -   & 32.87     & 11.26     & 5.62     & 1.11  & 7.65 & 2.66  & -     & - \\
    \hline
    \end{tabular}}
  \label{denoisetable}
\end{table*}
\paragraph{\textbf{Surface-based}} These methods aim to reconstruct surfaces from noisy point clouds, followed by sampling on the latent surfaces for denoising, as is shown in Fig. \ref{surface}.\par
NPD \cite{duan20193d} directly estimates reference planes and achieves denoising through a single projection step. This method improves denoising performance by reducing the influence of noise on the estimation process, and eliminates the need for defining neighboring points or performing eigen-decomposition. When only moving noisy points without properly reconstructing the underlying surface, it often leads to less-than-ideal denoising results. To deal with this issue, Luo et al. \cite{luo2020differentiable} presented a novel paradigm of reconstructing the underlying manifold of noisy point clouds. It leverages an adaptive differentiable pooling operation to selectively sample points with low noise levels, which are closer to the true underlying surface. These points are then used to infer and reconstruct local surface geometries or patch manifolds from embedded neighborhood features, as shown in Fig. \ref{denoisenet}a.
\paragraph{\textbf{Gradient-based}} These approaches utilize ascent techniques to iteratively update 3D point positions by calculating the gradient of the objective function, as illustrated in Fig. \ref{score}. These methods are also commonly referred to as score-based methods.\par
Motivated by the distributional properties of noisy point clouds, Luo et al. \cite{luo2021score} applied gradient ascent with the log-probability function. For a noisy point within an initial noisy patch, the training objective of ScoreDenoise is defined as the expectation over the square of the L2 norm between the predicted score and the ground truth score. Specifically, ScoreDenoise is trained to predict a score for points sampled from the k-nearest neighbors of a given noisy point. This score is conditioned on the latent feature of the noisy point and corresponds to the gradient of the log of the noise-convolved probability distribution for the point's position. The ground truth target is defined as the displacement from the point to its nearest neighbor in a clean patch. 
Meanwhile, for dynamic point cloud denoising, Hu et al. \cite{hu2022dynamic} utilized dynamic gradient fields and exploited temporal relationships between consecutive frames. It adjusts each point cloud patch as a rigid body within the gradient field formed by adjacent frames, continuing until a state of equilibrium is reached. This equilibrium signifies that the patch has been effectively aligned with the clean surface.
\subsubsection{\textbf{Normal-based}}
Normal-based methods excel in capturing surface variations, offering a more precise description than methods focusing solely on point position alterations. These techniques typically initiate by refining the normals of the point cloud. Following this initial step, point positions are adjusted to more accurately conform to the enhanced normals.\par
Lu et al. \cite{lu2020deep} developed a technique that enhances both normal estimation and point cloud filtering while preserving significant features. Their methodology involves creating a specialized dataset, classifying points into feature-rich and feature-poor categories, and employing distinct normal estimation networks tailored to these categories. In practical application, this method automatically derives normals from a given noisy point cloud and iteratively updates point positions based on these normals.
Motivated by the trend in computer vision  utilizing additional
task-specific information (i.e., domain knowledge) to improve the performance of base deep learning models, Wei et al. \cite{wei2021geodualcnn} presented GeoDualCNN. It is a dual convolutional neural network designed to support geometric analysis. This model is predicated on the assumption that the underlying surface of a noisy point cloud is piece-wise smooth, and maintaining accurate normal orientation is crucial for ensuring local surface continuity. The architecture features a homogeneous neighborhood to effectively isolate surface discontinuities, coupled with a dual-branch structure aimed at simultaneously removing noise and preserving essential features.

Unlike traditional methods that rely on specific priors or hand-crafted features, Refine-Net \cite{zhou2022refine} utilizes a multi-feature scheme to refine initial normals by integrating additional geometric information through several feature modules. The network architecture includes a new connection module for blending these features effectively. It also proposes a multi-scale fitting patch selection for initial normal estimation. PCDNF \cite{wang2023transformer} introduces  a shape-aware selector and a feature refinement module, to improve denoising by taking into account the relationship between point and normal features and their geometric priors. 
\subsubsection{\textbf{Others}} Edirimuni et al. \cite{edirimuni2022contrastive} presented a method utilizing a 3D patch-based contrastive learning framework to train a feature encoder. These representations inform a regression network supervised by a joint loss function that simultaneously addresses point normal and displacement estimation. GeoGCN \cite{chen2023geogcn} utilizes two types of surface normals: real normal (RN) and virtual normal (VN). It combines local and global geometric information to enhance denoising performance, with RNs helping to preserve local details and VNs aiding in maintaining the overall shape. The network operates through a dual-domain approach. It  first regresses point positions using spatial-based graph convolutions aided by VNs, and then estimates RNs via principal component analysis. Finally, it refines these RNs using normal-based graph convolutions. PointAPL \cite{zuo2023attention} uses an attention mechanism, specifically designed for outlier removal of rocket tank panels. PointProNets \cite{roveri2018pointpronets} leverages convolutional neural networks to tackle the inherent challenges of sparse and noisy point cloud data. It aims to faithfully reconstruct underlying surfaces by transforming unordered points into systematically sampled height maps.
\subsection{Unsupervised Methods} Because obtaining ground-truth clean data in many scenarios is not feasible, unsupervised techniques focus on denoising point clouds directly without relying on the guidance of clean point clouds. \par
The research toward unsupervised PCD is still at an early stage. 
Total denoising \cite{hermosilla2019total} operates under the premise that denser point clusters are more indicative of the true surface. Nevertheless, this technique struggles to maintain geometric details due to insufficient feature clarity during training. In pursuit of understanding the geometric structures within point clouds, Chen et al. \cite{chen2019deep} developed an unsupervised graph-based auto-encoder that employs folding techniques, graph topology deductions, and graph filtering to succinctly represent unorganized 3D point clouds. The decoding process of this model uses an adaptable graph structure to ensure that essential features are retained. Recently, Noise4Denoise \cite{wangnoise4denoise} learns the restoration of clean point clouds from noisy inputs alone, by training on two versions of the same point cloud: one with standard noise level and another with double  noise.
\subsection{Summary}
We briefly summarize the characteristics of representative denoising methods as follows:
\begin{enumerate}
\item \textbf{Point-based methods.} These methods primarily adjust the positions of individual points within the point cloud by predicting displacements or introducing additive noise. These techniques employ deep learning architectures to determine the optimal movements for each point to minimize noise. However, they often encounter issues such as shrinkage and outliers, and typically struggle to preserve sharp geometric features due to their reliance solely on 3D coordinate information. Moreover, some methods integrate denoising with other point cloud processing tasks like upsampling, edge extraction, and outlier removal. By leveraging the synergies between these tasks, these approaches enhance overall performance, exemplified by networks that simultaneously improve point resolution while reducing noise. This integrated approach naturally introduces multiple constraints on the data, leading to more robust denoising results. However, these methods suffer from redundant network parameters and require meticulous adjustments of network architectures to effectively balance various tasks.
\item \textbf{Normal-based methods.} These methods overcome some limitations of point-based approaches by concentrating on the normals of the point surfaces, offering a more precise representation of surface variations. The main challenges with these methods are their complex implementation and the need for additional computations to establish a consistent neighborhood around each point, which may hinder their practical applications.
\item \textbf{Unsupervised methods.} They are desirable choices in real-world applications. These techniques, which may utilize the density of surrounding points, strive to deduce the underlying surface without direct supervision. Although they forego the need for clean ground-truth data, their major drawback is the potential inaccuracy in preserving geometric details, particularly in complex or sparse point clouds.
\item \textbf{Common limitations.} Several challenges persist in current denoising methods. For instance,  the degradation of denoising performance as dataset scale and complexity grow, limited robustness when dealing with different kinds and levels of noise, and the weak generalization ability of current methods.
\end{enumerate}

\section{3D Point Cloud Completion}
Given a point cloud suffering from missing points, the objective of 3D point cloud completion is to infer the missing points and incorporate them into the original point cloud to reconstruct the complete shape of the object or scene.

\begin{figure*}[t]
    \centering
    \includegraphics[width=\textwidth]{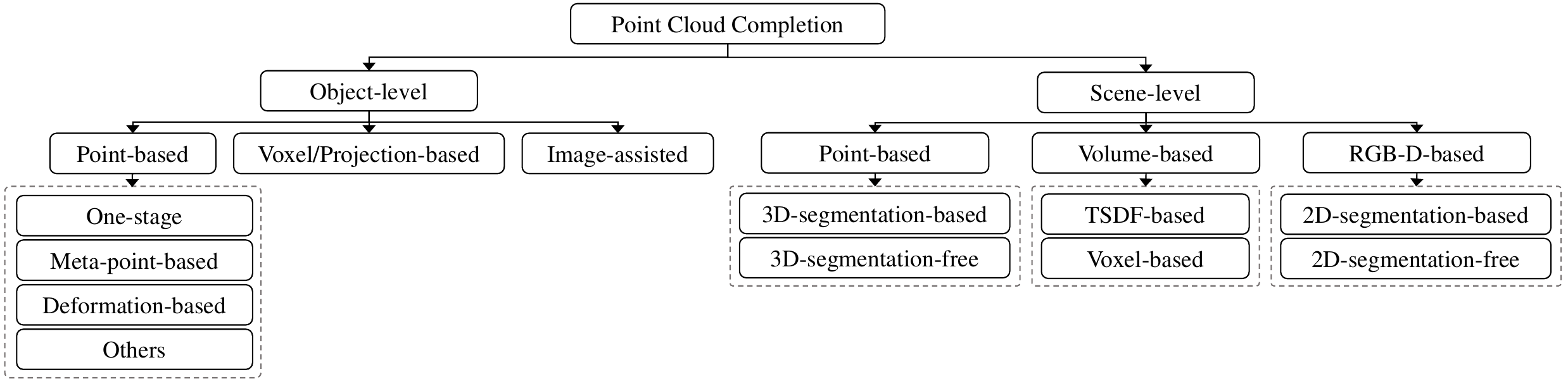}
    \caption{A detailed taxonomy of point cloud completion methods.}
    \label{fig:Completionframe}
\end{figure*}

\begin{figure*}[t]
    \centering
    \includegraphics[width=\textwidth]{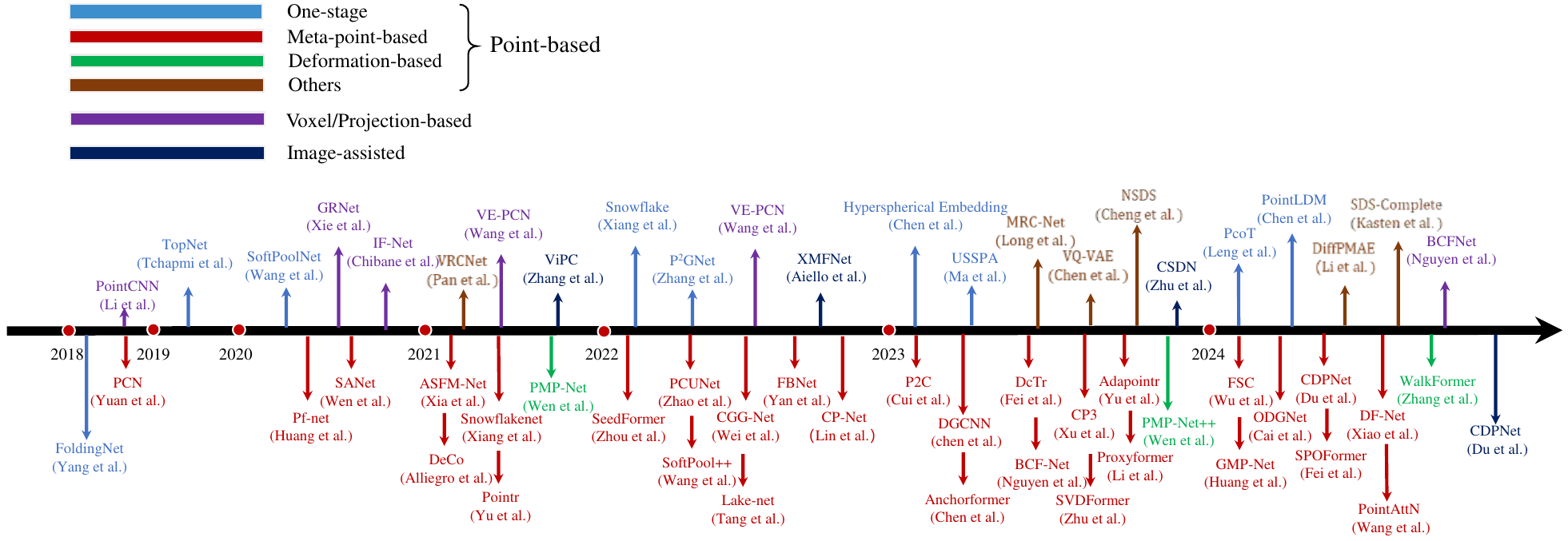}
    \caption{Chronological overview of representative point cloud completion networks for objects.}
    \label{point cloud completion for object}
\end{figure*}

\begin{figure*}[t]
    \centering
    \includegraphics[width=\textwidth]{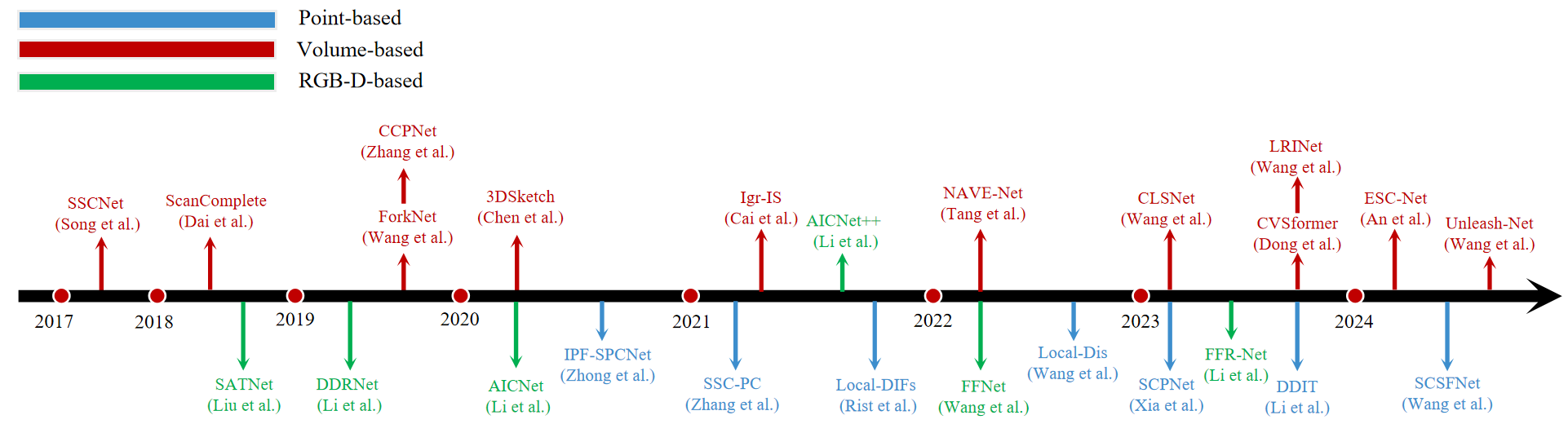}
    \caption{Chronological overview of representative point cloud completion networks for scenes.}
    \label{point cloud completion for scene}
\end{figure*}

\begin{figure*}[t]
    \centering
    \includegraphics[width=\textwidth]{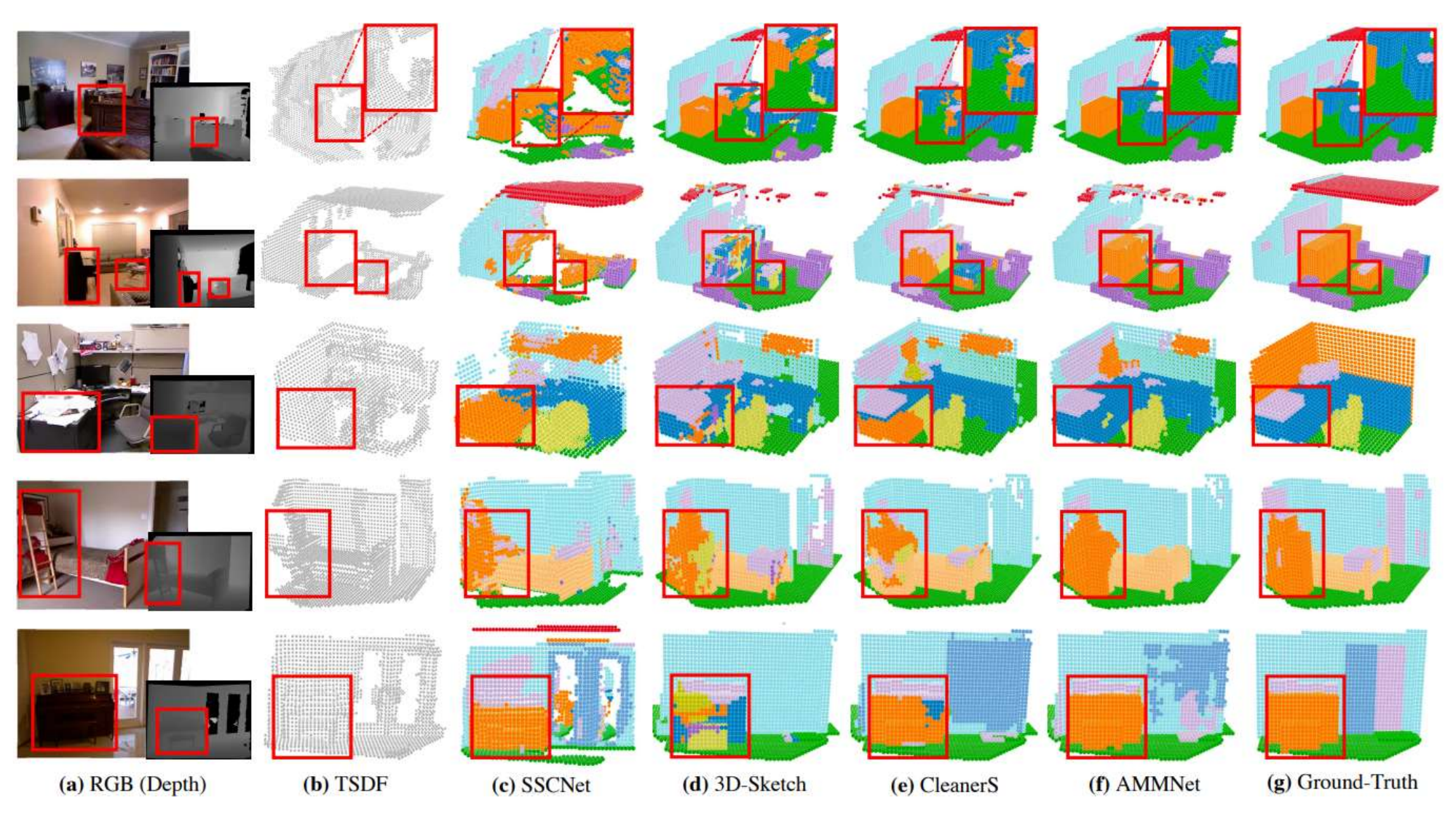}
    \caption{Qualitative comparisons on challenging indoor scenes from the test set of NYU with state-of-the-art methods, including SSCNet \cite{song2017semantic}, 3D-Sketch \cite{chen20203d}, and CleanerS (i. e CLSNet \cite{wang2023semantic}). These results are derived from AMMNet \cite{wang2024unleashing}.}
    \label{scenevisual}
\end{figure*}

Point cloud completion primarily concentrates on completion for objects and scenes. In object completion, the task entails predicting the missing parts of specific point clouds. In scene completion, besides predicting missing points of objects, it is crucial to understand the entire scene context as well. Thus from the perspective of input types, point cloud completion can be categorized into object-level and scene-level completion, as shown in Fig. \ref{fig:Completionframe}. The chronological overviews of object-level and scene-level completion networks are shown in Fig. \ref{point cloud completion for object} and Fig. \ref{point cloud completion for scene}, respectively. As shown in Fig \ref{point cloud completion for object}, point cloud completion for objects mainly uses point-based methods, with voxel-/projection-based and image-assisted methods being relatively less common. Earlier approaches primarily employed one-stage methods, while meta-point-based methods have now become mainstream. Additionally, some deformation-based methods have also emerged. As shown in Fig. \ref{point cloud completion for scene}, RGB-D and volume-based methods were predominantly used in the early stages, while point cloud-based approaches emerged later. In recent years, point cloud-based methods have become increasingly prevalent. The experimental results of object-level and scene-level completion networks are shown in Table \ref{completionobjecttable} and Table \ref{completionscenetable}. More visualizations can be seen in Fig. \ref{scenevisual}

\begin{table*}[t]
\centering
\caption{Comparative results of point cloud completion methods for objects on ShapeNet, PCN, KITTI, MVP and Completion3D datasets. ShapeNet-55 reports the detailed results for each method on 10 categories and the overall results on 55 categories. The value of CD is computed on 2048 points and multiplied by $10^4$, and we report F-Score@1\% metric. ‘-’ indicates that the result is unavailable.}

\vspace{-0.2cm}
\resizebox*{\textwidth}{!}{
\begin{tabular}{|c|c|cc|c|ccc|cc|cc|}
\hline
\multirow{2}{*}{Year}     & \multirow{2}{*}{Model} & \multicolumn{2}{c|}{ShapeNet-55}      & PCN    & \multicolumn{3}{c|}{KITTI}                                         & \multicolumn{2}{c|}{MVP}                               & \multicolumn{2}{c|}{Completion3D}    \\ \cline{3-12} 
                      &                        & \multicolumn{1}{c|}{F-Score} & CD     & CD     & \multicolumn{1}{c|}{MMD}   & \multicolumn{1}{c|}{Fidelity} & FD    & \multicolumn{1}{c|}{CD(4096 points)} & CD(2048 points) & \multicolumn{1}{c|}{CD}    & F-Score \\ \hline
\multirow{3}{*}{2018} & FoldingNet \cite{yang2018foldingnet}            & \multicolumn{1}{c|}{0.082}   & 0.312  & 1.274 & \multicolumn{1}{c|}{0.614} & \multicolumn{1}{c|}{1.892}    & 7.467 & \multicolumn{1}{c|}{8.760}            & 10.540           & \multicolumn{1}{c|}{14.320} & 0.186   \\ \cline{2-12} 
                      & PCN \cite{yuan2018pcn} & \multicolumn{1}{c|}{0.133}   & 0.266  & 1.181 & \multicolumn{1}{c|}{1.366} & \multicolumn{1}{c|}{2.235}    & 2.235 & \multicolumn{1}{c|}{7.140}            & 8.650            & \multicolumn{1}{c|}{8.810}  & 0.315   \\ \cline{2-12} 
                      & AtlasNet\cite{AtlasNet2018papier} & \multicolumn{1}{c|}{-}       & -      & 1.085  & \multicolumn{1}{c|}{1.105} & \multicolumn{1}{c|}{-}        & -     & \multicolumn{1}{c|}{-}               & -               & \multicolumn{1}{c|}{-}     & -       \\ \hline
\multirow{2}{*}{2019}& TopNet \cite{tchapmi2019topnet}& \multicolumn{1}{c|}{0.126}   & 0.291  & 1.507 & \multicolumn{1}{c|}{0.606} & \multicolumn{1}{c|}{1.999}    & 5.354 & \multicolumn{1}{c|}{7.690}            & 10.190           & \multicolumn{1}{c|}{11.560} & 0.257   \\ \cline{2-12} 
                     & CRN \cite{zhang2019crn}& \multicolumn{1}{c|}{0.205}   & 0.201  & 1.947 & \multicolumn{1}{c|}{0.672} & \multicolumn{1}{c|}{2.267}    & 0.550  & \multicolumn{1}{c|}{5.460}            & 6.830            & \multicolumn{1}{c|}{9.210}  & 0.408   \\ \hline
\multirow{5}{*}{2020}& ECG \cite{pan2020ecg}& \multicolumn{1}{c|}{0.321}   & 0.176 & 1.028 & \multicolumn{1}{c|}{0.530}  & \multicolumn{1}{c|}{0.256}    & 0.812 & \multicolumn{1}{c|}{7.310}            & 7.060            & \multicolumn{1}{c|}{10.430} & 0.300     \\ \cline{2-12} 
                      & GRNet \cite{xie2020grnet}& \multicolumn{1}{c|}{0.238}   & 0.197  & 1.278 & \multicolumn{1}{c|}{0.579} & \multicolumn{1}{c|}{0.918}    & 0.816 & \multicolumn{1}{c|}{5.730}            & 7.610            & \multicolumn{1}{c|}{8.540}  & 0.314   \\ \cline{2-12} 
                      & MSN \cite{liu2020msn}& \multicolumn{1}{c|}{-}       & -      & -      & \multicolumn{1}{c|}{2.259} & \multicolumn{1}{c|}{-}        & 0.434 & \multicolumn{1}{c|}{5.370}            & 7.080            & \multicolumn{1}{c|}{8.880}  & 0.359   \\ \cline{2-12} 
                      & NSFA \cite{zhang2020nsfa}& \multicolumn{1}{c|}{-}       & -      & 0.806  & \multicolumn{1}{c|}{0.891} & \multicolumn{1}{c|}{-}        & -     & \multicolumn{1}{c|}{-}               & -               & \multicolumn{1}{c|}{-}     & -       \\ \cline{2-12} 
                      & PFNet \cite{huang2020pf}& \multicolumn{1}{c|}{0.339}   & 0.522  & -      & \multicolumn{1}{c|}{-}     & \multicolumn{1}{c|}{-}        & -     & \multicolumn{1}{c|}{-}               & -               & \multicolumn{1}{c|}{-}     & -       \\ \hline
\multirow{8}{*}{2021} & RFNet \cite{huang2021rfnet}& \multicolumn{1}{c|}{-}       & -      & 0.813  & \multicolumn{1}{c|}{-}     & \multicolumn{1}{c|}{-}        & -     & \multicolumn{1}{c|}{-}               & -               & \multicolumn{1}{c|}{-}     & -       \\ \cline{2-12} 
                      & VE-PCN \cite{wang2021vepcn}& \multicolumn{1}{c|}{-}       & -      & 0.832  & \multicolumn{1}{c|}{-}     & \multicolumn{1}{c|}{-}        & -     & \multicolumn{1}{c|}{-}               & -               & \multicolumn{1}{c|}{-}     & -       \\ \cline{2-12} 
                      & ASFMNet \cite{xia2021asfm}& \multicolumn{1}{c|}{0.247}   & 0.170 & 1.674 & \multicolumn{1}{c|}{0.869} & \multicolumn{1}{c|}{1.737}    & -     & \multicolumn{1}{c|}{-}               & -               & \multicolumn{1}{c|}{-}     & -       \\ \cline{2-12} 
                      & PMP-Net \cite{wen2021pmp}& \multicolumn{1}{c|}{-}       & -      & 0.866  & \multicolumn{1}{c|}{-}     & \multicolumn{1}{c|}{-}        & -     & \multicolumn{1}{c|}{-}               & -               & \multicolumn{1}{c|}{9.230}  & -       \\ \cline{2-12} 
                      & ASHF-Net \cite{zong2021ashf}& \multicolumn{1}{c|}{-}       & -      & -  & \multicolumn{1}{c|}{-}     & \multicolumn{1}{c|}{0.541}        & 0.773     & \multicolumn{1}{c|}{-}               & -               & \multicolumn{1}{c|}{-}  & -       \\ \cline{2-12}
                      & PoinTr \cite{yu2021pointr}& \multicolumn{1}{c|}{0.464}   & 0.124  & 0.883 & \multicolumn{1}{c|}{0.549} & \multicolumn{1}{c|}{-}        & -     & \multicolumn{1}{c|}{4.290}            & 6.240            & \multicolumn{1}{c|}{7.210}  & 0.402   \\ \cline{2-12} 
                      & VRCNet \cite{pan2021vrcnet}& \multicolumn{1}{c|}{-}       & -      & -      & \multicolumn{1}{c|}{0.536} & \multicolumn{1}{c|}{-}        & 0.540  & \multicolumn{1}{c|}{4.620}            & 5.820            & \multicolumn{1}{c|}{6.690}  & 0.433   \\ \cline{2-12} 
                      & PDR \cite{lyu2021pdr}& \multicolumn{1}{c|}{-}       & -      & -      & \multicolumn{1}{c|}{0.524} & \multicolumn{1}{c|}{-}        & 0.472 & \multicolumn{1}{c|}{4.260}            & 5.660            & \multicolumn{1}{c|}{7.100}   & 0.451   \\ \cline{2-12} 
                      & SnowFlakeNet \cite{xiang2021snowflakenet}& \multicolumn{1}{c|}{0.398}   & 0.124  & 0.719  & \multicolumn{1}{c|}{-}     & \multicolumn{1}{c|}{-}        & -     & \multicolumn{1}{c|}{4.770}            & 6.050            & \multicolumn{1}{c|}{-}     & -       \\ \hline
\multirow{3}{*}{2022} & PCUNet \cite{zhao2022pcunet}& \multicolumn{1}{c|}{-}       & -      & 0.839  & \multicolumn{1}{c|}{-}     & \multicolumn{1}{c|}{-}        & -     & \multicolumn{1}{c|}{-}               & -               & \multicolumn{1}{c|}{-}     & -       \\ \cline{2-12} 
                      & FBNet \cite{yan2022fbnet}& \multicolumn{1}{c|}{-}       & -      & 0.694  & \multicolumn{1}{c|}{2.970}  & \multicolumn{1}{c|}{-}        & 0.520  & \multicolumn{1}{c|}{-}               & -               & \multicolumn{1}{c|}{-}     & -       \\ \cline{2-12} 
                      & SeedFormer \cite{zhou2022seedformer}& \multicolumn{1}{c|}{0.472}   & 0.092  & 0.674  & \multicolumn{1}{c|}{0.516} & \multicolumn{1}{c|}{-}        & 0.151 & \multicolumn{1}{c|}{4.760}            & -               & \multicolumn{1}{c|}{6.980}  & 0.442   \\ \hline 
\multirow{7}{*}{2023} & PMP-Net++ \cite{wen2022pmp}& \multicolumn{1}{c|}{-}       & -      & 0.756      & \multicolumn{1}{c|}{-} & \multicolumn{1}{c|}{-}        & -     & \multicolumn{1}{c|}{-}            & -               & \multicolumn{1}{c|}{7.970}  & -   \\ \cline{2-12} 
                      & AnchorFormer \cite{chen2023anchorformer}& \multicolumn{1}{c|}{-}       & -      & -      & \multicolumn{1}{c|}{0.458} & \multicolumn{1}{c|}{-}        & -     & \multicolumn{1}{c|}{4.350}            & -               & \multicolumn{1}{c|}{6.850}  & 0.467   \\ \cline{2-12}   
                      & SVDFormer \cite{zhu2023svdformer}& \multicolumn{1}{c|}{-}       & -      & 0.654  & \multicolumn{1}{c|}{0.967} & \multicolumn{1}{c|}{-}        & -     & \multicolumn{1}{c|}{-}               & -               & \multicolumn{1}{c|}{-}     & -       \\ \cline{2-12} 
                      & DcTr \cite{fei2023dctr}& \multicolumn{1}{c|}{-}       & -      & 0.767 & \multicolumn{1}{c|}{0.547} & \multicolumn{1}{c|}{-}        & -     & \multicolumn{1}{c|}{-}               & -               & \multicolumn{1}{c|}{-}     & -       \\ \cline{2-12} 
                      & Cp3 \cite{xu2023cp3}& \multicolumn{1}{c|}{-}       & -      & -      & \multicolumn{1}{c|}{-}     & \multicolumn{1}{c|}{0.013}    & -     & \multicolumn{1}{c|}{3.490}            & 5.1             & \multicolumn{1}{c|}{-}     & -       \\ \cline{2-12} 
                      & AdaPoinTr \cite{yu2023adapointr}& \multicolumn{1}{c|}{0.503}   & 0.081  & 0.653  & \multicolumn{1}{c|}{0.392} & \multicolumn{1}{c|}{-}        & 0.237 & \multicolumn{1}{c|}{-}               & -               & \multicolumn{1}{c|}{-}     & -       \\ \cline{2-12} 
                      & ProxyFormer \cite{li2023proxyformer}& \multicolumn{1}{c|}{0.483}   & 0.093  & 0.677  & \multicolumn{1}{c|}{0.508} & \multicolumn{1}{c|}{-}        & -     & \multicolumn{1}{c|}{-}               & -               & \multicolumn{1}{c|}{-}     & -       \\ \cline{2-12} 
                      & NSDS \cite{cheng2023nsds}& \multicolumn{1}{c|}{-}       & -      & 0.521  & \multicolumn{1}{c|}{-}     & \multicolumn{1}{c|}{-}        & -     & \multicolumn{1}{c|}{-}               & -               & \multicolumn{1}{c|}{-}     & -       \\ \hline
\multirow{8}{*}{2024} & PointLDM \cite{chen2024learning}& \multicolumn{1}{c|}{-}       & -      & -      & \multicolumn{1}{c|}{0.485} & \multicolumn{1}{c|}{-}        & -     & \multicolumn{1}{c|}{3.920}            & 5.280            & \multicolumn{1}{c|}{6.830}  & 0.478   \\ \cline{2-12} 
                      & WalkFormer \cite{zhang2024walkformer}& \multicolumn{1}{c|}{-}       & -      & 0.679  & \multicolumn{1}{c|}{0.503} & \multicolumn{1}{c|}{-}        & 0.094 & \multicolumn{1}{c|}{-}               & -               & \multicolumn{1}{c|}{6.590}  & -       \\ \cline{2-12} 
                      & FSC \cite{wu2024fsc}& \multicolumn{1}{c|}{-}       & -      & -      & \multicolumn{1}{c|}{0.239} & \multicolumn{1}{c|}{-}        & -     & \multicolumn{1}{c|}{-}               & -               & \multicolumn{1}{c|}{-}     & -       \\ \cline{2-12} 
                      & ODGNet \cite{cai2024odgnet}& \multicolumn{1}{c|}{0.437}   & 0.083  & 0.650   & \multicolumn{1}{c|}{0.349} & \multicolumn{1}{c|}{-}        & 1.280  & \multicolumn{1}{c|}{-}               & -               & \multicolumn{1}{c|}{-}     & -       \\ \cline{2-12} 
                      & PCoT-Net \cite{leng2024pcotnet}& \multicolumn{1}{c|}{-}       & -      & -      & \multicolumn{1}{c|}{-}     & \multicolumn{1}{c|}{-}        & -     & \multicolumn{1}{c|}{-}               & 5.560            & \multicolumn{1}{c|}{-}     & -       \\ \cline{2-12} 
                      & SPOFormer \cite{fei2024spoformer}& \multicolumn{1}{c|}{0.346}   & 0.125 & 0.811  & \multicolumn{1}{c|}{0.725} & \multicolumn{1}{c|}{-}        & -     & \multicolumn{1}{c|}{-}               & -               & \multicolumn{1}{c|}{-}     & -       \\ \cline{2-12} 
                      & PointAttN \cite{wang2024pointattn}& \multicolumn{1}{c|}{-}       & -      & 0.686  & \multicolumn{1}{c|}{-}     & \multicolumn{1}{c|}{-}        & -     & \multicolumn{1}{c|}{-}               & -               & \multicolumn{1}{c|}{6.630}  & -       \\ \cline{2-12}
                      & RPPE-RA \cite{chen2024rppera}& \multicolumn{1}{c|}{0.498}   & 0.091  & 0.704  & \multicolumn{1}{c|}{0.506} & \multicolumn{1}{c|}{0.238}    & -     & \multicolumn{1}{c|}{-}               & 5.910            & \multicolumn{1}{c|}{6.850}  & -       \\ \hline
\end{tabular}
}
\vspace{-0.2cm}
\label{completionobjecttable}
\end{table*}

\begin{table}[t]
  \centering
  \caption{Code URL for object-level point cloud completion networks. The symbol `-' indicates that the code is unavailable.}
  \vspace{-0.2cm}
\small
\resizebox{\linewidth}{!}{
\begin{tabular}{|c|c|c|}
\hline
                       &                         &                                                                    \\
\multirow{-2}{*}{Year} & \multirow{-2}{*}{Model} & \multirow{-2}{*}{URL}                                              \\ \hline
                       & FoldingNet \cite{yang2018foldingnet}             &  https://github.com/qinglew/FoldingNet \\ \cline{2-3} 
                       & PCN \cite{yuan2018pcn}                    & https://github.com/wentaoyuan/pcn                                  \\ \cline{2-3} 
\multirow{-3}{*}{2018} & AtlasNet \cite{AtlasNet2018papier}               & https://github.com/ThibaultGROUEIX/AtlasNet\\ \hline
                        & TopNet \cite{tchapmi2019topnet}                 & https://github.com/lynetcha/completion3d                           \\ \cline{2-3} 
\multirow{-2}{*}{2019} & CRN \cite{zhang2019crn}                    & -                                                                                         \\ \hline 
                       & ECG \cite{pan2020ecg}                    & https://github.com/paul007pl/ECG                                   \\ \cline{2-3} 
                       & GRNet \cite{xie2020grnet}                  & https://github.com/hzxie/GRNet                                     \\ \cline{2-3} 
                       & MSN \cite{liu2020msn}                    & https://github.com/Colin97/MSN-Point-Cloud-Completion              \\ \cline{2-3} 
                       & NSFA \cite{zhang2020nsfa}                   & -                                                                  \\ \cline{2-3} 
\multirow{-5}{*}{2020} & PFNet \cite{huang2020pf}                  & -                                                                  \\ \hline
                       & RFNet \cite{huang2021rfnet}                  & https://github.com/Tianxinhuang/RFNet                              \\ \cline{2-3} 
                       & VE-PCN \cite{wang2021vepcn}                 & -                                                                  \\ \cline{2-3} 
                       & ASFMNet \cite{xia2021asfm}                & -                                                                  \\ \cline{2-3} 
                       & PMP-Net \cite{wen2021pmp}                & https:// github.com/ diviswen/ PMP-Net                                                                  \\ \cline{2-3}
                       & ASHF-Net \cite{zong2021ashf}                & -                                                                 \\ \cline{2-3} 
                       & PoinTr \cite{yu2021pointr}                 & https://github.com/yuxumin/PoinTr                                  \\ \cline{2-3} 
                       & VRCNet \cite{pan2021vrcnet}                 & -                                                                  \\ \cline{2-3} 
                       & PDR \cite{lyu2021pdr}                    & -                                                                  \\ \cline{2-3} 
\multirow{-9}{*}{2021} & SnowFlakeNet \cite{xiang2021snowflakenet}           & https://github.com/AllenXiangX/SnowflakeNet                        \\ \hline
                       & PCUNet \cite{zhao2022pcunet}                 & -                                                                  \\ \cline{2-3} 
                       & FBNet \cite{yan2022fbnet}                  & -                                                                  \\ \cline{2-3} 
\multirow{-3}{*}{2022} & SeedFormer \cite{zhou2022seedformer}             & -                                                    
                            \\ \hline
                        & PMP-Net++ \cite{wen2022pmp}           & https:// github.com/ diviswen/ PMP-Net                                                                 \\ \cline{2-3} 
                       & AnchorFormer \cite{chen2023anchorformer}           & -                                                                  \\ \cline{2-3} 
                       & SVDFormer \cite{zhu2023svdformer}              & https://github.com/czvvd/SVDFormer\_PointSea                       \\ \cline{2-3} 
                       & DcTr \cite{fei2023dctr}                   & https://github.com/Fayeben/DcTr                                    \\ \cline{2-3} 
                       & Cp3 \cite{xu2023cp3}                    & https://github.com/MingyeXu/cp3                                    \\ \cline{2-3} 
                       & AdaPoinTr \cite{yu2023adapointr}              & -                                                                  \\ \cline{2-3} 
                       & ProxyFormer \cite{li2023proxyformer}            & -                                                                  \\ \cline{2-3} 
\multirow{-7}{*}{2023} & NSDS \cite{cheng2023nsds}                   & -                                                                  \\ \hline
                       & PointLDM \cite{chen2024learning}               & -                                                                  \\ \cline{2-3} 
                       & WalkFormer \cite{zhang2024walkformer}             & -                                                                  \\ \cline{2-3} 
                       & FSC \cite{wu2024fsc}                    & -                                                                  \\ \cline{2-3} 
                       & ODGNet \cite{cai2024odgnet}                 & https://github.com/corecai163/ODGNet                               \\ \cline{2-3} 
                       & PCoT-Net \cite{leng2024pcotnet}               & -                                                                  \\ \cline{2-3} 
                       & SPOFormer \cite{fei2024spoformer}              & -                                                                  \\ \cline{2-3} 
                       & PointAttN \cite{wang2024pointattn}              & https://github.com/ohhhyeahhh/PointAttN
                       \\ \cline{2-3}
\multirow{-8}{*}{2024} & RPPE-RA  \cite{chen2024rppera}               & -                                                                  \\ \hline
\end{tabular}
}

  \label{denoisingurl}%
\end{table}%

\begin{table}[t]
  \centering
  \caption{Code URL for scene-level point cloud completion networks. The symbol `-' indicates that the code is unavailable.}
  \vspace{-0.2cm}
\scalebox{0.92}{
\begin{tabular}{|c|c|c|}
\hline
\multicolumn{1}{|c|}{\multirow{2}{*}{Year}} & \multirow{2}{*}{Method}  & \multirow{2}{*}{URL} \\
\multicolumn{1}{|c|}{}                      &                          &                      \\ \hline
2017                                        & SSCNet \cite{song2017semantic}     & https://github.com/shurans/sscnet                     \\ \hline
\multirow{3}{*}{2018}                       & CCDNet \cite{guedes2018semantic}       & -                     \\ \cline{2-3}
                                            & SATNet \cite{liu2018see}       & https://github.com/ShiceLiu/SATNet                     \\ \cline{2-3} 
                                            & ESSCNet \cite{zhang2018efficient}    & https://github.com/zjhthu/SGC-Release                     \\ \hline
\multirow{4}{*}{2019}                       & DDRNet \cite{li2019rgbd}        & https://github.com/waterljwant/SSC                     \\ \cline{2-3} 
                                            & TS3D+DNet \cite{behley2019semantickitti} & -                     \\ \cline{2-3} 
                                            & ForkNet \cite{wang2019forknet}     & -                     \\ \cline{2-3} 
                                            & CCPNet \cite{zhang2019crn}     & -                     \\ \hline
\multirow{4}{*}{2020}                       & 3DSketch \cite{chen20203d}    & https://github.com/charlesCXK/TorchSSC                     \\ \cline{2-3} 
                                            & IPF-SPCNet \cite{zhong2020semantic}     & -                     \\ \cline{2-3} 
                                            & AMFNet \cite{li2020attention}        & -                     \\ \cline{2-3} 
                                            & AICNet (C) \cite{li2020anisotropic}       & https://github.com/waterljwant/SSC                     \\ \hline
\multirow{3}{*}{2021}                       & EdgeNet \cite{dourado2021edgenet}       & -                    \\ \cline{2-3}
                                            & Igr-IS \cite{cai2021semantic}       & https://github.com/yjcaimeow/SISNet                     \\ \cline{2-3} 
                                            & AICNet++ (J) \cite{li2021anisotropic}       & https://waterljwant.github.io/SSC/                     \\ \cline{2-3} 
                                            & Local-DIFs \cite{rist2021semantic}  & -                     \\ \hline
\multirow{2}{*}{2022}                       & Local-Dis \cite{wang2022learning}   & https://github.com/wangyida/disp3d                     \\ \cline{2-3} 
                                            & FFNet \cite{wang2022ffnet}       & https://github.com/alanWXZ/FFNet                     \\ \hline
\multirow{6}{*}{2023}                       & PCANet \cite{li2023front}        & -                     \\ \cline{2-3} 
                                            & CVSformer \cite{dong2023cvsformer}   & https://github.com/donghaotian123/CVSformer                     \\ \cline{2-3} 
                                            & CLSNet \cite{wang2023semantic}      & https://github.com/fereenwong/CleanerS                     \\ \cline{2-3} 
                                            & AdaPoinTr \cite{yu2023adapointr}     & https://github.com/yuxumin/PoinTr                     \\ \cline{2-3} 
                                            & SCPNet \cite{xia2023scpnet}       & https://github.com/SCPNet/Codes-for-SCPNet                     \\ \cline{2-3} 
                                            & CasFusionNet \cite{xu2023casfusionnet}  & https://github.com/JinfengX/CasFusionNet                     \\ \hline
\multirow{2}{*}{2024}                       & AMMNet \cite{wang2024unleashing}      & https://github.com/fereenwong/AMMNet                     \\ \cline{2-3} 
                                            & SCSFNet \cite{wang2024semantic}     & https://scsfnet.github.io/                     \\ \hline
\end{tabular}
}
  \label{denoisingurl}%
\end{table}%

\begin{table*}[t]
      \centering
      \caption{Comparative results of point cloud completion methods for scenes on NYUv2 $(60\times36\times60)$, NYUCAD $(60\times36\times60)$, SUNCG $(60\times36\times60)$, and SementicKITTI $(256\times32\times256)$ datasets. The values in parentheses are commonly used resolutions. The symbol `-' means the result is unavailable. `$a$' means results with SUNCG pretraning. `$b$' means results at different resolutions, whereas CCPNet $(240\times144\times240)$ and ForkNet $(80\times48\times80)$.`$c$' means results on SUNCG-RGBD, a miniature version of SUNCG.}
      \vspace{-0.2cm}
      \resizebox*{\textwidth}{!}{
\begin{tabular}{|l|c|ccc|cccc|cccc|cccc|cccc|}
\hline
\multicolumn{1}{|c|}{}                       &                          & \multicolumn{3}{c|}{Input Encoding}                                                   & \multicolumn{4}{c|}{NYUv2}                                                                                                                                                             & \multicolumn{4}{c|}{NYUCAD}                                                                                                                                                           & \multicolumn{4}{c|}{SUNCG}                                                                                                                                                             & \multicolumn{4}{c|}{SemanticKITTI}                                                                                                                                                   \\ \cline{3-21} 
\multicolumn{1}{|c|}{\multirow{-2}{*}{Year}} & \multirow{-2}{*}{Method} & \multicolumn{1}{l|}{Point} & \multicolumn{1}{l|}{Volume} & \multicolumn{1}{l|}{RGB-D} & \multicolumn{1}{c|}{prec.}                        & \multicolumn{1}{c|}{recall}                      & \multicolumn{1}{c|}{IoU}                          & mIoU                        & \multicolumn{1}{c|}{prec.}                        & \multicolumn{1}{c|}{recall}                      & \multicolumn{1}{c|}{IoU}                         & mIoU                        & \multicolumn{1}{c|}{prec.}                        & \multicolumn{1}{c|}{recall}                      & \multicolumn{1}{c|}{IoU}                          & mIoU                         & \multicolumn{1}{c|}{prec.}                       & \multicolumn{1}{c|}{recall}                      & \multicolumn{1}{c|}{IoU}                         & mIoU                        \\ \hline
2017                                         & SSCNet \cite{song2017semantic}      & \multicolumn{1}{c|}{}      & \multicolumn{1}{c|}{\checkmark}      &                            & \multicolumn{1}{c|}{{57.0}}  & \multicolumn{1}{c|}{{94.5}} & \multicolumn{1}{c|}{{55.1}}  & {24.7} & \multicolumn{1}{c|}{{75.4}}  & \multicolumn{1}{c|}{{96.3}} & \multicolumn{1}{c|}{{73.2}} & {40.0} & \multicolumn{1}{c|}{{76.3}}  & \multicolumn{1}{c|}{{95.2}} & \multicolumn{1}{c|}{{73.5}}  & {46.4}  & \multicolumn{1}{c|}{{31.7}} & \multicolumn{1}{c|}{{83.4}} & \multicolumn{1}{c|}{{29.8}} & {9.5}  \\ \hline
                                             & CCDNet \cite{guedes2018semantic}       & \multicolumn{1}{c|}{}      & \multicolumn{1}{c|}{}       & \checkmark                          & \multicolumn{1}{c|}{{62.5}} & \multicolumn{1}{c|}{{82.3}}  & \multicolumn{1}{c|}{{54.3}}  & 27.5  & \multicolumn{1}{c|}{{-}}     & \multicolumn{1}{c|}{{-}}    & \multicolumn{1}{c|}{{-}}    & {-}    & \multicolumn{1}{c|}{-} & \multicolumn{1}{c|}{-} & \multicolumn{1}{c|}{-}  & {-}  & \multicolumn{1}{c|}{{-}}    & \multicolumn{1}{c|}{{-}}    & \multicolumn{1}{c|}{{-}}    & {-}    \\ \cline{2-21} 
                                             & SATNet \cite{liu2018see}       & \multicolumn{1}{c|}{}      & \multicolumn{1}{c|}{}       & \checkmark                          & \multicolumn{1}{c|}{{67.3}$^{a}$} & \multicolumn{1}{c|}{{85.8}$^{a}$ } & \multicolumn{1}{c|}{{60.6}$^{a}$}  & {34.4}$^{a}$  & \multicolumn{1}{c|}{{-}}     & \multicolumn{1}{c|}{{-}}    & \multicolumn{1}{c|}{{-}}    & {-}    & \multicolumn{1}{c|}{{80.7}} & \multicolumn{1}{c|}{{96.5}} & \multicolumn{1}{c|}{{78.5}}  & {64.3}  & \multicolumn{1}{c|}{{-}}    & \multicolumn{1}{c|}{{-}}    & \multicolumn{1}{c|}{{-}}    & {-}    \\ \cline{2-21} 
\multirow{-3}{*}{2018}                       & ESSCNet \cite{zhang2018efficient}    & \multicolumn{1}{c|}{}      & \multicolumn{1}{c|}{\checkmark}      &                            & \multicolumn{1}{c|}{{71.9}}  & \multicolumn{1}{c|}{{71.9}} & \multicolumn{1}{c|}{{56.2}}  & {26.7} & \multicolumn{1}{c|}{{-}}     & \multicolumn{1}{c|}{{-}}    & \multicolumn{1}{c|}{{-}}    & {-}    & \multicolumn{1}{c|}{{92.6}}  & \multicolumn{1}{c|}{{90.4}} & \multicolumn{1}{c|}{{84.5}}  & {70.5}  & \multicolumn{1}{c|}{{62.6}} & \multicolumn{1}{c|}{{55.6}} & \multicolumn{1}{c|}{{41.8}} & {17.5} \\ \hline
                                             & DDRNet \cite{li2019rgbd}        & \multicolumn{1}{c|}{}      & \multicolumn{1}{c|}{}       & \checkmark                          & \multicolumn{1}{c|}{{71.5}}  & \multicolumn{1}{c|}{{80.8}} & \multicolumn{1}{c|}{{64.0}}  & {30.4} & \multicolumn{1}{c|}{{88.7}}  & \multicolumn{1}{c|}{{88.5}} & \multicolumn{1}{c|}{{79.4}} & {42.8} & \multicolumn{1}{c|}{{-}}     & \multicolumn{1}{c|}{{-}}    & \multicolumn{1}{c|}{{-}}     & {-}     & \multicolumn{1}{c|}{{-}}    & \multicolumn{1}{c|}{{-}}    & \multicolumn{1}{c|}{{-}}    & {-}    \\ \cline{2-21} 
                                             & TS3D+DNet \cite{behley2019semantickitti} & \multicolumn{1}{c|}{}      & \multicolumn{1}{c|}{\checkmark}      &                            & \multicolumn{1}{c|}{{-}}     & \multicolumn{1}{c|}{{-}}    & \multicolumn{1}{c|}{{-}}     & {-}    & \multicolumn{1}{c|}{{-}}     & \multicolumn{1}{c|}{{-}}    & \multicolumn{1}{c|}{{-}}    & {-}    & \multicolumn{1}{c|}{{-}}     & \multicolumn{1}{c|}{{-}}    & \multicolumn{1}{c|}{{-}}     & {-}     & \multicolumn{1}{c|}{{25.9}} & \multicolumn{1}{c|}{{88.3}} & \multicolumn{1}{c|}{{25.0}} & {10.2} \\ \cline{2-21} 
                                             & ForkNet \cite{wang2019forknet}     & \multicolumn{1}{c|}{}      & \multicolumn{1}{c|}{\checkmark}      &                            & \multicolumn{1}{c|}{{-}}     & \multicolumn{1}{c|}{{-}}    & \multicolumn{1}{c|}{{63.4}$^{ab}$} & {37.1}$^{ab}$ & \multicolumn{1}{c|}{{-}}     & \multicolumn{1}{c|}{{-}}    & \multicolumn{1}{c|}{{-}}    & {-}    & \multicolumn{1}{c|}{{-}}     & \multicolumn{1}{c|}{{-}}    & \multicolumn{1}{c|}{{86.9}$^{b}$} & {63.4}$^{b}$  & \multicolumn{1}{c|}{{-}}    & \multicolumn{1}{c|}{{-}}    & \multicolumn{1}{c|}{{-}}    & {-}    \\ \cline{2-21} 
\multirow{-4}{*}{2019}                       & CCPNet \cite{zhang2019crn}     & \multicolumn{1}{c|}{}      & \multicolumn{1}{c|}{\checkmark}      &                            & \multicolumn{1}{c|}{{74.2}$^{ab}$} & \multicolumn{1}{c|}{{90.8}$^{ab}$} & \multicolumn{1}{c|}{{63.5}$^{ab}$}  & {38.5}$^{ab}$ & \multicolumn{1}{c|}{{91.3}$^{ab}$} & \multicolumn{1}{c|}{{92.6}$^{ab}$} & \multicolumn{1}{c|}{{82.4}$^{ab}$} & {53.2}$^{ab}$ & \multicolumn{1}{c|}{{98.2}$^{b}$} & \multicolumn{1}{c|}{{96.8}$^{b}$} & \multicolumn{1}{c|}{{91.4}$^{b}$}  & {74.2}$^{b}$  & \multicolumn{1}{c|}{{-}}    & \multicolumn{1}{c|}{{-}}    & \multicolumn{1}{c|}{{-}}    & {-}    \\ \hline                            
                                             & 3DSketch \cite{chen20203d}    & \multicolumn{1}{c|}{}      & \multicolumn{1}{c|}{\checkmark}      &                            & \multicolumn{1}{c|}{{85.0}}  & \multicolumn{1}{c|}{{81.6}} & \multicolumn{1}{c|}{{71.3}}  & {41.1} & \multicolumn{1}{c|}{{90.6}}  & \multicolumn{1}{c|}{{92.2}} & \multicolumn{1}{c|}{{84.2}} & {55.2} & \multicolumn{1}{c|}{{-}}     & \multicolumn{1}{c|}{{-}}    & \multicolumn{1}{c|}{{88.2}} & {76.5} & \multicolumn{1}{c|}{{-}}    & \multicolumn{1}{c|}{{-}}    & \multicolumn{1}{c|}{{-}}    & {-}    \\ \cline{2-21} 
                                             & IPF-SPCNet \cite{zhong2020semantic}     & \multicolumn{1}{c|}{\checkmark}     & \multicolumn{1}{c|}{}       &                            & \multicolumn{1}{c|}{70.5}                         & \multicolumn{1}{c|}{46.7}                        & \multicolumn{1}{c|}{39.0}                         & 35.1                        & \multicolumn{1}{c|}{83.3}                         & \multicolumn{1}{c|}{72.7}                        & \multicolumn{1}{c|}{63.5}                        & 50.7                        & \multicolumn{1}{c|}{-}                            & \multicolumn{1}{c|}{-}                           & \multicolumn{1}{c|}{-}                            & -                            & \multicolumn{1}{c|}{-}                           & \multicolumn{1}{c|}{-}                           & \multicolumn{1}{c|}{-}                           & -                           \\ \cline{2-21} 
                                             & AMFNet \cite{li2020attention}        & \multicolumn{1}{c|}{}      & \multicolumn{1}{c|}{}       & \checkmark                          & \multicolumn{1}{c|}{67.9}                         & \multicolumn{1}{c|}{{82.3}} & \multicolumn{1}{c|}{{59.0}}  & {33.0} & \multicolumn{1}{c|}{{-}}     & \multicolumn{1}{c|}{{-}}    & \multicolumn{1}{c|}{{-}}    & {-}    & \multicolumn{1}{c|}{{-}}     & \multicolumn{1}{c|}{{-}}    & \multicolumn{1}{c|}{{-}}     & {-}     & \multicolumn{1}{c|}{{-}}    & \multicolumn{1}{c|}{{-}}    & \multicolumn{1}{c|}{-}                           & -                           \\ \cline{2-21} 
\multirow{-4}{*}{2020}                       & AICNet (C) \cite{li2020anisotropic}        & \multicolumn{1}{c|}{}      & \multicolumn{1}{c|}{}       & \checkmark                          & \multicolumn{1}{c|}{62.4}                         & \multicolumn{1}{c|}{{91.8}} & \multicolumn{1}{c|}{{59.2}}  & {33.3} & \multicolumn{1}{c|}{{88.2}}  & \multicolumn{1}{c|}{{90.3}} & \multicolumn{1}{c|}{{80.5}} & {45.8} & \multicolumn{1}{c|}{{-}}     & \multicolumn{1}{c|}{{-}}    & \multicolumn{1}{c|}{{-}}     & {-}     & \multicolumn{1}{c|}{{-}}    & \multicolumn{1}{c|}{{-}}    & \multicolumn{1}{c|}{-}                           & -                           \\ \hline
                                             & EdgeNet \cite{dourado2021edgenet}       & \multicolumn{1}{c|}{}      & \multicolumn{1}{c|}{}      & \checkmark                           & \multicolumn{1}{c|}{76.0}                         & \multicolumn{1}{c|}{{68.3}} & \multicolumn{1}{c|}{{56.1}}  & {27.8} & \multicolumn{1}{c|}{-}  & \multicolumn{1}{c|}{-} & \multicolumn{1}{c|}{-} & {-} & \multicolumn{1}{c|}{{93.1}} & \multicolumn{1}{c|}{{90.4}} & \multicolumn{1}{c|}{{84.8}}  & {69.5}  & \multicolumn{1}{c|}{{-}}    & \multicolumn{1}{c|}{{-}}    & \multicolumn{1}{c|}{-}                           & -                           \\ \cline{2-21}                                              
                                             & Igr-IS \cite{cai2021semantic}       & \multicolumn{1}{c|}{}      & \multicolumn{1}{c|}{\checkmark}      &                            & \multicolumn{1}{c|}{92.1}                         & \multicolumn{1}{c|}{{83.8}} & \multicolumn{1}{c|}{{78.2}}  & {52.4} & \multicolumn{1}{c|}{{94.1}}  & \multicolumn{1}{c|}{{91.2}} & \multicolumn{1}{c|}{{86.3}} & {63.5} & \multicolumn{1}{c|}{{92.6}$^{c}$} & \multicolumn{1}{c|}{{96.3}$^{c}$} & \multicolumn{1}{c|}{{89.3}$^{c}$}  & {79.0}$^{c}$  & \multicolumn{1}{c|}{{-}}    & \multicolumn{1}{c|}{{-}}    & \multicolumn{1}{c|}{-}                           & -                           \\ \cline{2-21} 
                                             & AICNet (J) \cite{li2021anisotropic}      & \multicolumn{1}{c|}{}      & \multicolumn{1}{c|}{}       & \checkmark                          & \multicolumn{1}{c|}{69.8}                         & \multicolumn{1}{c|}{{83.2}} & \multicolumn{1}{c|}{{61.2}}  & {34.7} & \multicolumn{1}{c|}{{88.7}}  & \multicolumn{1}{c|}{{90.3}} & \multicolumn{1}{c|}{{80.7}} & {47.0} & \multicolumn{1}{c|}{{-}}     & \multicolumn{1}{c|}{{-}}    & \multicolumn{1}{c|}{{-}}     & {-}     & \multicolumn{1}{c|}{{-}}    & \multicolumn{1}{c|}{{-}}    & \multicolumn{1}{c|}{-}                           & -                           \\ \cline{2-21} 
\multirow{-4}{*}{2021}                       & Local-DIFs \cite{rist2021semantic}  & \multicolumn{1}{c|}{\checkmark}     & \multicolumn{1}{c|}{}       &                            & \multicolumn{1}{c|}{-}                            & \multicolumn{1}{c|}{{-}}    & \multicolumn{1}{c|}{{-}}     & {-}    & \multicolumn{1}{c|}{{-}}     & \multicolumn{1}{c|}{{-}}    & \multicolumn{1}{c|}{{-}}    & {-}    & \multicolumn{1}{c|}{{-}}     & \multicolumn{1}{c|}{{-}}    & \multicolumn{1}{c|}{{-}}     & {-}     & \multicolumn{1}{c|}{{-}}    & \multicolumn{1}{c|}{{-}}    & \multicolumn{1}{c|}{57.7}                        & 22.7                        \\ \hline
                                             & Local-Dis \cite{wang2022learning}   & \multicolumn{1}{c|}{\checkmark}     & \multicolumn{1}{c|}{}       &                            & \multicolumn{1}{c|}{-}                            & \multicolumn{1}{c|}{{-}}    & \multicolumn{1}{c|}{{-}}     & {42.4} & \multicolumn{1}{c|}{{-}}     & \multicolumn{1}{c|}{{-}}    & \multicolumn{1}{c|}{{-}}    & {-}    & \multicolumn{1}{c|}{{-}}     & \multicolumn{1}{c|}{{-}}    & \multicolumn{1}{c|}{{-}}     & {-}     & \multicolumn{1}{c|}{{-}}    & \multicolumn{1}{c|}{{-}}    & \multicolumn{1}{c|}{-}                           & -                           \\ \cline{2-21} 
\multirow{-2}{*}{2022}                       & FFNet \cite{wang2022ffnet}      & \multicolumn{1}{c|}{}      & \multicolumn{1}{c|}{}       & \checkmark                          & \multicolumn{1}{c|}{89.3}                         & \multicolumn{1}{c|}{{78.5}} & \multicolumn{1}{c|}{{71.8}}  & {44.4} & \multicolumn{1}{c|}{{94.8}}  & \multicolumn{1}{c|}{{90.3}} & \multicolumn{1}{c|}{{85.5}} & {57.4} & \multicolumn{1}{c|}{{-}}     & \multicolumn{1}{c|}{{-}}    & \multicolumn{1}{c|}{{-}}     & {-}     & \multicolumn{1}{c|}{{-}}    & \multicolumn{1}{c|}{{-}}    & \multicolumn{1}{c|}{-}                           & -                           \\ \hline
                                             & PCANet \cite{li2023front}        & \multicolumn{1}{c|}{}      & \multicolumn{1}{c|}{}       & \checkmark                          & \multicolumn{1}{c|}{89.5}                         & \multicolumn{1}{c|}{{87.5}} & \multicolumn{1}{c|}{{78.9}}  & {48.9} & \multicolumn{1}{c|}{{92.1}}  & \multicolumn{1}{c|}{{91.8}} & \multicolumn{1}{c|}{{84.3}} & {59.6} & \multicolumn{1}{c|}{{92.2}$^{c}$} & \multicolumn{1}{c|}{{90.2}$^{c}$} & \multicolumn{1}{c|}{{83.9}$^{c}$}  & {65.9}$^{c}$  & \multicolumn{1}{c|}{{-}}    & \multicolumn{1}{c|}{{-}}    & \multicolumn{1}{c|}{-}                           & -                           \\ \cline{2-21} 
                                             & CVSformer \cite{dong2023cvsformer}   & \multicolumn{1}{c|}{}      & \multicolumn{1}{c|}{\checkmark}      &                            & \multicolumn{1}{c|}{97.7}                         & \multicolumn{1}{c|}{{82.1}} & \multicolumn{1}{c|}{{73.7}}  & {52.6} & \multicolumn{1}{c|}{{94.0}}  & \multicolumn{1}{c|}{{91.0}} & \multicolumn{1}{c|}{{86.0}} & {63.9} & \multicolumn{1}{c|}{{-}}     & \multicolumn{1}{c|}{{-}}    & \multicolumn{1}{c|}{{-}}     & {-}     & \multicolumn{1}{c|}{{-}}    & \multicolumn{1}{c|}{{-}}    & \multicolumn{1}{c|}{-}                           & -                           \\ \cline{2-21} 
                                             & CLSNet \cite{wang2023semantic}      & \multicolumn{1}{c|}{}      & \multicolumn{1}{c|}{\checkmark}      &                            & \multicolumn{1}{c|}{88.0}                         & \multicolumn{1}{c|}{{83.5}} & \multicolumn{1}{c|}{{75.0}}  & {47.7} & \multicolumn{1}{c|}{{-}}     & \multicolumn{1}{c|}{{-}}    & \multicolumn{1}{c|}{{-}}    & {-}    & \multicolumn{1}{c|}{{-}}     & \multicolumn{1}{c|}{{-}}    & \multicolumn{1}{c|}{{-}}     & {-}     & \multicolumn{1}{c|}{{-}}    & \multicolumn{1}{c|}{{-}}    & \multicolumn{1}{c|}{-}                           & -                           \\ \cline{2-21} 
                                             & AdaPoinTr \cite{yu2023adapointr}     & \multicolumn{1}{c|}{\checkmark}     & \multicolumn{1}{c|}{}       &                            & \multicolumn{1}{c|}{90.1}                         & \multicolumn{1}{c|}{82.9}                        & \multicolumn{1}{c|}{74.6}                         & 44.1                        & \multicolumn{1}{c|}{94.4}                         & \multicolumn{1}{c|}{90.6}                        & \multicolumn{1}{c|}{86.1}                        & 56.1                        & \multicolumn{1}{c|}{-}                            & \multicolumn{1}{c|}{-}                           & \multicolumn{1}{c|}{-}                            & -                            & \multicolumn{1}{c|}{-}                           & \multicolumn{1}{c|}{-}                           & \multicolumn{1}{c|}{-}                           & -                           \\ \cline{2-21} 
                                             & SCPNet \cite{xia2023scpnet}       & \multicolumn{1}{c|}{\checkmark}     & \multicolumn{1}{c|}{}       &                            & \multicolumn{1}{c|}{-}                            & \multicolumn{1}{c|}{-}                           & \multicolumn{1}{c|}{-}                            & -                           & \multicolumn{1}{c|}{-}                            & \multicolumn{1}{c|}{-}                           & \multicolumn{1}{c|}{-}                           & -                           & \multicolumn{1}{c|}{-}                            & \multicolumn{1}{c|}{-}                           & \multicolumn{1}{c|}{-}                            & -                            & \multicolumn{1}{c|}{-}                           & \multicolumn{1}{c|}{-}                           & \multicolumn{1}{c|}{-}                           & 36.7                        \\ \cline{2-21} 
\multirow{-6}{*}{2023}                       & CasFusionNet \cite{xu2023casfusionnet}  & \multicolumn{1}{c|}{\checkmark}     & \multicolumn{1}{c|}{}       &                            & \multicolumn{1}{c|}{-}                            & \multicolumn{1}{c|}{-}                           & \multicolumn{1}{c|}{-}                            & -                           & \multicolumn{1}{c|}{-}                            & \multicolumn{1}{c|}{-}                           & \multicolumn{1}{c|}{-}                           & 49.5                        & \multicolumn{1}{c|}{-}                            & \multicolumn{1}{c|}{-}                           & \multicolumn{1}{c|}{-}                            & -                            & \multicolumn{1}{c|}{-}                           & \multicolumn{1}{c|}{-}                           & \multicolumn{1}{c|}{-}                           & -                           \\ \hline
                                             & AMMNet \cite{wang2024unleashing}     & \multicolumn{1}{c|}{}      & \multicolumn{1}{c|}{\checkmark}      &                            & \multicolumn{1}{c|}{88.7}                         & \multicolumn{1}{c|}{84.5}                        & \multicolumn{1}{c|}{76.3}                         & 56.1                        & \multicolumn{1}{c|}{92.8}                         & \multicolumn{1}{c|}{89.0}                        & \multicolumn{1}{c|}{83.3}                        & 67.2                        & \multicolumn{1}{c|}{-}                            & \multicolumn{1}{c|}{-}                           & \multicolumn{1}{c|}{-}                            & -                            & \multicolumn{1}{c|}{-}                           & \multicolumn{1}{c|}{-}                           & \multicolumn{1}{c|}{-}                           & -                           \\ \cline{2-21} 
\multirow{-2}{*}{2024}                       & SCSFNet \cite{wang2024semantic}     & \multicolumn{1}{c|}{\checkmark}     & \multicolumn{1}{c|}{}       &                            & \multicolumn{1}{c|}{-}                            & \multicolumn{1}{c|}{-}                           & \multicolumn{1}{c|}{-}                            & -                           & \multicolumn{1}{c|}{-}                            & \multicolumn{1}{c|}{-}                           & \multicolumn{1}{c|}{-}                           & -                           & \multicolumn{1}{c|}{-}                            & \multicolumn{1}{c|}{-}                           & \multicolumn{1}{c|}{-}                            & -                            & \multicolumn{1}{c|}{-}                           & \multicolumn{1}{c|}{-}                           & \multicolumn{1}{c|}{34.5}                        & 16.1                        \\ \hline
\end{tabular}
}
\vspace{-0.2cm}
      \label{completionscenetable}
\end{table*}

\subsection{Object-level Completion}
The goal of object completion is to predict the missing shape of an object. We divide object-level completion into two categories: point-based and voxel/projection-based.

\begin{figure}[t]
    \centering
    \includegraphics[width=\linewidth]
    {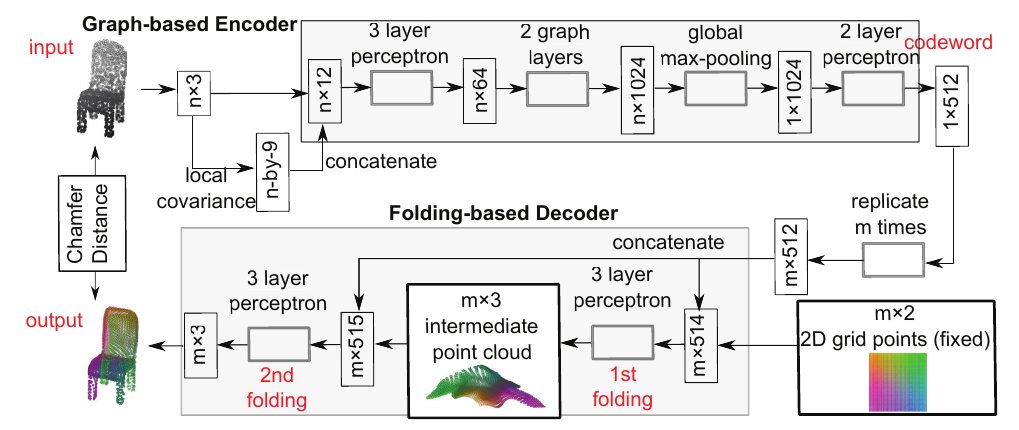}
    \caption{Illustration of FoldingNet \cite{yang2018foldingnet} architecture.}
    \label{FoldingNet}
\end{figure} 
\begin{figure}[t]
    \centering
    \includegraphics[width=\linewidth]
    {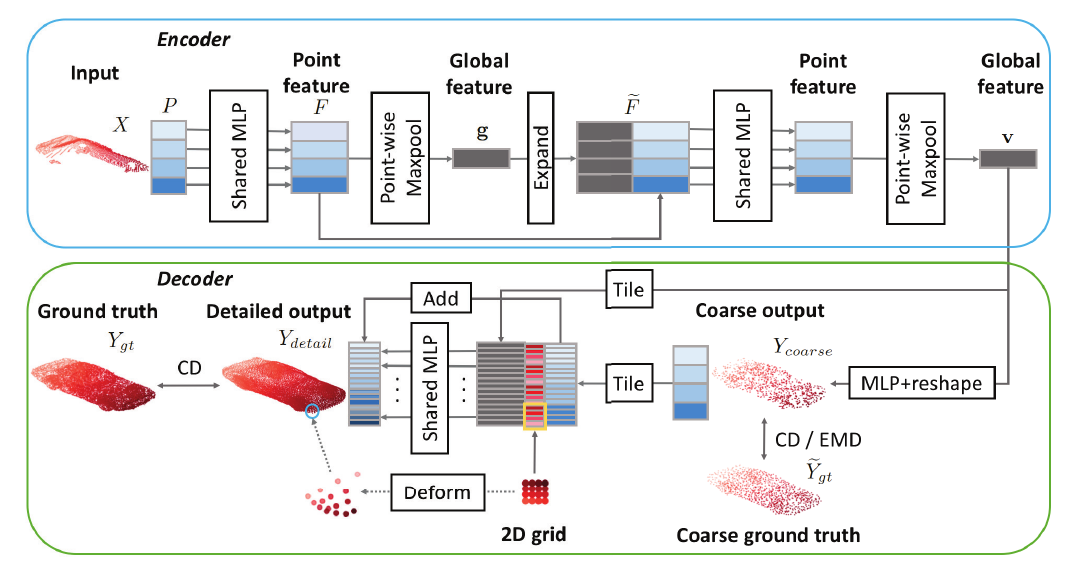}
    \caption{Illustration of PCN \cite{yuan2018pcn} architecture.}
    \label{PCN}
\end{figure}

\begin{figure*}[t]
    \centering
    \includegraphics[width=\textwidth]{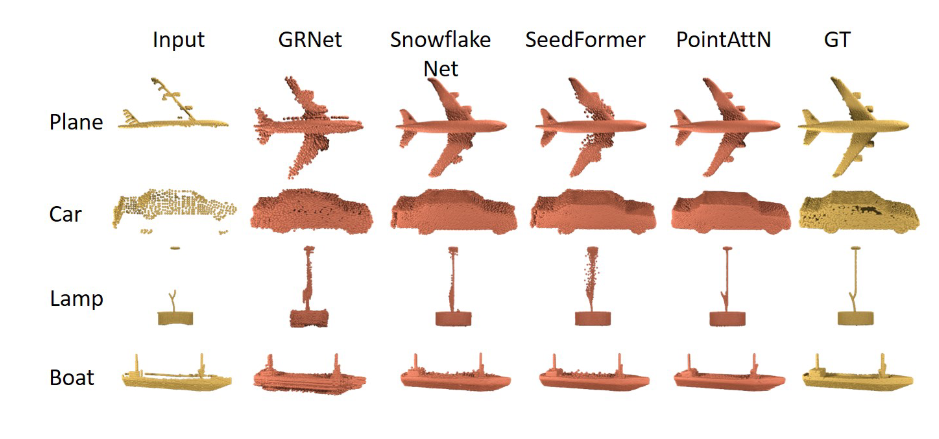}
    \caption{Qualitative comparisons on PCN \cite{yuan2018pcn} with state-of-the-art methods, including GRNet \cite{xie2020grnet}, SnowFlakeNet \cite{xiang2021snowflakenet}, SeedFormer \cite{zhou2022seedformer} and PointAttn \cite{wang2024pointattn}. These results are derived from PointAttn \cite{wang2024pointattn}.}
    \label{objectvisual}
\end{figure*}

\subsubsection{\textbf{Point-based}}
These techniques extract point feature characteristics from the input point cloud and directly generate the final complete point cloud. We divide point-based object completion into four categories: one-stage, meta-point-based, deformation-based, and the others.
\paragraph{\textbf{One-stage}} These methods directly utilize characteristics derived from the inputs to generate the entire point cloud, without generating any intermediate auxiliary point clouds.

 Groueix et al.~\cite{AtlasNet2018papier} proposed AtlasNet, which represents 3D shapes as collections of parametric surface elements. AtlasNet employs learnable transformations to map a set of 2D squares onto a surface, similar to placing paper strips onto a form to create a papier-mâché model. These transformation parameters are derived from both the neural network's learned weights and a representation of the shape. Tchapmi et al. \cite{tchapmi2019topnet} proposed TopNet, which integrates the encoders from PointNet \cite{qi2017pointnet} with the suggested decoders. The decoder exhibits high flexibility and can learn any arbitrary grouping of points, regardless of the topology of the point set. Subsequently, Wang et al. \cite{wang2020softpoolnet} presented SoftPoolNet, which replaces max-pooling to retain multiple high-scoring features, preserving more information while maintaining permutation invariance. Additionally, regional convolutions are introduced for the decoding stage to enhance global activation entropy. Inspired by the local refining procedure in PCN \cite{yuan2018pcn}, a patch-deforming operation is also proposed to simulate deconvolutional operations.
This improves the preservation of details in 3D reconstruction. Afterwards, P$^2$GNet \cite{zhang2022partial} utilizes a locality-aware feature-disentangled encoder to effectively incorporate incomplete point cloud data into a global feature representation while also handling missing data hierarchically. Ma et al. \cite{ma2023symmetric} presented a method that prioritizes the preservation of symmetry and detail in real scene objects. The precision and homogeneity of the reconstructed shapes are improved by utilizing a symmetry learning module and a refinement auto-encoder. Chen et al. \cite{chen2024learning} introduced PointLDM. This model utilizes a transformer-based VAE to encode shape latents. It then employs a diffusion process to achieve accurate latent formulation. Inspired by  \cite{li2022contextual}, Leng et al. \cite{leng2024pcotnet} proposed a unique point contextual transformer, which combines static and dynamic local characteristics using attention mechanisms.

\paragraph{\textbf{Meta-point-based}} Meta-points represent the intermediate data generated during feature learning. These methods extract global geometric features from the input to generate approximately complete point clouds, i.e., meta-points. After that, they refine the meta-points using local geometry information to produce detailed point cloud completion results. It can be divided into two categories: whole-shape-aware and missing-region-aware.

\textbf{Whole-shape-aware.} These methods  \cite{yuan2018pcn, zhao2022pcunet,wang2022softpool++,zhou2022seedformer, zhang2022point} generate meta-points encompassing the whole object for completion.

FoldingNet \cite{yang2018foldingnet} utilizes a graph-based encoder and an innovative folding-based decoding mechanism that maps a 2D grid onto the 3D surface of the point cloud. This approach guarantees minimal reconstruction errors and offers a reliable technique for unsupervised feature learning in 3D geometric data, as shown in Fig. \ref{FoldingNet}. PCN \cite{yuan2018pcn} introduces a decoder that produces detailed completions while keeping the parameter count low, as shown in Fig. \ref{PCN}. Inspired by FoldingNet \cite{yang2018foldingnet}, Xia et al. \cite{xia2021asfm} introduced ASFM-Net, which utilizes PCN as the underlying network architecture for the auto-encoder. This network can generate a more comprehensive global feature for items that are not fully defined. Zhang et al. \cite{zhang2022point} introduced the skeleton-detail transformer. It improves the accuracy of preserving geometric details compared to earlier approaches, boosting the quality of point cloud reconstruction. 

Utilizing PointNet++ \cite{qi2017pointnet++} as a primary encoder, LAKe-Net \cite{tang2022lake} extracts local and global features simultaneously. The asymmetric keypoint locator incorporates an unsupervised multi-scale keypoint detector and a comprehensive keypoint generator. This system can accurately identify keypoints for both whole and partial objects across many categories. ASHF-Net \cite{zong2021ashf} uses a denoising auto-encoder with adaptive sampling to learn robust local features and a hierarchical folding decoder with gated skip-attention for detailed structure reconstruction. The model also employs a KL regularization term to ensure uniform point distribution. Xiang et al. \cite{xiang2021snowflakenet} introduced SnowflakeNet, which uses snowflake point deconvolution to capture local geometries for comprehensive point cloud completion precisely. This method exhibits versatility and surpasses current strategies in many point cloud tasks. Chen et al. \cite{chen2023anchorformer} included an attention block in the previous approaches, generating a dual attention block. The system utilizes pattern-aware discriminative nodes, known as anchors, to actively capture regional information of objects. Unlike previous single-channel transfomer-based methods, DcTr \cite{fei2023dctr} introduces a dual-channel transformer. It combines a dual-channel transformer with cross-attention techniques to enhance the robustness of point cloud completion to noise. Subsequently, Cui et al. \cite{cui2023p2c} proposed Partial2Complete (P2C), which uses only a single partial observation to avoid using entire samples or many observations. 

To eliminate local region operations such as $k$-nearest neighbours ($k$NN) in transformer-based approaches, Wang et al. presented PointAttN \cite{wang2024pointattn}. It utilizes cross-attention and self-attention processes along with implicit local region division. They also proposed two key components, Geometric Details Perception and Self-Feature Augment, to enhance feature extraction and completion. To tackle the problem of shape completion with extremely sparse point clouds, Wu et al. \cite{wu2024fsc} introduced the few-point shape completion (FSC) model. It incorporates a revolutionary dual-branch feature extractor designed to handle inputs with very low density effectively. It also includes an extensive branch that maximizes the exploitation of data points. 
Afterwards, ODGNet \cite{cai2024odgnet} enhances previous approaches by using a unique seed generation U-Net that utilizes multi-level feature extraction to improve the representation of seed points with more details. Motivated by prompting techniques in natural language processing, CP3 \cite{xu2023cp3} uses pretrain-prompt-predict for point cloud completion, which improves the reliability and precision of producing and revising point clouds. SVDFormer \cite{zhu2023svdformer} is proposed to analyze multiple views and a dual-generator technique to refine the structure.

\textbf{Missing-region-aware.} In contrary to whole-shape-aware methods, missing-region-aware methods \cite{huang2020pf, xiao2022df,yu2023adapointr,li2023proxyformer} generate meta-points that specifically address the missing region of the objects. 

PF-Net \cite{huang2020pf} presents an innovative method for accurately filling empty areas in point clouds while preserving the integrity of the existing points. This approach outperforms conventional methods that tend to overlook local features. SA-Net \cite{wen2020point} utilizes skip-attention to prioritize local features, resulting in improved geometry prediction. The system's decoder also utilizes hierarchical folding to improve the representation of structural features at various resolutions. Lin et al. \cite{lin2022cosmos} presented the cosmos propagation network, highlighting its  mirror expand module that decreases information duplication and enhances point distribution. Huang et al. \cite{huang2024gmp} combined double perspective contrastive learning to improve feature extraction, and a unique tree structure decoder with graph attention to produce and refine missing point clouds with high accuracy. 

\paragraph{\textbf{Deformation-based}}
The deformation-based methods involve iterative updating to produce a complete point cloud. During each deformation phase, the positions of the input points are modified by incorporating the coordinate offsets acquired from the displacement prediction module.

Zhang et al. \cite{zhang2024walkformer} introduced WalkFormer, a method that uses partial deformation to predict entire point clouds accurately. The performance of point cloud completion is enhanced by employing a multi-step partial deformation approach.  Wen et al. \cite{wen2021pmp} introduced PMP-Net, which aims to enhance the quality of point cloud completion by efficiently shifting each point from an incomplete shape to a complete one. It forecasts a distinct trajectory for each point. Consequently, the network acquires a precise and exclusive mapping at the level of individual points, enhancing the accuracy of the anticipated overall form. To further boost PMP-Net, PMP-Net++ \cite{wen2022pmp} formulates completion as a point cloud deformation process. This method employs a transformer-enhanced neural network to predict the shortest point moving paths, ensuring detailed topology and structure. Key features include multi-step path searching and a recurrent path aggregation module.

\paragraph{\textbf{Others}}
Li et al. \cite{li2023generalized} proposed a method that enables the production of shapes with high accuracy and diversity for different tasks. Cheng et al. \cite{cheng2023nsds} introduced NSDS, a method that utilizes a pretrained diffusion model to enhance the null-space content of degraded point clouds. This approach achieves a high level of completion accuracy without requiring extra training. VRCNet \cite{pan2021vrcnet} employs a dual-path design and attention processes, resulting in an enhanced representation of both local and global structural links. 

\subsubsection{\textbf{Voxel/Projection-based}}
These techniques facilitate the transformation of point clouds into different formats. For instance, point clouds can be voxelized or converted to depth maps.

Xie et al. \cite{xie2020grnet} developed two innovative differentiable layers that can convert point clouds into 3D grids and vice versa while preserving the integrity of the structural data. Instead of employing a single vector to represent a 3D form, implicit feature networks (IF-Nets) \cite{chibane2020implicit} extract a learnable 3D multi-scale tensor of deep features. It can provide continuous outputs and cater for different topologies. In addition, they are capable of generating complete forms for incomplete or limited input data while maintaining the favorable attributes of freshly acquired implicit functions. To solve the problem that previous approaches fail to reconstruct realistic structures due to excessive flattening of fine-grained features, Wang et al. \cite{wang2021vepcn} initially included point clouds into standardized voxel grids and subsequently produced entire objects utilizing the assistance of the imagined shape edges. Then, Wei et al. \cite{wei2022cyclic} presented a methodology known as CGG-Net. The method employs a distinctive blend of layered folding and globally directed upsampling to improve the accuracy and structural soundness of the produced point clouds. Instead of voxelizing the point cloud, Phong Nguyen et al. \cite{nguyen2023bilateral} integrated 2D images as guidance to tackle the issue of structural degradation in incomplete point clouds. They proposed a multi-level codeword combination approach to forecast and preserve structural intricacies accurately. Afterwards, Long et al. \cite{long20233d} introduced MRC-Net. This network combines multi-scale degradation and multi-discriminator techniques with the GAN inversion paradigm.

\subsubsection{\textbf{Image-assisted}}
As only partial information can be utilized through incomplete point clouds, it is challenging to reconstruct the missing data. The idea of image-assisted completion method is introduced to provide more information for point cloud completion. These methods leverage a single-view image to gather more information, offering a more applicable solution for handling incomplete point clouds.\par
ViPC \cite{zhang2021view} is the first image-assisted method in the point cloud completion task. ViPC uses an extra single-view image to explicitly provide the crucial global structural prior information for completion. Specifically, ViPC uses a modality transformer to convert images directly into a skeleton point cloud, which is then combined with the occluded point cloud. It then refines this merged data by utilizing both the concatenated image features and point cloud features. Moreover, the ShapeNet-ViPC dataset is proposed for training and testing. As ViPC is limited by the need to estimate a point cloud from the single-view image, XMFnet \cite{aiello2022cross} performs fusion in a latent domain. Specifically, XMFnet utilizes cross-attention operations to combine the features of two modalities and a flexible decoder to achieve completion with varied sizes. Compared to previous works, CSDN \cite{zhu2023csdn} develops a novel coarse-to-fine network, which significantly enhances completion performance. CSDN proposes a shape fusion module for generating a coarse complete shape. Then it is refined through the dual-refinement module. In CDPNet \cite{du2024cdpnet}, it utilizes the image and patches to learn global and local information respectively. A patch generator is also proposed to generate fine-grained patches through coarse patch features. To find the critical information contained in a corresponding image, EGIINet \cite{xu2024explicitly} identifies the critical information within images by explicitly guiding the information interaction. It achieves completion through two steps: modal alignment and information fusion.

\subsection{Scene-level Completion}
Due to the close relationship between semantic information and scene completion, the process of scene completion emphasizes both the reconstruction of the scene itself and the accurate completion of its semantic labels. Since our method is based on 3D point cloud completion, we consider all semantic scene completion methods that utilize 3D input encoding. 

To achieve semantic scene completion, we have classified existing methods into three categories based on the different 3D input encodings of the model network: point-based, volume-based, and RGB-D-based. All 3D input encodings are capable of being converted to each other. Moreover, each category's input encoding is derived from the same dataset before training. Although the input encodings differ after the dataset conversion, the output results are measured against a unified standard. The various input encoding techniques can aid in the reconstruction of missing shapes and corresponding semantic information in diverse ways. 

\subsubsection{\textbf{Point-based}}
As a prevalent form of 3D representation, point clouds contain valuable information, such as color and spatial position. Nevertheless, the coexistence of multiple objects and the large number of points within a scene pose challenges for semantic recognition and scene understanding. Thus, existing solutions can be classified as 3D-segmentation-based and 3D-segmentation-free.

\paragraph{\textbf{3D-segmentation-based}} These methods typically employ 3D point cloud segmentation methods to accomplish semantic scene completion. They often extract the necessary 3D semantic information from the scene point cloud \cite{zhang2021point, yan2021sparse, li2023ddit}. 

PCSSC-Net \cite{zhang2021point} uses the patch-based contextual encoder to extract scene point clouds both geometry and contextual information at different scales. After that, extracted features guide semantic label prediction point clouds completion of the scene through the semantic-guided completion decoder. Subsequently, to exploit the correlation and complementarity between scene completion and semantic segmentation, CasFusionNet \cite{xu2023casfusionnet} adopts a cascaded network to achieve scene completion and semantic label prediction from coarse to fine. In CasFusionNet, the global completion module (GCM) first generates rough completion results. Then, the semantic segmentation module predicts the semantic labels of the generated points in GCM. Finally, the local refinement module refines the rough completion results and semantic labels. To further address challenges such as sparse and incomplete inputs and the presence of a large number of objects of varying scales, SCPNet \cite{xia2023scpnet} transfers the rich semantic information from the multi-frame teacher to the single-frame student to improve its representation learning ability.

\paragraph{\textbf{3D-segmentation-free}} These methods directly extract various features from point clouds to assist scene reconstruction and semantic annotation \cite{rist2020scssnet, zhong2020semantic, wang2022learning, yu2023adapointr}. 

IPF-SPCNet \cite{zhong2020semantic} first uses point clouds as input encoding for semantic scene completion. IPF-SPCNet is based on an encoder-decoder structure, which uses an observed point encoder to extract features from observed points and an observed occluding-point decoder to map the extracted features to occluded points. Unlike other semantic scene completion methods that use voxelization, Rist et al. \cite{rist2021semantic} proposed a continuously local depth implicit function to continuously represent the entire scene. They assembled a global scene completion function by assembling the scene point cloud at different spatial resolutions. SCSFNet \cite{wang2024semantic} first achieves semantic scene complete and future scene forecasting simultaneously. Taking a 4D dynamic point cloud sequence as input, SCSFNet uses an attention-based skip connection scheme and auxiliary visibility grids for high-resolution scene completion and scene forecasting.

\subsubsection{\textbf{Volume-based}}
As the earliest proposed solution, volume-based methods can directly utilize the rich 3D information contained in various kinds of volumes. The extracted information can be utilized to assign semantic labels to objects within the scene, and scene completion can be achieved by identifying interrelated features among objects. 
These methods can be categorized into two types: TSDF-based and voxel-based.

\paragraph{\textbf{TSDF-based}} These methods usually convert depth maps into TSDF inputs \cite{zhang2018efficient, cherabier2018learning, wang2019forknet}. Typically, features are extracted separately from TSDF and RGB images. Scene completion and semantic label prediction can be achieved subsequently by integrating both types of features. 

Song et al. \cite{song2017semantic} first considered scene completion and semantic labels of depth maps simultaneously. They employed a 3D context module based on dilation techniques to effectively increase the receptive field and facilitate the learning of 3D context. In addition, they defined metrics like IoU and mIoU for evaluation and datasets like SUNGCG for training and testing, which are widely used by subsequent methods. Dai et al. \cite{dai2018scancomplete} proposed ScanComplete to address the challenges posed by large-scale scenes of varying spatial extent. In ScanComplete, it efficiently handles the exponential growth in data size that occurs as the scene size increases. To restore object details and achieve full-resolution predictions effectively, CCPNet \cite{zhang2019crn} introduces a cascaded context pyramid to enhance labeling coherence. In CCPNet, the guided residual refinement module is designed to progressively restore the fine structures of objects based on low-level features. To obtain rich geometric information from low-resolution representations, Chen et al. \cite{chen20203d} proposed a novel approach that leverages a geometry-based strategy to embed depth information via a low-resolution voxel representation. They also introduced a method to generate structure prior information from the TSDF input, which is then used to predict occupancy and object labels for each voxel. Cai et al. \cite{cai2021semantic} simultaneously considered the completion of instances and scenes. They iterated the information extracted from both instances and scenes multiple times to generate a coarse-to-fine semantic scene completion result. These methods directly extract 3D features from voxel grids. 

To acquire a noise-free model, Wang et al. \cite{wang2023semantic} proposed to train the teacher model with noise-free data so that the model will focus more on scene completion and not be affected by noise. Then, they distilled the noise-free knowledge from the teacher model and applied it to train the student model with noisy TSDF input. AMMNet \cite{wang2024unleashing} combines features from different modal inputs to correct semantic labels. It also presents a discriminator to avoid overfitting the dataset.

\paragraph{\textbf{Voxel-based}} These methods directly extract features from voxels \cite{garbade2019two, wilson2022motionsc, an2024esc}. 

Garbade et al. \cite{garbade2019two} utilized RGB images to aid in scene completion and semantic scene completion. They designed a two-stream network comprising the color input stream and the depth input stream to extract features respectively. MotionSC \cite{wilson2022motionsc} aims to achieve semantic scene completion for dynamic scenes by integrating existing deep-learning frameworks. Moreover, this work fills a gap in semantic scene completion data by introducing a novel outdoor dataset featuring accurate and complete dynamic scenes. To extract context features from long-range scenes, DST-Net \cite{wang2023learning} employs a two-stage completion strategy. At the first stage, DST-Net divides the input scene into multiple blocks and conducts completion on each block. At the second stage, the blocks are merged as a whole and further refined to enhance integrity. An et al. \cite{an2024esc} proposed ESC-Net to deal with extremely sparse scene completion. They corrected the semantic labels of the scene by using the multi-view multi-task attention loss function and thus restored the high-quality boundaries of the scene.

\subsubsection{\textbf{RGB-D-based}}
RGB-D images contain both color and depth information. Extracting effective information from 2D RGB-D images to facilitate 3D scene completion and semantic label prediction poses a challenge. To fully harness the information within RGB-D images, RGB-D-based methods typically extract feature information from the RGB-D images before mapping the image data into a 3D space. RGB-D-based methods can be categorized into 2D-segmentation-based and 2D-segmentation-free.

\paragraph{\textbf{2D-segmentation-based}}
These methods typically utilize 2D segmentation networks to extract semantic features from 2D images \cite{liu2018see, li2020attention, li2023front}. 

Liu et al. \cite{liu2018see} first proposed a framework to utilize the information contained in RGB-D images. They divided semantic scene completion into three steps: segmentation of 2D images, reprojecting of 2D depth images into 3D space, and semantic scene completion for 3D scenes. This framework has been followed up with subsequent works. Li et al. \cite{li2020attention} designed a 2D semantic segmentation network and a 3D semantic completion network. They fused multimodal features and input them into the network to enhance the ability to extract input information. Considering the large disparity in RGB-D images and the uncertainty measurements of depth data, FFNet \cite{wang2022ffnet} correlates the RGB-D images in the frequency domain, which can better learn color and depth information. Zhang et al. \cite{zhang2023point} proposed an end-to-end approach for colored semantic point cloud scene completion from a single RGB-D image, which can directly predict the coordinates, colors, and semantic information of missing point clouds. To complete and segment the rear view of the scenes, Li et al. \cite{li2023front} tried to extend the high-precision prediction results of the front view to the area of the back view to achieve semantic completion for the entire scene.

\paragraph{\textbf{2D-segmentation-free}}
These methods typically utilize convolution-based neural networks to extract informative features \cite{li2019rgbd, li2020anisotropic, li2021anisotropic}. 

Li et al. \cite{li2019rgbd} tried to integrate the depth and color features of RGB-D images through multi-level feature fusion. They also designed dimensional decomposition residual blocks to form the entire network, significantly reducing the number of model parameters. To handle variations of objects in the scene, Li et al. \cite{li2020anisotropic} proposed anisotropic convolution to effectively leverage the 3D context for modeling diverse objects or environments. The anisotropic convolution-based module can automatically choose proper receptive fields for different voxels and model the dimensional anisotropy voxelwise. Based on \cite{li2020anisotropic}, Li et al. \cite{li2021anisotropic} optimized the anisotropic module to the kernel-selection anisotropic module, the kernel-modulation module, and the dimensional decomposition residual module, resulting in a more efficient network structure.

\subsection{Summary}
We summarize the development and characteristics of point cloud completion methods as follows:

\begin{enumerate}
    \item \textbf{Object-level completion methods.} In 3D object completion task, the unordered nature of point clouds is a significant challenge. Point-based approaches independently operate on individual data points and maintain permutation invariance property. They sometimes overlook the geometric relationship between neighboring points. Convolution-based techniques often utilize 3D CNNs to convert point clouds into grids or voxels in order to handle their unordered structure. They meet challenges in determining the dimensions of grids or voxels and can be inefficient when dealing high-resolution upsampling. At present, generative-based and transformer-based methods generally yield outstanding performance and are increasingly popular.
    
    
    
    
    \item \textbf{Scene-level completion methods.} The primary objective of scene-level completion is semantic scene completion. As shown in Fig. \ref{point cloud completion for scene}, early methods like SSCNet \cite{song2017semantic}, SATNet \cite{liu2018see} and CCPNet \cite{zhang2019crn} predominantly utilize TSDF or RGB-D images as inputs. The formal exploration of using point clouds as direct input encoding is IPF-SPCNet \cite{zhong2020semantic}, which emerges relatively late. This delay is attributed primarily to the limitation of hardware and feature representation ability of neural networks in early stages. However, with recent advancements in deep learning and hardware capabilities, point clouds have increasingly become a preferred representation for scene completion.

    \item \textbf{Challenges and limitations.} While current point cloud completion methods have achieved promising results, they still face challenges or limitations in several aspects. For object-level completion, current work focuses mainly on synthetic datasets, which may hardly render real-world data traits. For scene-level completion, current methods fail to fully utilize object-level information. Challenges also include heavy occlusion in real-world applications and the issue of texture completion for color point clouds.
    
    \item \textbf{Research tendency.} First, performing object-level and scene-level completion simultaneously is necessary, as scenes usually contain objects. Their information are complementary. Second, multi-modal completion for point clouds becomes the trend, especially for scene completion, as scene data usually are equipped with color cues. Simultaneously recovering texture and shape remains an open problem.

\end{enumerate}
\section{3D Point Cloud Upsampling}
\begin{figure*}[t]
    \centering
    \includegraphics[width=\textwidth]{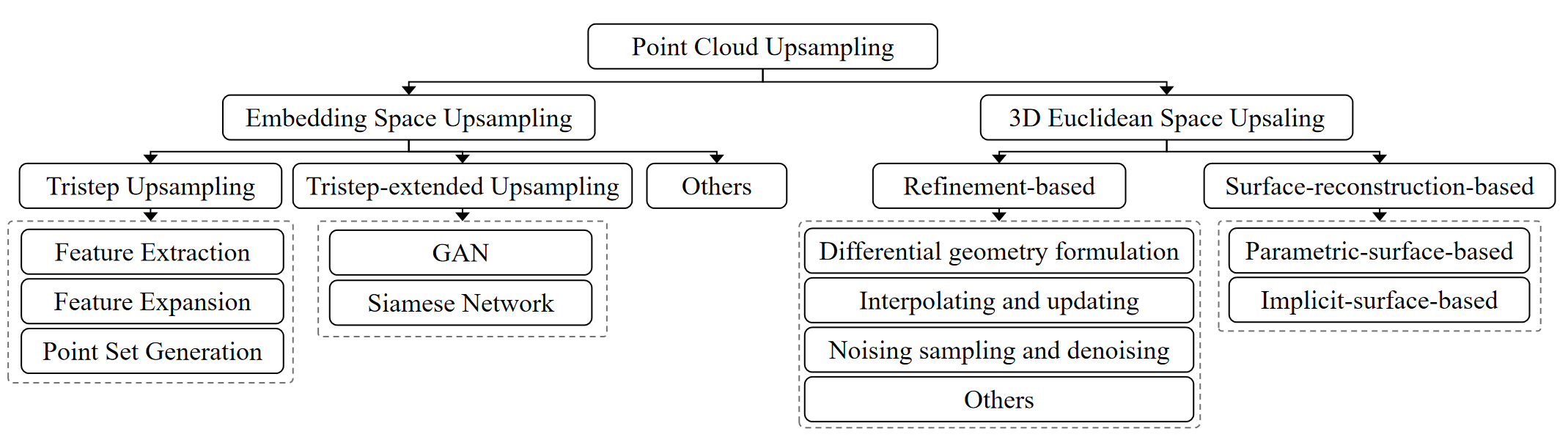}
    \caption{A detailed taxonomy of point cloud upsampling methods.}
    \label{fig:1}
\end{figure*}
\begin{figure*}[t]
    \centering
    \includegraphics[width=\textwidth]{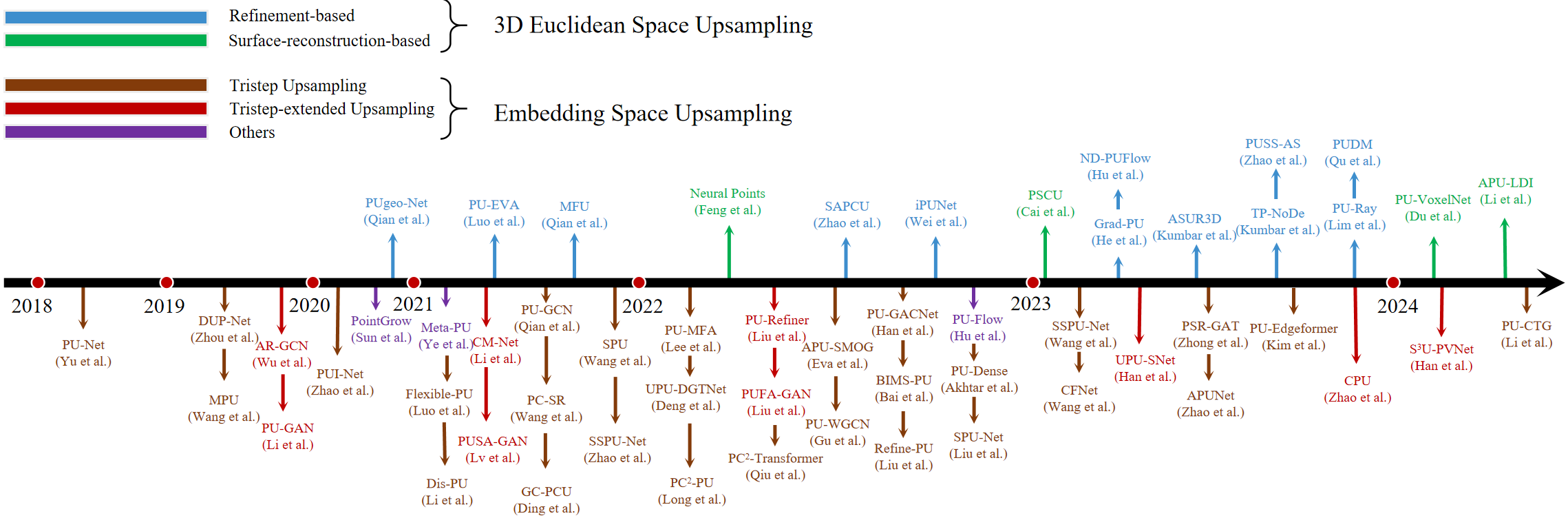}
    \caption{Chronological overview of representative point cloud upsampling networks.}
    \label{fig:2}
\end{figure*}
The point cloud upsampling task aims to generate denser point clouds from sparse point cloud data, in order to improve the resolution of the point cloud while maintaining the integrity of the original geometric structure as much as possible. Given the original sparse point cloud as $P_{sparse} = \{p_i | i = 1,2,\ldots,N\}, p_i \in \mathbf{R}^3$, where $N$ denotes the number of points. The goal of the point cloud upsampling task is to produce a dense point cloud $P_{dense} = \{q_j | j = 1, 2, \ldots, rN\}, q_i \in \mathbf{R}^3$, where $r$ denotes the upsampling rate. The upsampled point cloud $P_{dense}$ should cover the same object surface more densely and uniformly, while preserving the geometric information of the original point cloud $P_{sparse}$.\par
\begin{figure*}[t]
    \centering
    \includegraphics[width=\textwidth]{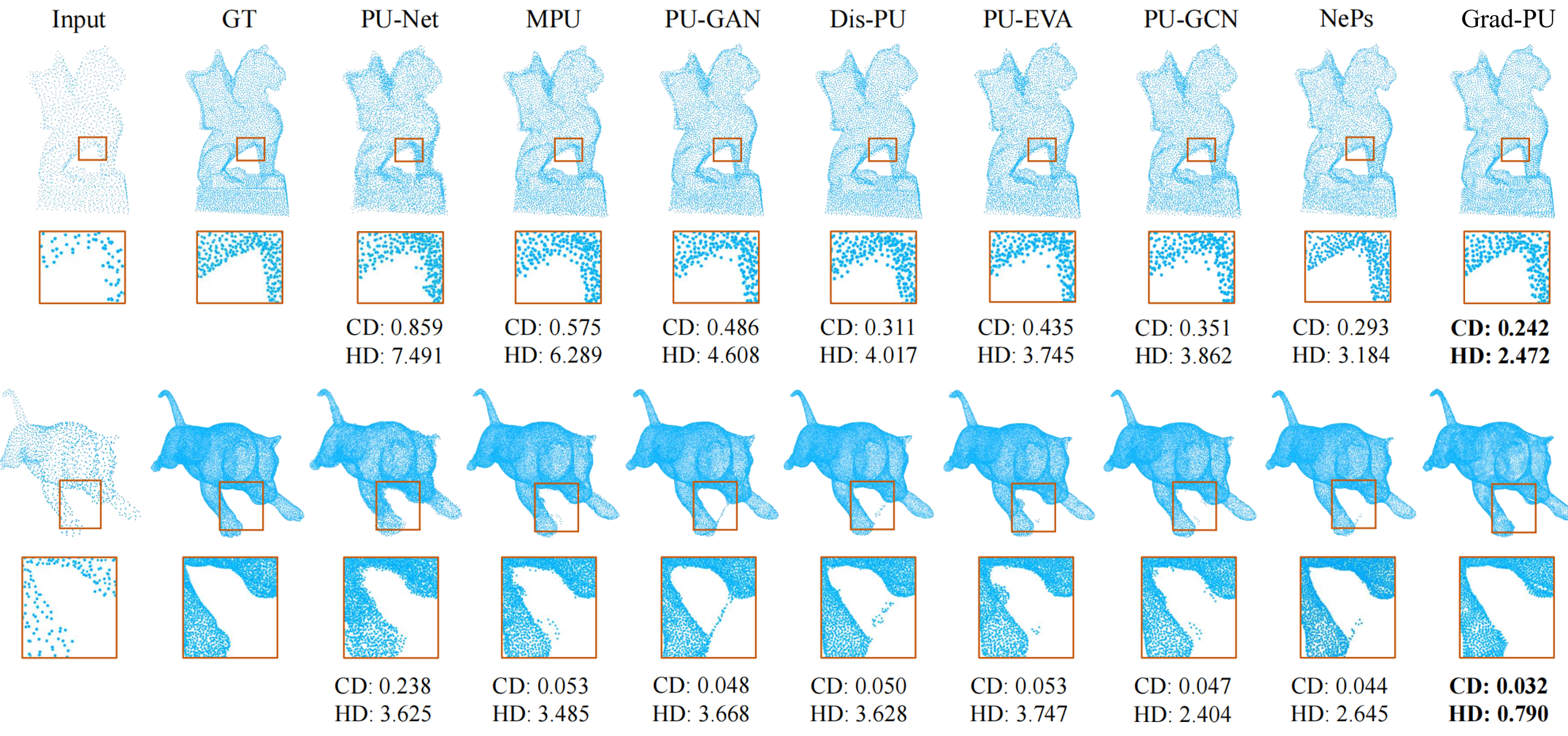}
    \caption{Qualitative results on the PU-GAN dataset under two evaluation scales: the first row corresponds to a 4× evaluation, while the second row pertains to a 16× evaluation. Both the Chamfer Distance (CD) and Hausdorff Distance (HD) metrics are measured in units of $10^{-3}$. These results are derived from Grad-PU \cite{he2023grad}.}
    \label{upsamplingvisual}
\end{figure*}

Deep-learning-based point cloud upsampling involves using a deep neural network to transform sparse point clouds into denser versions by learning a mapping from sparse to dense configurations, capitalizing on the network's capacity to identify inherent patterns in the data for high-quality upsampling results.
Our research categorizes upsampling methods into embedding-space based upsampling and 3D Euclidean space based upsampling, as presented in Fig. \ref{fig:1}. The chronological overview of point cloud upsampling approaches is provided in Fig. \ref{fig:2}. It is found that embedding space-based upsampling is the mainstream approach, significantly outnumbering 3D-Euclidean-space-based upsampling methods. In the category of embedding-space-based upsampling, tristep-based upsampling methods have attracted the most attention. We also report the experimental results of representative point cloud upsampling methods in Table \ref{upsamplingtable}. Some visual results are depicted in Fig. \ref{upsamplingvisual}.

\begin{table}[t]
  \centering
  \caption{Code URL for point cloud upsampling methods. The symbol `-' indicates that the complete code is still unavailable currently.}
  \vspace{-0.2cm}
\resizebox*{\linewidth}{!}{
\begin{tabular}{|c|c|c|}
\hline
\multicolumn{1}{|c|}{\multirow{2}{*}{Year}} & \multirow{2}{*}{Method}  & \multirow{2}{*}{URL} \\
\multicolumn{1}{|c|}{}                      &                          &                      \\ \hline
\multirow{2}{*}{2018}                                        & PU-Net \cite{yu2018pu}     & https://github.com/yulequan/PU-Net   
\\ \cline{2-3} & EC-Net \cite{yu2018ec} & https://github.com/yulequan/EC-Net
\\ \hline
\multirow{3}{*}{2019}                       & {AR-GCN} \cite{wu2019point}       & https://github.com/wuhuikai/PointCloudSuperResolution                     \\ \cline{2-3} 
                                            & MPU \cite{yifan2019patch}    & https://github.com/yifita/3pu 
     \\ \cline{2-3}
                                            & PU-GAN \cite{li2019pu} &  https://github.com/liruihui/PU-GAN
                                            \\ \hline
\multirow{1}{*}{2020}                       & PUGeo-Net \cite{qian2020pugeo}        & https://github.com/ninaqy/PUGeo                         \\ \hline
\multirow{6}{*}{2021}                       & SSPU-Net \cite{zhao2021sspu}    & -                     \\ \cline{2-3} 
                                            & Dis-PU \cite{li2021point}     & https://github.com/liruihui/Dis-PU                     \\ \cline{2-3} 
                                            & PU-GCN \cite{qian2021pu}        & https://github.com/guochengqian/PU-GCN                    \\ \cline{2-3} 
                                            & PU-EVA \cite{luo2021pu}       & -   
                                            \\ \cline{2-3} & CM-Net \cite{li2021cm}
                                            & -
                                            \\ \cline{2-3} & Flexible-PU \cite{qian2021deep} & https://github.com/ninaqy/Flexible-PU
                                            \\ \hline
\multirow{21}{*}{2022}                       & APU-SMOG \cite{dell2022arbitrary}       & https://github.com/apusmog/apusmog
                                            \\ \cline{2-3} & PU-Refiner \cite{liu2022pu} & -
\\ \cline{2-3} 
                                            & PC2-PU \cite{long2022pc2}       &  https://github.com/chenlongwhu/PC2-PU                     \\ \cline{2-3} 
                                            & NePs \cite{feng2022neural}  & https://github.com/WanquanF/NeuralPoints  
                                            \\ \cline{2-3} & SAPCU \cite{zhao2022self} &  https://github.com/xnowbzhao/sapcu
                                            \\ \cline{2-3} & ZSPU \cite{zhou2022zero} &  https://github.com/ky-zhou/ZSPU.
                                            \\ \cline{2-3} & PU-Dense \cite{akhtar2022pu} & https://github.com/aniqueakhtar/PointCloudUpsampling
                                            \\ \cline{2-3} & PUFA-GAN \cite{liu2022pufa} & https://github.com/yuanhui0325/PUFA-GAN
                                            \\ \cline{2-3} & SPU-Net \cite{liu2022spu} & https://github.com/liuxinhai/SPU-Net
                                            \\ \cline{2-3} & PU-Flow \cite{mao2022pu} & https://github.com/unknownue/pu-flow
                                            \\ \cline{2-3} & {iPUNet} \cite{wei2023ipunet} & https://github.com/GSW-D/iPUNet
                                            \\ \cline{2-3} & {Meta-PU} \cite{ye2021meta} & https://github.com/pleaseconnectwifi/Meta-PU
                                            \\ \cline{2-3} & Refine-PU \cite{liu2022refine} & -
                                            \\ \cline{2-3} & PU-Transformer \cite{qiu2022pu} & -
                                            \\ \cline{2-3} & PU-WGCN \cite{gu2022pu} & -
                                            \\ \cline{2-3} & PU-GACNet \cite{han2022pu} & https://github.com/BingHan0458/PU-GACNet
                                            \\ \cline{2-3} & BIMS-PU \cite{bai2022bims} & -
                                            \\ \cline{2-3} & UPU-DGTNet \cite{deng2022upu} & -
                                            \\ \cline{2-3} & PU-CRN \cite{du2022point} & https://github.com/hikvision-research/3DVision
                                            \\ \cline{2-3} & PU-MFA \cite{lee2022pu} & -
                                            \\ \cline{2-3} & PU-CycGAN \cite{li2022weakly} & https://github.com/cognaclee/PU-CycGAN
                                            \\ \hline
\multirow{13}{*}{2023}                       & PSCU \cite{cai2023parametric}        & https://github.com/corecai163/PSCU                     \\ \cline{2-3} 
                                            & CPU \cite{zhao2023cpu}   & -                     \\ \cline{2-3} 
                                            & PUDM \cite{qu2023conditional}      & https://github.com/QWTforGithub/PUDM                     \\ \cline{2-3} 
                                            & PU-Ray \cite{lim2023pu}     & https://github.com/sum1lim/PU-Ray                     \\ \cline{2-3} 
                                            & Grad-PU \cite{he2023grad}       & https://github.com/yunhe20/Grad-PU                     \\ \cline{2-3} 
                                            & {PU-SSAS} \cite{zhao2023self}  & https://github.com/xnowbzhao/PU-SSAS  
                                            \\ \cline{2-3} & S3U-PVNet \cite{han2024s3u} & -
                                            \\ \cline{2-3} & CFNet \cite{wang2023cfnet} & -
                                            \\ \cline{2-3} & TP-NoDe \cite{kumbar2023tp} & https://github.com/Akash-Kumbar/TP-NoDe
                                            \\ \cline{2-3} & ASUR3D \cite{kumbar2023asur3d} & -
                                            \\ \cline{2-3} & APUNet \cite{zhao2023apunet} & -
                                            \\ \cline{2-3} & ND-PUFlow \cite{hu2023noising} & -
                                            \\ \cline{2-3} & UPU-SNet \cite{han2023upu} & -
                                            \\ \hline
\multirow{7}{*}{2024}                       & PU-VoxelNet \cite{du2024arbitrary}      & -                     \\ \cline{2-3} 
                                            & APU-LDI \cite{li2024learning}     & https://github.com/lisj575/APU-LDI   
                                            \\ \cline{2-3} & PU-CTG \cite{li2024pu} & - \\
                                            \cline{2-3} & {PU-Mask} \cite{liu2024pu} & https://github.com/liuhaoyun/PU-Mask \\
                                            \cline{2-3} & {PU-SDF} \cite{pan2024pu} & - \\
                                            \cline{2-3} & {SPU-PMD} \cite{liu2024spu} & - \\
                                            \cline{2-3} & {RepKPU} \cite{rong2024repkpu} & https://github.com/EasyRy/RepKPU
                                            \\ \hline
\end{tabular}
}
  \label{upsamplingurl}%
\end{table}%

\subsection{Embedding Space Upsampling}
We categorize the upsampling networks further based on their structures into tristep upsampling networks, tristep-extended ones  and the others.
\subsubsection{\textbf{Tristep Upsampling Networks}}
\begin{figure}
    \centering
    \includegraphics[width=\linewidth]{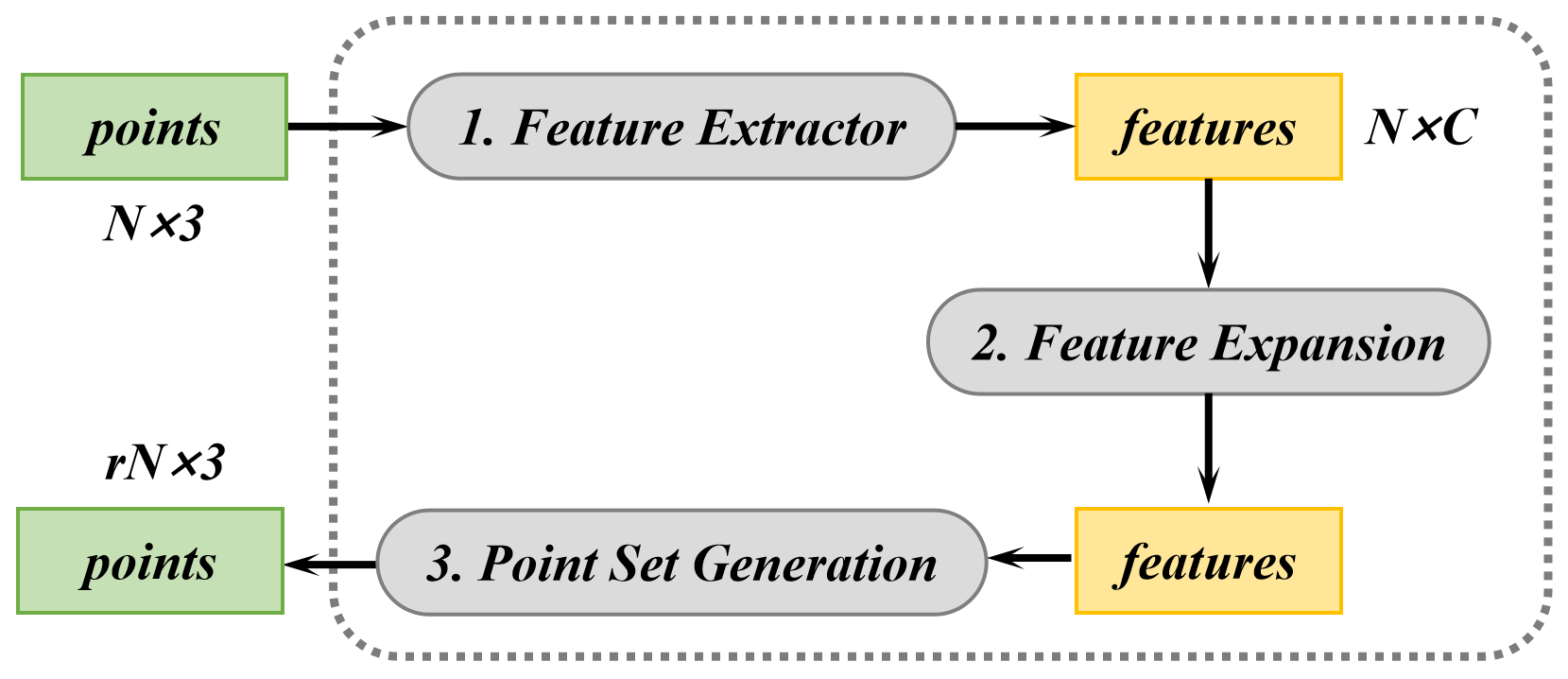}
    \caption{A schematic diagram of tristep upsampling frameworks. Here, \textit{N} denotes the number of points, \textit{C} denotes the dimensionality of the learned features for each point, and \textit{r} denotes the upsampling rate.}
    \label{tristep}
    \vspace{-0.3cm}
\end{figure}

The tristep upsampling framework mainly consists of feature extraction, feature expansion, and point set generation. A schematic diagram of the tristep upsampling framework is shown in Fig. \ref{tristep}.
\paragraph{\textbf{Feature Extraction}} It is a core component in upsampling networks, and the diversity of their techniques provide new perspectives and possibilities for point cloud upsampling. Currently, feature extraction can be roughly categorized into four main types, i.e., PointNet-based, graph-convolutional-network-based (GCN-based), voxel-convolution-based, and attention-based.\par
\begin{figure*}[t]
    \centering
    \includegraphics[width=0.9\linewidth]{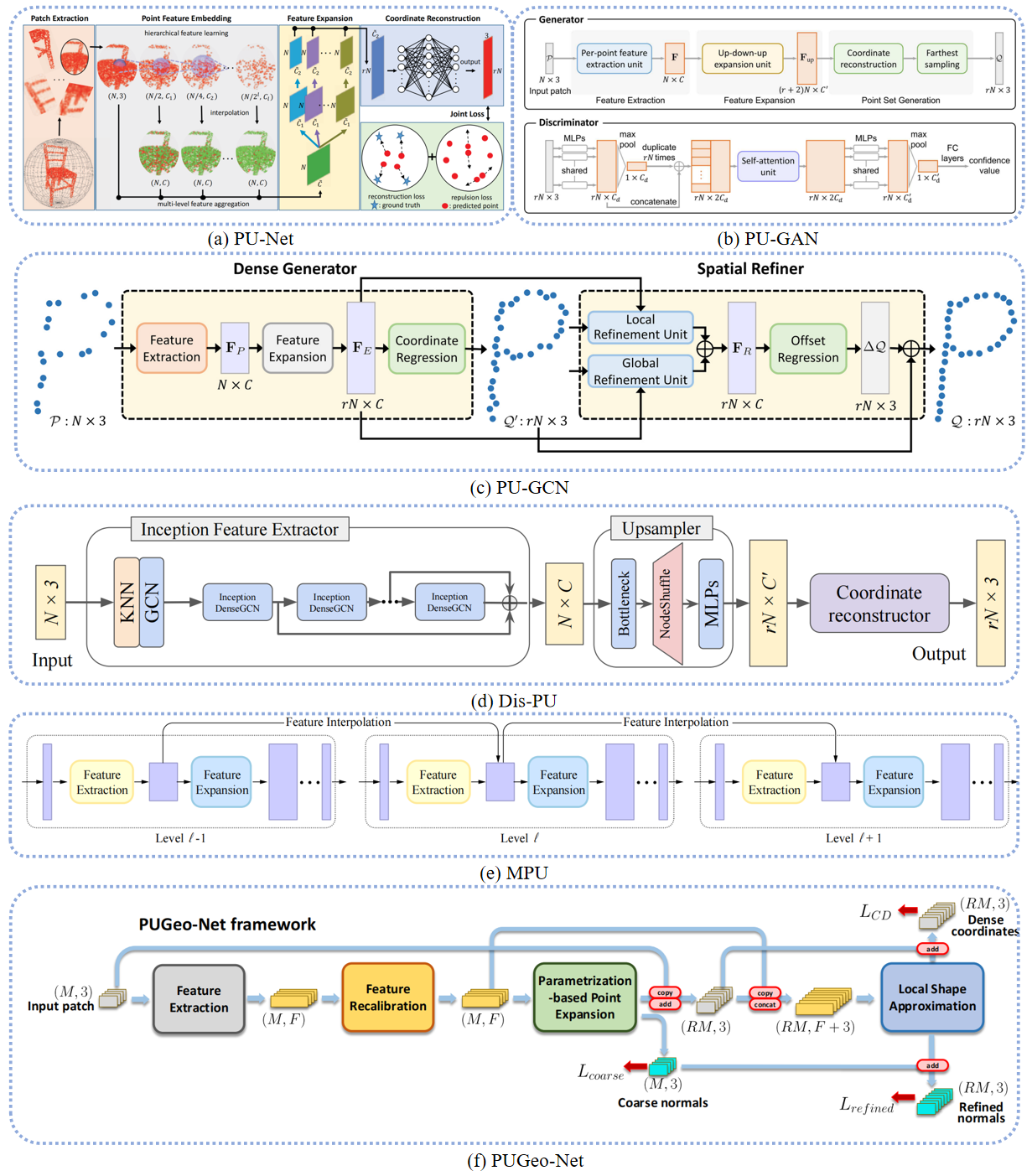}
    \caption{Pipelines of representative point cloud upsampling methods.}
    \label{classical}
\end{figure*}

\textbf{PointNet-based.}
PointNet \cite{qi2017pointnet} is a pioneering deep learning model tailored for the direct manipulation of 3D point clouds through a permutation-invariant architecture. PointNet++ \cite{qi2017pointnet++} is developed later to further improve the model's robustness to local structures at varying scales via a hierarchical strategy. This strategy involves the recursive application of PointNet across increasingly granular subdivisions of the input dataset, significantly bolstering the model's proficiency in discerning complex patterns and configurations. In a subsequent advancement, PuNet \cite{yu2018pu} extends the foundational concepts of PointNet and PointNet++ to learn from sparse inputs and generate denser point clouds, as illustrated in Fig. \ref{classical}a. Despite these improvements, PuNet encounters challenges in bridging substantial gaps or augmenting minute, poorly sampled structures. It constrains its applicability in scenarios demanding thorough and intricate detail, such as the conservation of cultural heritage or industrial inspections.\par
Based on these innovations, Sharma et al. \cite{sharma2021point} leveraged PointNet for the extraction of global features and PointNet++ for local feature extraction, thereby enhancing accuracy and efficiency. Furthermore, Zeng et al. \cite{zeng2021point} introduced an approach for multi-level spatial local feature aggregation within the frameworks of PointNet and PointNet++. It significantly elevates the networks' prowess in understanding spatial contexts and learning hierarchical features. PointRNN, introduced by Fan et al. \cite{fan2019pointrnn}, integrates recurrent neural networks with the PointNet architecture to capture temporal dependencies, facilitating dynamic modeling of point cloud data. Moreover, numerous studies \cite{zhang2019data,zhou2019dup,li2019pu,zhao2021sspu} have utilized the foundational architectures of PointNet and PointNet++ for the purpose of feature extraction, demonstrating the remarkable resilience and flexibility of the PointNet frameworks for point cloud upsampling.\par
\textbf{GCN-based.} Graph convolutional networks (GCN) revolutionize the traditional CNN paradigm to handle graph-structured data. By strategically aggregating features from neighboring nodes, GCNs are proficient in capturing both the local and global patterns inherent within the data. They effectively preserve the geometric and topological information of the underlying structures.\par
Wu et al. \cite{wu2019point} pioneered a deep network architecture that harnesses the power of graph convolution for processing point clouds. It improves the transition from low to high-resolution outputs, through the incorporation of residual connections and direct skip connections. These enhancements leverage local similarities to great effect. By contrast, Wang et al. \cite{wang2019dynamic} introduced the novel Dynamic Graph CNNs, which boasts a dynamic mechanism for updating the graph structure at each layer. This flexibility allows the network to adaptively focus on varying geometric details at different levels of abstraction. A cornerstone of this architecture, i.e., the EdgeConv operation, excels at capturing local geometric structures by computing features on the edges connecting graph nodes, thereby facilitating sophisticated learning mechanisms.\par
Building on the foundation laid by EdgeConv, Qian et al. \cite{qian2021pu} unveiled the inception DenseGCN, optimized for capturing intricate local geometries within point clouds across multiple scales, as depicted in Fig. \ref{classical}c. This model employs parallel graph convolution branches, each designed with unique receptive fields, drawing inspiration from the Inception model used in image processing. This design allows for effective handling of a wide array of structural complexities. Furthermore, DenseGCN enhances intra- and inter-feature integration through the implementation of dense connections.\par
Besides, Liu et al. \cite{liu2022refine} proposed the dense EdgeConv (DEC) architecture. DEC melds local and global features through a densely connected convolutional framework, incorporating tailored DenseGCN blocks with specific kernel sizes, expansion rates, and growth rates. This approach enables DEC to extend its receptive field and efficiently extract multi-scale features, ensuring high resolution in the resultant point clouds.\par
Moreover, Gu et al. \cite{gu2022pu} developed the PU-WGCN network, integrating a weighted graph neural network module to refine point cloud processing, specifically targeting prevalent issues such as hole overfilling and boundary blurring. This strategy adjusts for the high feature similarity in local neighborhoods and the uniform treatment of nodes by emphasizing boundary information and feature distinction through weighted graph connections.\par
Graph attention networks merge the capabilities of graph convolutional networks with attention mechanisms, enabling them to focus selectively on significant areas of the input graph. 
Han et al. \cite{han2022pu} introduced PU-GACNet by utilizing a graph attention convolution module designed to dynamically assign attentional weights. It enables the capture of both global and local structured features of point clouds by combining spatial positions and feature attributes. In addition, Liu et al. \cite{liu2023tucnet} proposed CSAE module, which is able to capture semantic patterns in local regions by sequentially generating attention maps across both channel and spatial dimensions.\par
\textbf{Voxel-convolution-based.} Early adaptations of 2D CNNs to 3D contexts, such as those by Wu et al. \cite{wu20153d} and Tchapmi et al. \cite{tchapmi2017segcloud}, rely on segmenting point clouds into uniform cubes followed by 3D convolutions. However, due to the sparsity of point clouds, these methods are proved to be computationally inefficient. Alternatives include those proposed by Kanezaki et al. \cite{kanezaki2018rotationnet}, Lang et al. \cite{lang2019pointpillars}, and Su et al. \cite{su2015multi}, which project 3D data onto 2D planes. However, they fail to fully leverage the 3D properties of the data.\par
In response, more sophisticated voxel-based techniques have been developed. They utilize a gridding method for mapping points to voxel vertices combined with 3D convolutions, enhancing the handling of point cloud data.\par
Akhtar et al. \cite{akhtar2022pu} introduced an architecture optimized for point cloud processing that includes sparse convolutions tailored to adaptively manage geometric data. This architecture integrates inception-residual blocks and a 3D dilated pyramid block to enable multi-scale feature learning, as well as a pruning mechanism for efficient feature selection.\par
Furthermore, Han et al. \cite{han2024s3u} developed the point-voxel feature fusion modules, which merge detailed geometric capture through a point-based branch with the structured processing of a voxel-based branch. This fusion technique effectively combines the detailed spatial analysis capability of point-based processing with the structural advantages of voxel-based approaches, thereby enhancing precision and organizational efficiency in handling point cloud data.\par
\textbf{Attention-based.} Attention mechanisms enhance point cloud upsampling by dynamically modifying features to cope with the non-uniformity and diverse distributions in raw point clouds. They enable the modeling of long-range dependencies essential for understanding the overall structure and accurately guiding the insertion of new points. By using multi-head attention, these models can effectively capture information across various scales, addressing both macro structures and micro details.\par
Qiu et al. \cite{qiu2022pu} introduced the PU-Transformer, which includes a positional fusion module and a shifted channel multi-head self-attention (SC-MSA) mechanism. The positional fusion module effectively incorporates 3D spatial coordinates into the feature space, enhancing the model's spatial understanding and feature recognition. SC-MSA, on the other hand, shifts feature channels before applying self-attention, improving the model’s capability to process complex spatial relationships. Following this, Lee et al. \cite{lee2022pu} employed a U-Net architecture that uniquely utilizes multi-scale features combined with a cross-attention mechanism. This design allows for adaptive and efficient use of features at different scales. Zhao et al. \cite{zhao2023apunet} proposed APUNet, which features the DisTransformer, an innovative component that integrates distance priors into its attention mechanism. It allows for more accurate modeling of point relationships and enhances object representation by combining global and local features, focusing on both patch and point correlations. Li et al. \cite{li2024pu} introduced the PU-CTG network, which incorporates a cross-transformer-fused module and a GRU-corrected module. This network focuses on enhancing feature interaction and scale fusion, and provides fine-grained feature reconstruction by correcting feedback errors.\par
\begin{figure*}[t]
    \centering
    \begin{subfigure}{0.325\textwidth}
    \centering
    \includegraphics[width=0.99\linewidth,height=0.75\linewidth]{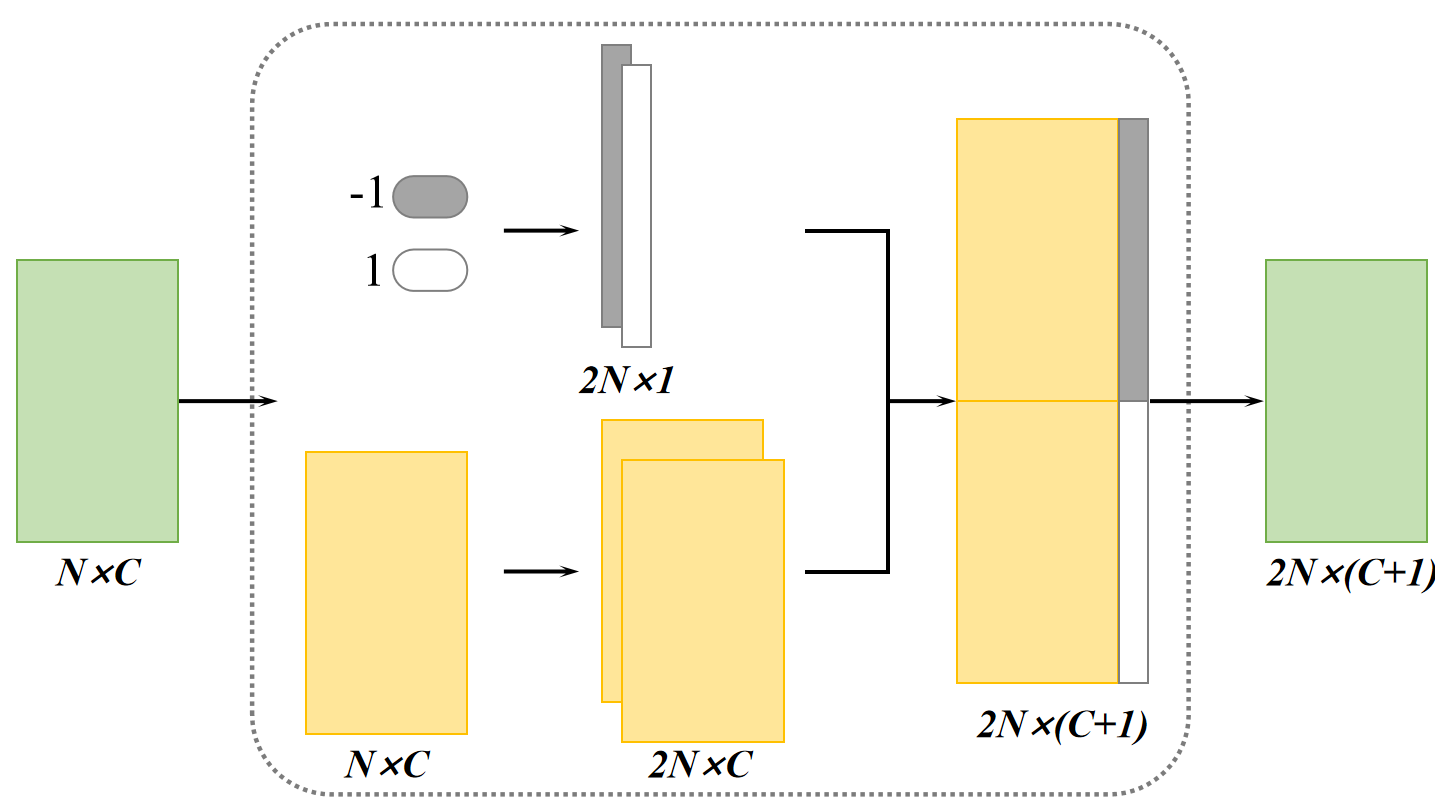}
    \caption{Folding Operation}
    \label{folding}
    \end{subfigure}
    \centering
    \begin{subfigure}{0.325\textwidth}
    \centering
    \includegraphics[width=0.99\linewidth,height=0.76\linewidth]{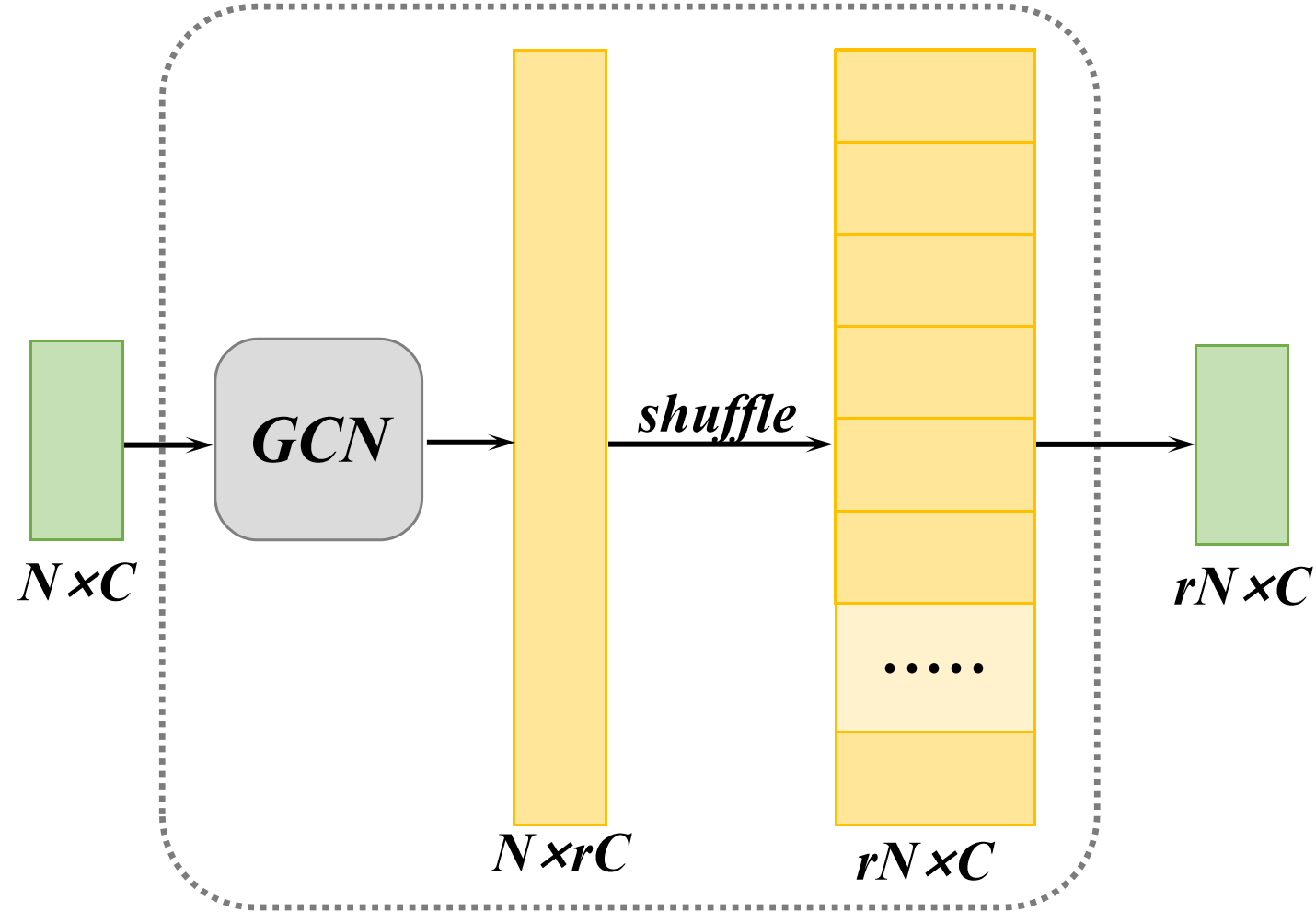}
    \caption{NodeShuffle}
    \label{nodeshuffle}
    \end{subfigure}
    \centering
    \begin{subfigure}{0.325\textwidth}
    \centering
    \includegraphics[width=0.99\linewidth,height=0.75\linewidth]{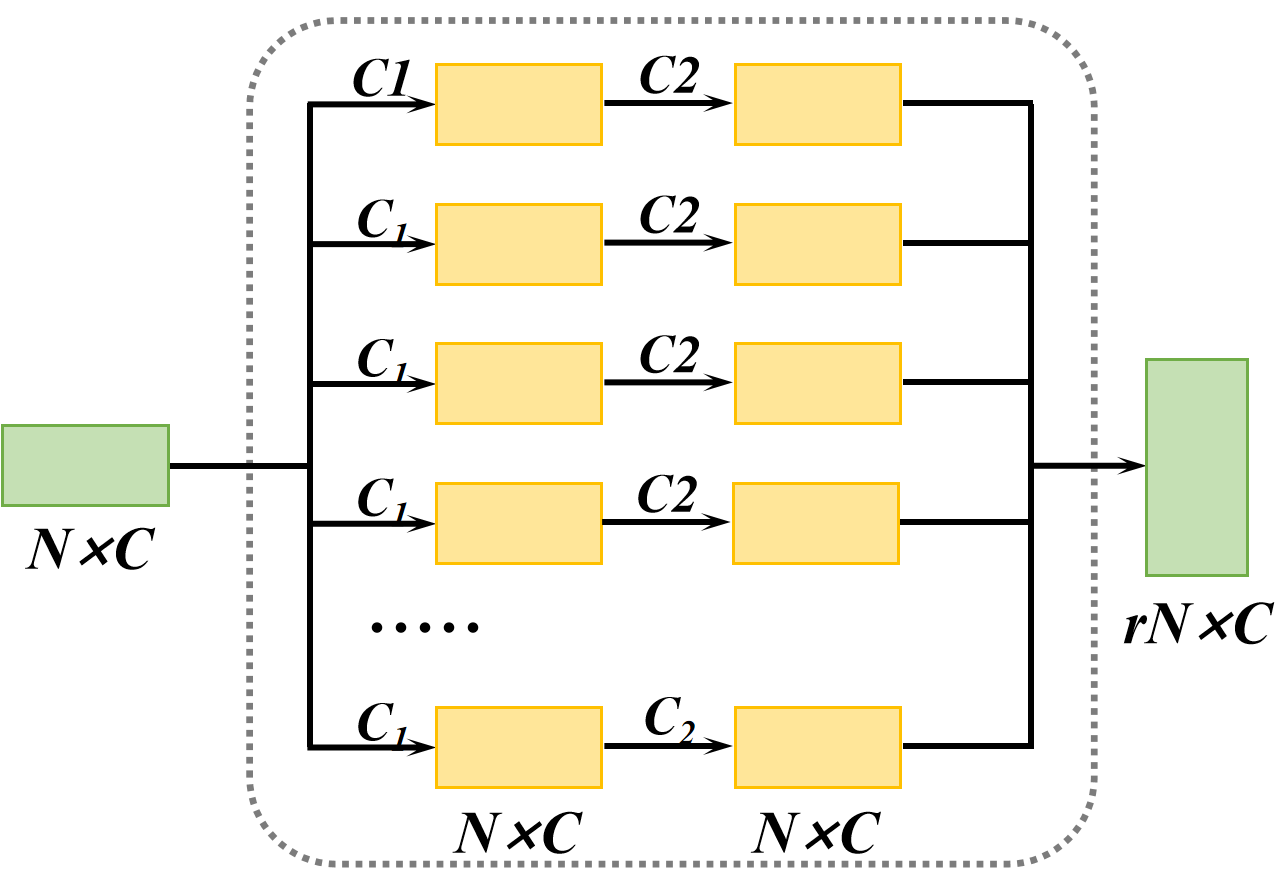}
    \caption{Multi-CNN}
    \label{multi-cnn}
    \end{subfigure}
    \caption{Illustration of the feature epansion module. (a) Folding operation. (b) NodeShuffle. (c) Multi-CNN. Here, \textit{N} denotes the number of points, \textit{C} denotes the dimensionality of the learned features for each point, \textit{r} denotes the upsampling rate;  \textit{$C_1$} and \textit{$C_2$} denote multiple sets of $1 \times 1 $ convolutions.}
    \label{expansion}
\end{figure*}

\paragraph{\textbf{Feature Expansion}} This process involves inflating the original feature set of a sparse point cloud to generate a denser and more informative representation, as shown in Fig. \ref{expansion}. The expanded features could capture more granular details of the 3D structure.\par
\textbf{Folding operation.} A seminar work is the FoldingNet \cite{yang2018foldingnet}, which utilizes a folding operation to transform a 2D grid into a 3D point cloud via a low-dimensional latent space. This method accurately captures the geometry of the object by learning to deform a 2D grid based on the latent space attributes.\par
Wang et al. \cite{yifan2019patch} enhanced this architecture by adding a simple 1D code that controls feature transformations across different locations (Fig. \ref{folding}), facilitating the generation of diverse point clouds without additional loss functions. Similarly, Li et al. \cite{li2019pu} and Zhao et al. \cite{zhao2021sspu} implemented a self-attention mechanism followed by multiple MLPs to upscale feature outputs. Additionally, Liu et al. \cite{liu2022pufa} developed a model incorporating cascaded hierarchical residual attention units that integrate residual blocks and a gating unit to refine feature extraction.\par
\textbf{NodeShuffle.} Inspired by the success of PixelShuffle in the realm of image super-resolution, NodeShuffle \cite{qian2021pu} harnesses the power of graph convolution layers to augment features, coupled with strategic rearrangement via shuffle operations for enhanced representation, as shown in Fig. \ref{nodeshuffle}. Unlike conventional CNN-based methods, NodeShuffle opts for graph convolutions to enrich feature sets, thereby encoding spatial nuances from point neighborhoods and synthesizing new points from the latent space without mere duplication of existing points.\par
Building on this, Han et al. \cite{han2022pu} developed the edge-aware NodeShuffle module, which not only duplicates and expands point features but also reshuffles them. This results in a denser feature set. Furthermore, Han et al. \cite{han2023upu} proposed the TranShuffle module, which incorporates a transformer block for feature expansion followed by a shuffling operation to efficiently rearrange the features. This method distinguishes itself by utilizing transformers to capture spatial relationships at multiple scales, contrasting with the multi-branch MLPs and GCNs used in earlier approaches such as MLPShuffle and NodeShuffle.\par
\textbf{Multi-CNN.} Yu et al. \cite{yu2018pu} proposed a sub-pixel convolution layer which begins with the application of multiple sets of $1 \times 1 $ convolutions on the embedded features of the input points, as shown in Fig. \ref{multi-cnn}. These convolutions serve to transform the original feature dimensions into a larger, more enriched feature space. Following this transformation, a reshape operation is employed to adjust the dimensions of the resulting feature tensor. \par
\paragraph{\textbf{Point Set Generation}} This process aims to generate upsampled points by utilizing expanded features or through the integration of original points with expanded features.\par


\begin{figure*}[t]
   \centering
   \begin{subfigure}{0.49\textwidth}
   \centering
   \includegraphics[width=\linewidth]{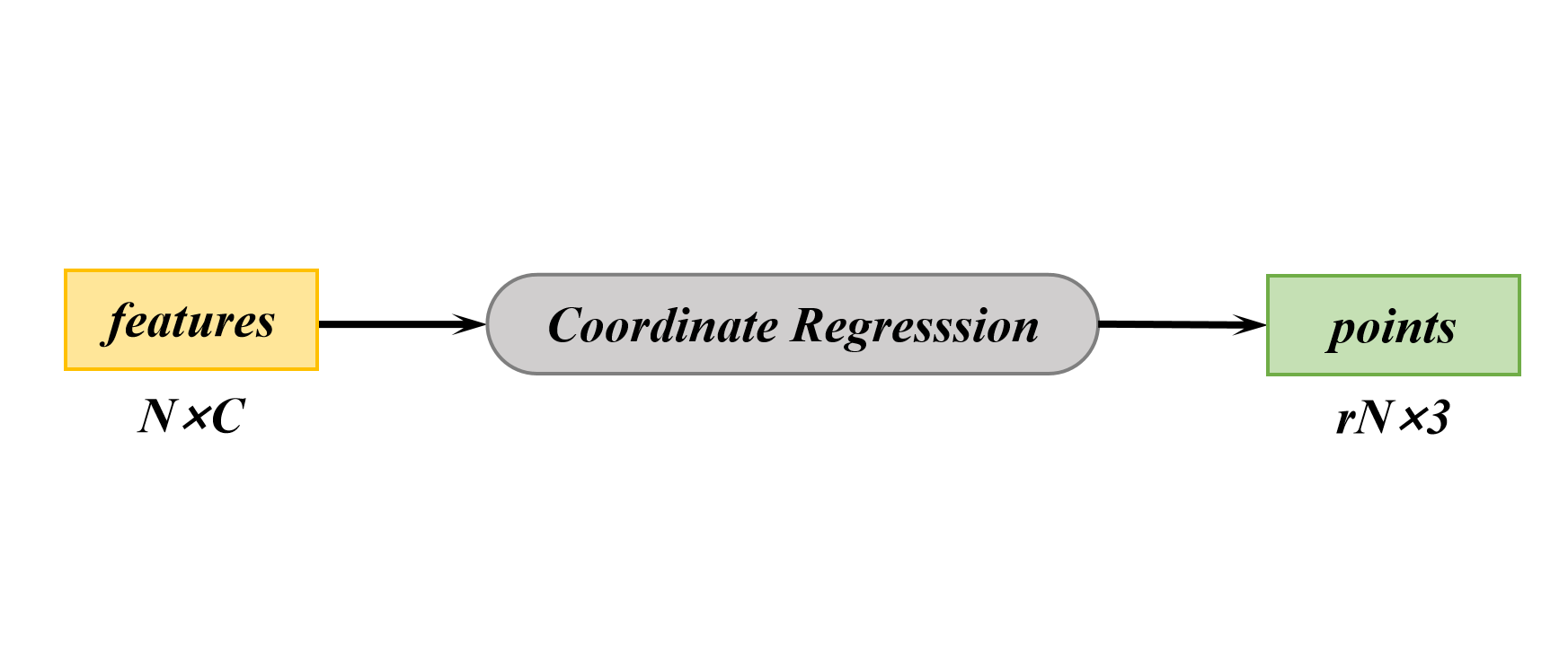}
   \vspace{-1.5cm}
   \caption{Regression to 3D coordinates}
   \label{direct}
   \end{subfigure}
   \vspace{1pt}
   \begin{subfigure}{0.49\textwidth}
   \centering
   \includegraphics[width=\linewidth]{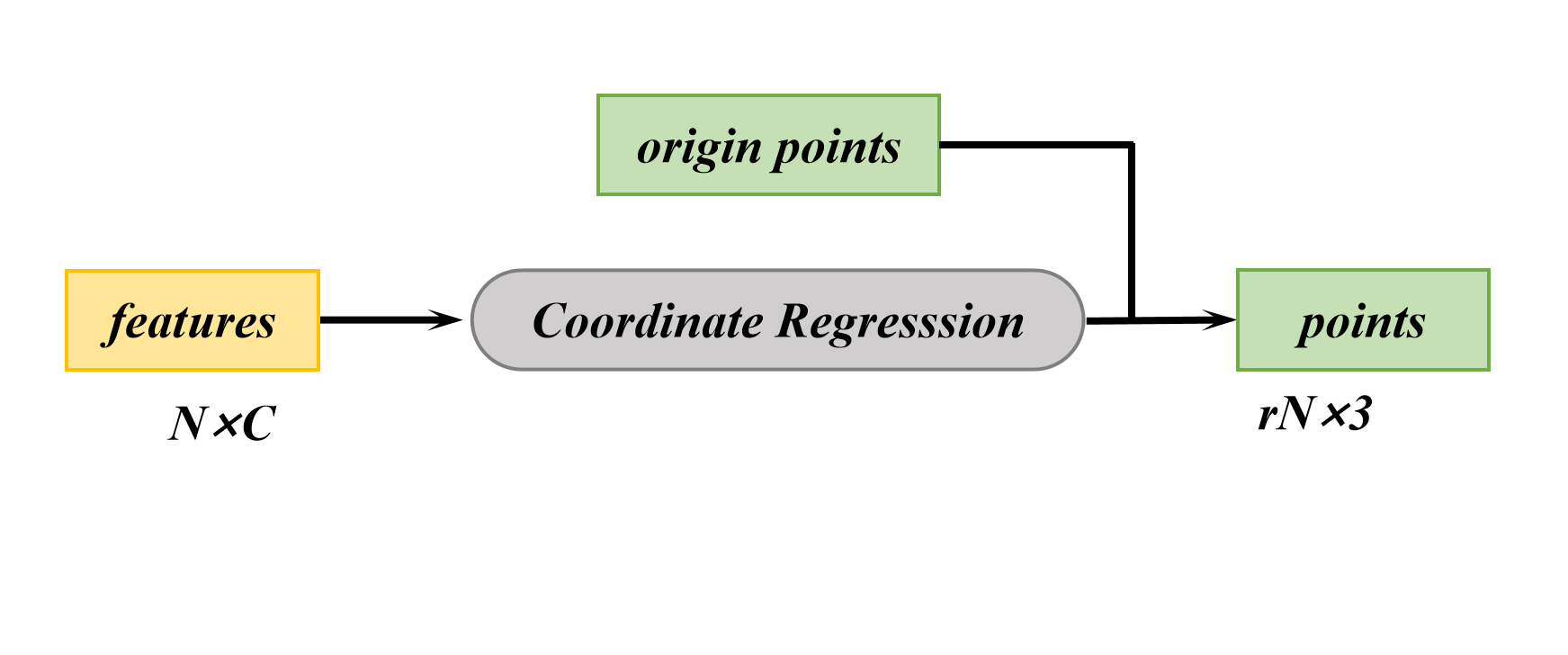}
   \vspace{-1.5cm}
   \caption{Regression to 3D coordinates offsets}
   \label{delta}
   \end{subfigure}
   \begin{subfigure}{0.49\textwidth}
   \centering
   \includegraphics[width=\linewidth]{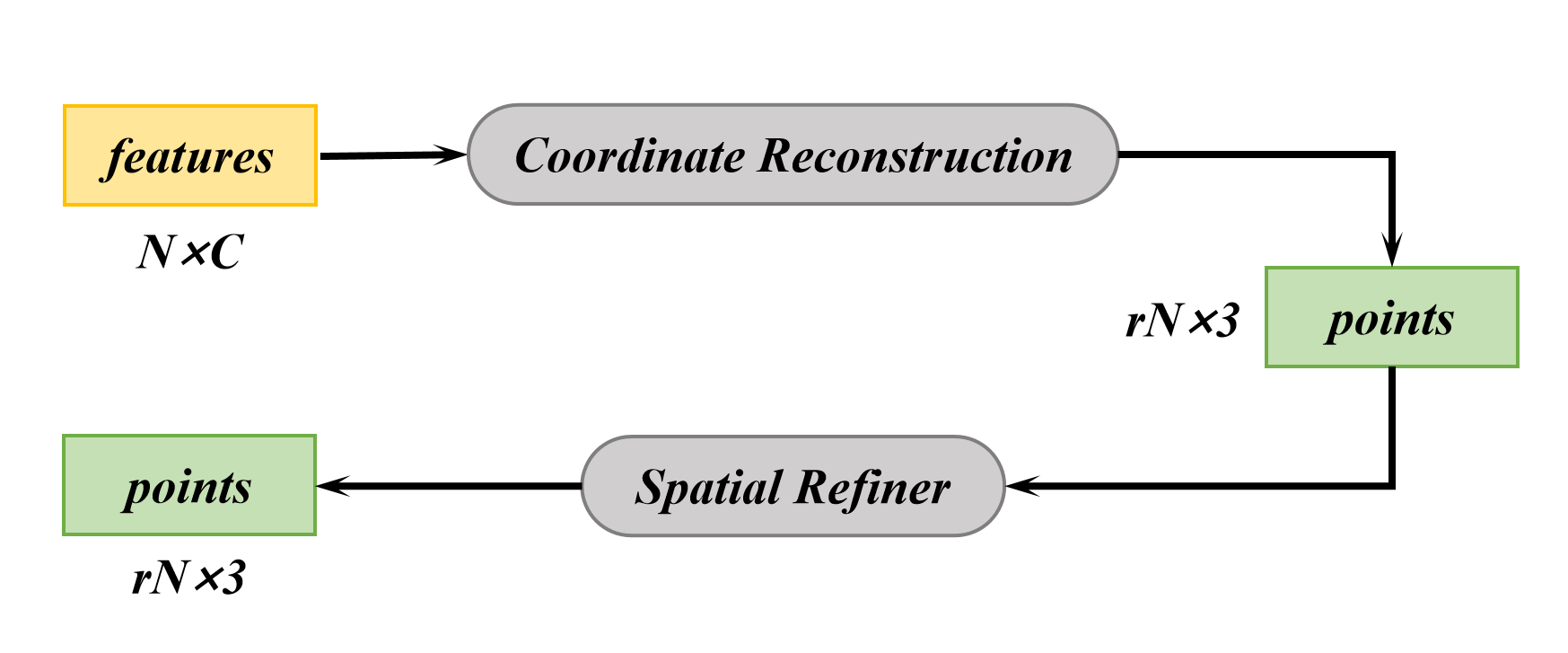}
   \caption{Disentangled refinement}
   \label{disentangle}
   \end{subfigure}
   \begin{subfigure}{0.49\textwidth}
   \centering
   \includegraphics[width=\linewidth]{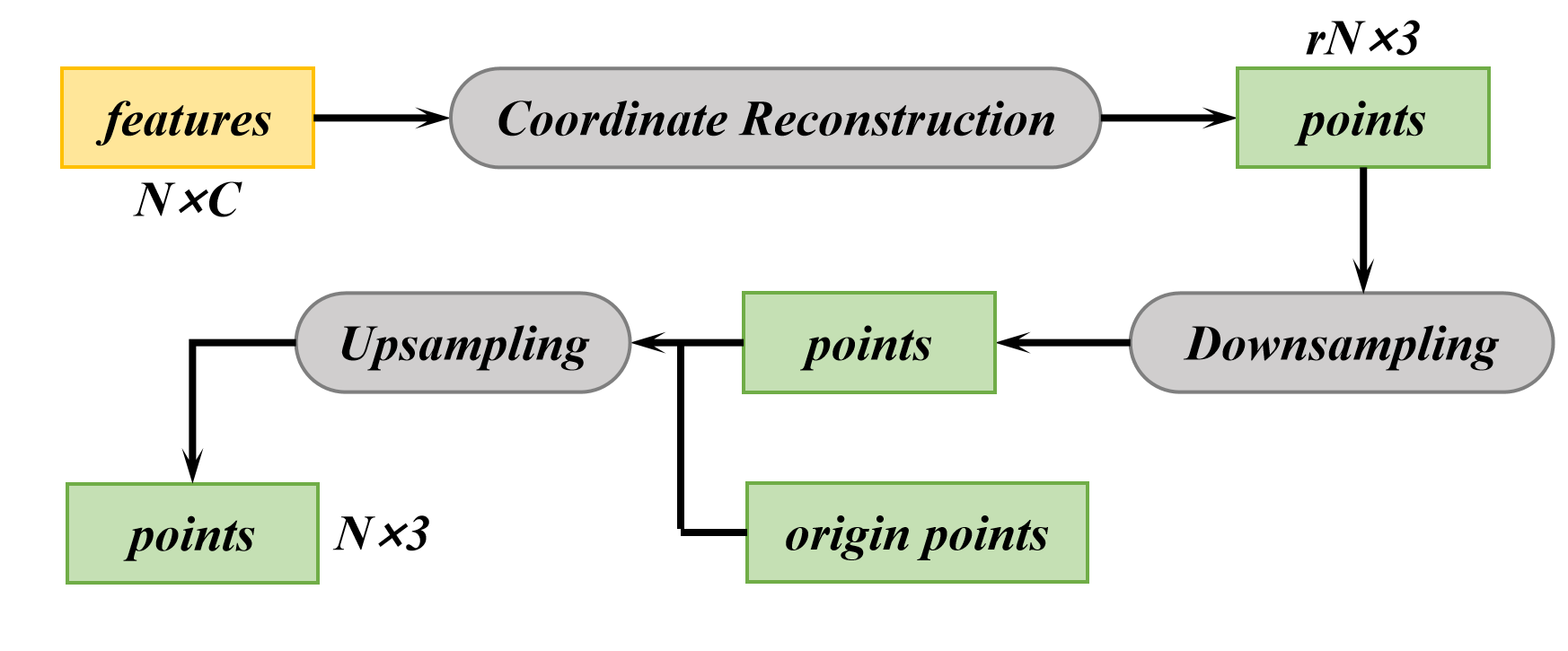}
   \caption{Up-down upsampling}
   \label{updown}
   \end{subfigure}
   \begin{subfigure}{0.49\textwidth}
   \centering
   \includegraphics[width=\linewidth]{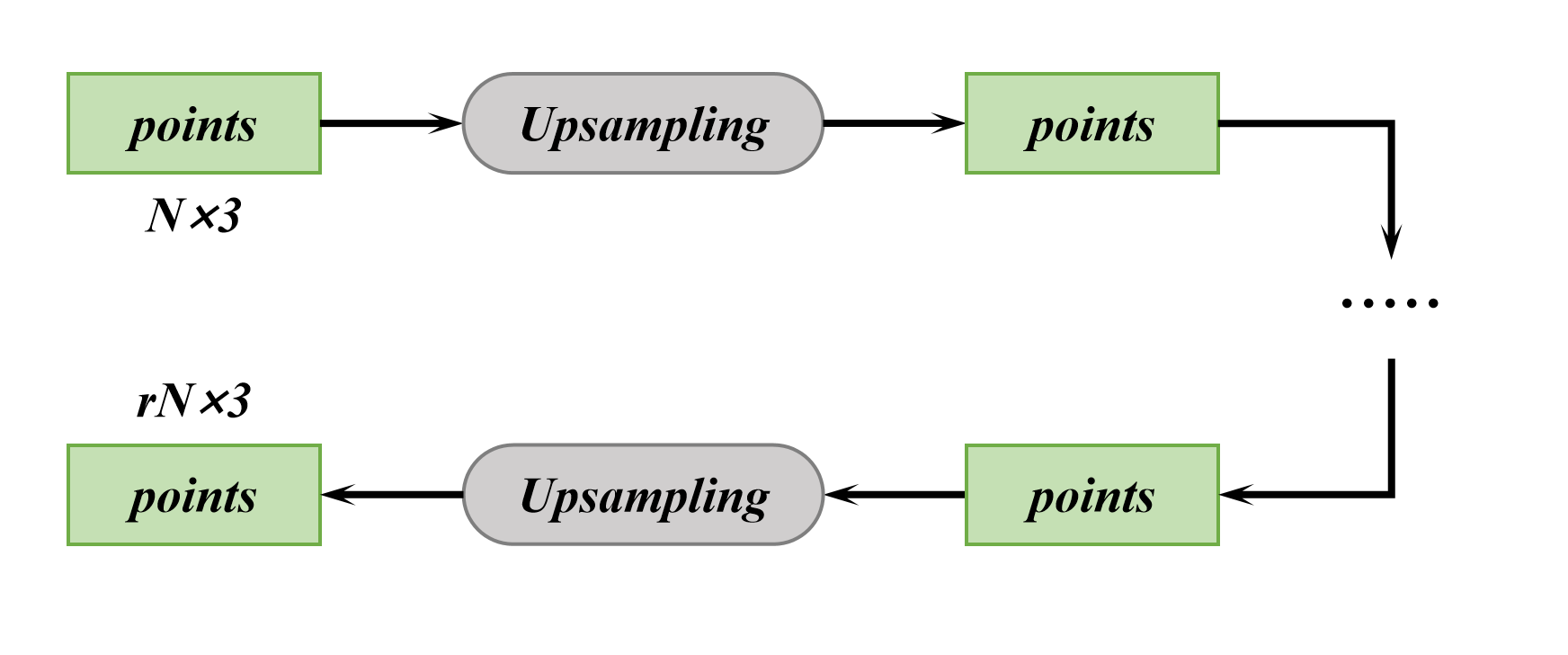}
   \caption{Multi-step upsampling}
   \label{multistep}
   \end{subfigure}
   \begin{subfigure}{0.49\textwidth}
   \centering
   \includegraphics[width=\linewidth]{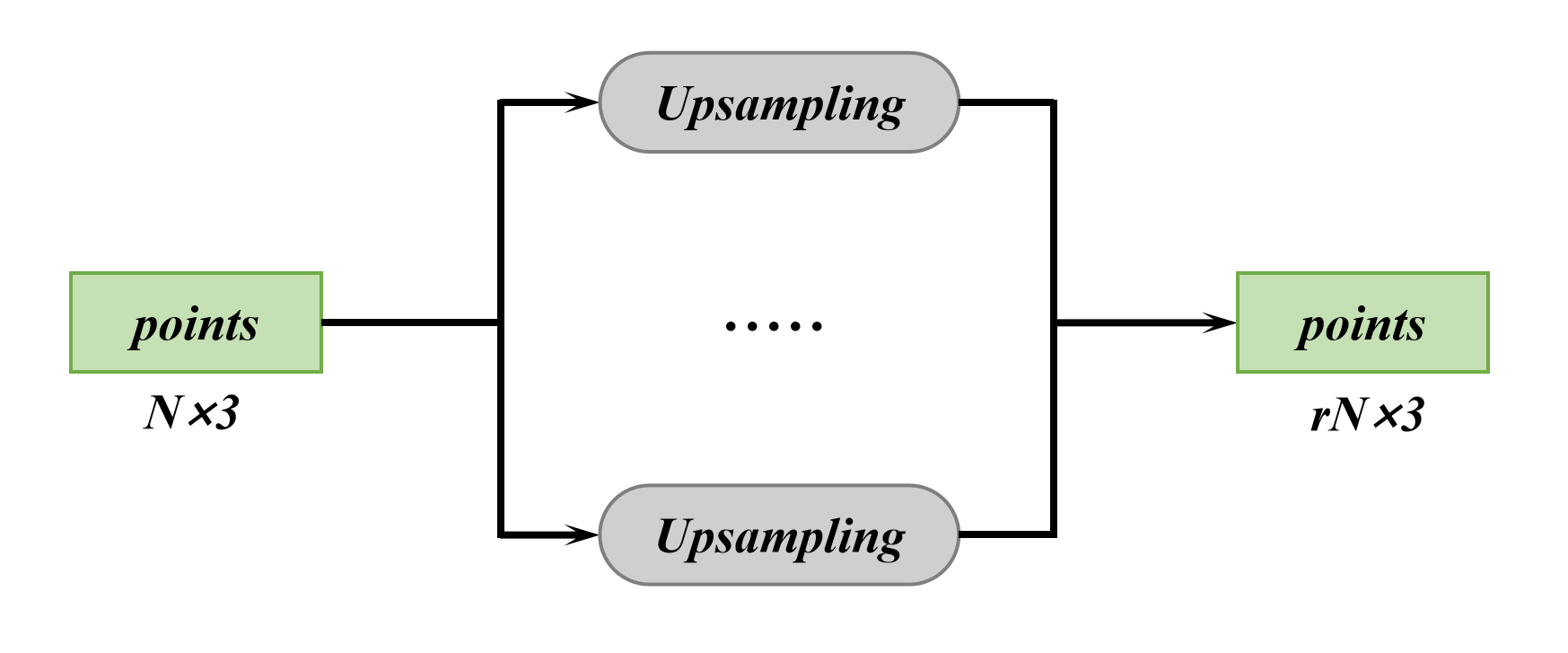}
   \caption{Multi-branch upsampling}
   \label{multibranch}
   \end{subfigure}
   \caption{Illustration of point set generation. (a) Direct regression. (b) Regression to 3D coordinate offsets. (c) Disentangled refinement. (d) Up-Down upsampling. (e) Multi-step upsampling. (f) Multi-branch upsampling. Here, \textit{N} denotes the number of points, \textit{C} denotes the dimensionality of the learned features for each point, and \textit{r} denotes the upsampling rate.}
\end{figure*}
\textbf{Coordinate regression.} Coordinate regression in 3D point cloud upsampling can be categorized into two primary approaches.\par
The first is direct regression, where features are directly mapped to 3D coordinates, enhancing geometric detail without altering the underlying structure \cite{yu2018pu,zeng2021point,qian2021pu,ding2021point}, as is shown in Fig. \ref{direct}.\par
The second is the regression to 3D coordinate offsets, which adjusts the positions of points by predicting their deviations from original locations. This approach allows for refined adjustments, particularly around edges and surfaces, thereby improving the realism and quality of the point cloud \cite{zhao2021sspu}, as shown in Fig. \ref{delta}.\par
\textbf{Disentangled refinement.} Li et al. \cite{li2021point} introduced Dis-PU, a network model that enhances point cloud upsampling by segmenting the process into two distinct phases, i.e., feature expansion and spatial refinement, as depicted in Fig. \ref{disentangle}. The architecture of the network is illustrated in Fig. \ref{classical}d. In the feature expansion phase, the network expands initial features to generate a coarse point set. The spatial refinement phase then adjusts these points to refine their spatial locations, resulting in a uniformly distributed and high-quality dense point set. Dis-PU enables more focused optimization and improved synergy in generating detailed point clouds.\par
Furthermore, SSPU \cite{wang2023sspu} utilizes a dilation strategy to create neighboring point sets at various scales and employs a DenseGCN architecture with three layers to capture local patterns effectively. The features from different scales are concatenated and fused, with attentional feature fusion redistributing these features to enhance the point cloud's representation significantly.\par
\textbf{Up-down upsampling.} Li et al. \cite{li2019pu} introduced a method for refining point cloud features through a sequence of upsampling and downsampling processes, as shown in Fig. \ref{updown}. Initially, the point features are upsampled to expand the feature set, which is then downsampled to extract distinct feature differences. These differences are later upscaled and merged back with the original expanded features for enhancement.\par
In a similar vein, Zhang et al. \cite{zhang2021progressive} employed an upsampling technique coupled with a channel attention mechanism. It is a two-stage downsampling process. The first stage maintains the original point count to extract and connect features from neighboring points, while the second selects crucial points and features for enhanced processing.\par
\textbf{Multi-step upsampling.} Wang et al. \cite{yifan2019patch} utilized a patch-based methodology that selectively enhances specific sections of the point cloud during each phase of upsampling, as depicted in Fig. \ref{multistep}. The architecture of the network is illustrated in Fig. \ref{classical}e.\par
Besides, Shi et al. \cite{shi2023progressive} started with a coarse point cloud, segmenting it into multiple parts that are individually refined through layers that integrate local, global, and point-wise features. Du et al. \cite{du2022point} described a cascaded refinement network that increases point density and precision in three stages, each designed with specific objectives within a unified network architecture. Other researchers, such as \cite{wu2019point} and \cite{li2021point}, have adopted similar strategies.\par
A common limitation in existing methods is that they mainly use feed-forward cascaded networks, and focus only on a coarse-to-fine approach. Wang et al. \cite{wang2023cfnet} introduced the cascaded feedback network, which incorporates a feedback mechanism to improve low-level feature learning by integrating higher-level layer information, combining feed-forward and feedback processes for enhanced feature refinement.\par
\textbf{Multi-branch upsampling.} Yu et al. \cite{yu2018pu} introduced a novel approach involving a multi-branch feature expansion module that employs parallel sub-pixel convolutional layers for processing, as shown in Fig. \ref{multibranch}.\par
Many existing methods aggregate hierarchical features from multiple resolutions through upsampling sub-networks, and thus suffer from high computational costs. To alleviate this issue,  Bai et al. \cite{bai2022bims} developed BIMS-PU, a network that integrates a bi-directional multi-scale upsampling module into a feature pyramid architecture. This network decomposes the upsampling and downsampling tasks into manageable sub-steps with smaller scaling factors, enabling parallel enhancement of multi-scale upsampled point cloud features.
Moreover, Liu et al. \cite{liu2023arbitrary} proposed a method that segments the input point cloud into several sub-sparse point clouds, each of which is then upsampled separately to enhance density and detail. The method includes a resampling process that integrates these densified sub-sparse point clouds into a cohesive structure.\par
\textbf{Local filling.} Liu et al. \cite{liu2024pu} proposed a virtual mask-based network called PU-Mask, viewing point cloud upsampling as a ``local filling'' problem. They introduced a virtual mask generation module to locate and form each virtual mask. Following this, a mask-guided transformer-style asymmetric auto-encoder (MTAA) was used to restore the upsampled features. By employing a pooling technique, PU-Mask produces a coarse upsampled point cloud specific to the virtual masks. Finally, a learnable pseudo Laplacian operator was designed to refine the coarse upsampled point cloud and generate a more accurate upsampled point cloud.
\subsubsection{\textbf{Tristep-extended Upsampling Networks}} These networks are structured on the tristep upsampling framework. By integrating tristep upsampling in a shell, such as GAN, the extended versions substantially refine the network's ability to generate high-resolution points.\par
\textbf{Generative adversarial network.} A number of works have adopted generative adversarial networks to enhance the density and quality of point clouds, as shown in Fig. \ref{classical}b. Within this framework, the generator often embodies a structure analogous to a tristep upsampling network. The discriminator components of these GANs exhibit a diverse array of architectures.\par
Li et al. \cite{li2019pu} enhanced feature learning in neural networks by incorporating a self-attention mechanism into the architecture initially proposed by Yuan et al. \cite{yuan2018pcn}. This excels in integrating local and global information efficiently. The work in \cite{li2021cm} further develops this concept by designing a discriminator with two layers of feature extraction. It focuses on detailed analysis of both local and global features through a combination of nonlinear convolutional layers and a feature connection layer, thus setting the stage for advanced feature discrimination.\par
Lv et al. \cite{lv2021point} introduced the PointNet-Mix discriminator that leverages both max-pooling and average pooling features to achieve permutation invariance in the extracted features. Moreover, the discriminator proposed by Liu et al. \cite{liu2022pu} employs a sequence of four MLPs followed by a max pooling operation to derive a global response vector. It is then recombined with the MLP outputs to enrich feature representation. This setup is further refined with a self-attention mechanism to enhance feature interactions.\par
Additionally, Liu et al. \cite{liu2022pufa} developed a discriminator featuring dual heads. One for global features and the other for high-frequency (HF) features. The dual heads share a common backbone but differentiated by a frequency-aware design using a graph filter. This filter emphasizes high-frequency components, aiding in the isolation of HF points that highlight complex details such as edges and corners. Besides, Nguyen et al. \cite{nguyen2022combination} attempted to learn mutli-modal feature from low-resolution point clouds and corresponding 2D images to enhance point cloud quality.\par
\textbf{Siamese network.} Inspired by the Siamese networks, Han et al. \cite{han2023upu} introduced UPU-SNet, which employs dual weight-sharing branches to extract and process features from raw point clouds. One branch focuses on feature extraction and standard coordinate reconstruction, while the other incorporates a shared feature extractor and a novel feature expander, culminating in a shared coordinate reconstruction process.\par
\textbf{Vector-quantized codebook learning.} Zhao et al. \cite{zhao2023cpu} introduced CPU, which utilizes vector quantization and a knowledge distillation training scheme. This framework reformulates the upsampling task as a more deterministic code prediction task within a discrete proxy space. It leverages a teacher-student model for learning high-fidelity geometric data in a codebook, and employs a transformer for precise code prediction.
\subsubsection{\textbf{Others}}
Sun et al. \cite{sun2020pointgrow} introduced an auto-regressive framework PointGrow, which employs a probabilistic approach to model these correlations. This method treats the generation of a point cloud as a sampling process in a high-dimensional space. By leveraging the chain rule, it decomposes the joint probability distribution of the point cloud into a series of conditional probabilities. Zhang et al. \cite{zhang2022upsampling} introduced an upsampling auto-encoder (UAE). UAE operates by randomly subsampling points from the input point cloud at a low rate, then utilizing an asymmetric encoder-decoder architecture to reconstruct the unselected points. Specifically, the encoder only deals with the subsampled points, while the decoder focuses on upsampling. Mao et al. \cite{mao2022pu} presented PU-Flow, which integrates normalizing flows and a weight prediction technique. This approach capitalizes on the reversible properties of normalizing flows to facilitate seamless transformations between Euclidean and latent spaces. This transformation enables the upsampling process to be conceptualized as an ensemble of neighboring points within the latent space, with ensemble weights determined adaptively based on the local geometric context. Ye et al. \cite{ye2021meta} introduced Meta-PU via a meta-learning framework. Central to this architecture is a meta-subnetwork designed to dynamically adjust the weights of a meta-RGC block based on input scale factors.\par
\begin{table*}[t]
      \centering
      \caption{Comparative point cloud upsampling results on the PU1K dataset $(r \in [4])$ and PU-GAN dataset $(r \in [4,16])$. The symbol `-' means the result is unavailable. The CD, HD and P2F values are multiplied by $10^{3}$. The low-resolution point clouds for evaluating contain 2,048 points except CFNet on PU-GAN (1,024), CPU on PU-GAN (1,024), AR-GCN on PU-GAN (4,096) and ZSPU on PU-GAN (4,096). Some of the special settings for some specific methods all follow the best choice.}
      \vspace{-0.2cm}
      \resizebox*{\textwidth}{!}{
      \begin{tabular}{|c|c|c|c|c|c|c|c|c|c|c|c|c|c|}
      \hline
      \multicolumn{3}{|c|}{\multirow{3}[0]{*}{Method}} & \multirow{3}[0]{*}{Arbitrary?} & \multirow{3}[0]{*}{Supervised?} & \multicolumn{3}{c|}{PU1K Dataset} & \multicolumn{6}{c|}{PU-GAN Dataset} \\
      \cline{6-14}
      \multicolumn{3}{|c|}{}  &       &       & \multicolumn{3}{c|}{4$\times$} & \multicolumn{3}{c|}{4$\times$} & \multicolumn{3}{c|}{16$\times$} \\
      \cline{6-14}
      \multicolumn{3}{|c|}{}  &       &       & CD$\downarrow$    & HD$\downarrow$    & P2F$\downarrow$   & CD$\downarrow$    & HD$\downarrow$    & P2F$\downarrow$   & CD$\downarrow$    & HD$\downarrow$    & P2F$\downarrow$ \\
      \hline
      \multirow{32}[0]{*}{Embedding Space Upsampling} & \multirow{26}[0]{*}{Tristep Upsampling} & PU-Net \cite{yu2018pu} & $\times$   & \checkmark   & 1.155 & 15.170 & 4.834 & 0.720  & 8.940  & 6.840  & 0.510  & 8.206 & 6.041 \\
      \cline{3-14}
              &       & MPU \cite{yifan2019patch}   & $\times$    & \checkmark   & 0.935 & 13.327 & 3.511 & 0.490  & 6.110  & 3.960  & 0.219 & 7.054 & 3.085 \\
              \cline{3-14}
              &       & PU-GCN \cite{qian2021pu} & $\times$    & \checkmark   & 0.669 & 10.087 & 2.706 & 0.260  & 1.890  & 2.720  & 0.161 & 4.283 & 2.632 \\
              \cline{3-14}
              &       & Dis-PU \cite{li2021point} & $\times$    & \checkmark   & 0.485 & 6.145 & 1.802 & 0.320  & 4.210  & 4.150  & 0.167 & 4.923 & 2.261 \\
              \cline{3-14}
              &       & PU-Transformer \cite{qiu2022pu} & $\times$    & \checkmark   & 0.451 & 3.843 & 1.277 & 0.270  & 2.610  & 1.840  & 0.241 & 2.310  & 1.687 \\
              \cline{3-14}
              &       & EAR \cite{yu2018ec}   & $\times$    & $\times$   & -     & -     & -     & 0.520  & 7.370  & 5.820  & -     & -     & - \\
              \cline{3-14}
              &       & SPU-Net \cite{liu2022spu} & $\times$    & $\times$    & -     & -     & -     & 0.380  & 2.240  & 5.970  & -     & -     & - \\
              \cline{3-14}
              &       & $S^{3}$U-PVNet \cite{han2024s3u} & \checkmark   & $\times$    & -     & -     & -     & 0.280  & 1.640  & 2.530 & 0.121 & 1.525 & 2.404 \\
              \cline{3-14}
              &       & SSPU-Net \cite{zhao2021sspu} & $\times$    & $\times$    & -     & -     & -     & 0.370  & 3.470  & 4.430  & -     & -     & - \\
              \cline{3-14}
              &       & P$C^{2}$-PU \cite{long2022pc2} & $\times$    & \checkmark   & -     & -     & -     & 0.255 & 2.504 & 2.119 & 0.111 & 2.806 & 2.351 \\
              \cline{3-14}
              &       & PU-CTG \cite{li2024pu} & $\times$    & \checkmark   & 0.690  & 10.500  & -     & 0.240  & 3.450  & 2.360  & -     & -     & - \\
              \cline{3-14}
              &       & PU-GACNet \cite{han2022pu} & $\times$    & \checkmark   & 0.665 & 9.053 & 2.429 & -     & -     & -     & -     & -     & - \\
              \cline{3-14}
              &       & BIMS-PU \cite{bai2022bims} & $\times$    & \checkmark   & -     & -     & -     & 0.280  & 4.280  & 2.050  & 0.160  & 4.700   & 2.590 \\
              \cline{3-14}
              &       & SSPU-Net \cite{wang2023sspu} & $\times$    & \checkmark   & -     & -     & -     & -     & -     & -     & -     & -     & - \\
              \cline{3-14}
              &       & UPU-DGTNet \cite{deng2022upu} & $\times$    & $\times$    & -     & -     & -     & 1.050  & 5.620  & 3.170  & -     & -     & - \\
              \cline{3-14}
              &       & Refine-PU \cite{liu2022refine} & $\times$    & \checkmark   & 0.534 & 7.071 & 2.071 &   -   &   -   &   -   & -     & -     & - \\
              \cline{3-14}
              &       & PU-CRN \cite{du2022point} & $\times$    & \checkmark   & 0.471 & 7.123 & 1.925 & 0.520  & 6.102 & 3.165 & 0.284 & 7.143 & 3.724 \\
              \cline{3-14}
              &       & Flexible-PU \cite{qian2021deep} & $\times$    & \checkmark   & -     & -     & -     & 0.218 & 1.450  & 6.880  & 0.700   & 2.530  & 4.890 \\
              \cline{3-14}
              &       & APU-SMOG \cite{dell2022arbitrary} & $\times$    & \checkmark   & 0.528 & 2.549 & 1.667 & 0.276 & 1.909 & 2.634 & 0.149 & 1.948 & 2.666 \\
              \cline{3-14}
              &       & CFNet \cite{wang2023cfnet} & $\times$    & \checkmark   & 0.450  & 5.918 & -     & 0.497 & 4.676 & -     & -     & -     & - \\
              \cline{3-14}
              &       & PU-WGCN \cite{gu2022pu} & $\times$    & \checkmark   & 0.619 & 8.926 & 2.377 & -     & -     & -     & -     & -     & - \\
              \cline{3-14}
              &       & PU-MFA \cite{lee2022pu} & $\times$    & \checkmark   & -     & -     & -     & 0.233 & 1.094 & 2.545 & 0.501 & 5.414 & 9.111 \\
              
              \cline{3-14}
              &       & AR-GCN\cite{wu2019point} & $\times$ & \checkmark & - & - & - & 0.230 & 1.780 & 3.020 & - & - & - \\
              \cline{3-14}
              &       & ZSPU \cite{zhou2022zero} & $\times$    & $\times$    & -     & -     & -     & 0.200   & 1.100   & 1.890  & -     & -     & - \\
              \cline{3-14}
              &       & APUNet \cite{zhao2023apunet} & $\times$    & \checkmark   & -     & -     & -     & 0.228 & -     & -     & 0.760  & -     & - \\
              \cline{3-14}
              &      & PU-Mask
              \cite{liu2024pu} & $\times$ & \checkmark & 0.425 & 6.476 & 1.542 & 0.252 & 3.462 & 2.167 & - & - & - \\
              \cline{2-14}
              & \multirow{7}[0]{*}{Tristep-extended Upsampling} & PU-GAN \cite{li2019pu} & $\times$    & \checkmark   & 0.738 & 10.074 & 2.992 & 0.280  & 4.640  & 2.330  & 0.207 & 6.963 & 2.556 \\
              \cline{3-14}
              &       & PU-CycGAN \cite{li2022weakly} & $\times$    & \checkmark   & 0.551 & 2.919 & 2.080  & 0.320  & 2.920  & 3.350  & -     & -     & - \\
              \cline{3-14}
              &       & UPU-SNet \cite{han2023upu} & $\times$    & \checkmark   & -     & -     & -     & 0.218 & 1.210  & 6.910  & 0.850  & 1.290  & 5.040 \\
              \cline{3-14}
              &       & PU-Refiner \cite{liu2022pu} & $\times$    & \checkmark   & -     & -     & -     & 0.270  & 3.870  & 2.480  & -     & -     & - \\
              \cline{3-14}
              &       & PUFA-GAN \cite{liu2022pufa} & $\times$    & \checkmark   & 0.439 & 6.516 & 1.623 & 0.258 & 3.571 & 2.392 & -     & -     & - \\
              \cline{3-14}
              &       & CM-Net\cite{li2021cm} & $\times$    & \checkmark   & -     & -     & -     & 0.410  & 5.270  & 2.360  & -     & -     & - \\
              \cline{3-14}
              &       & CPU \cite{zhao2023cpu}   & $\times$    & \checkmark   & 0.358 & 2.818 & 1.113 & 0.503 & 3.987 & 2.874 & 0.213 & 3.927 & 3.344 \\
              \cline{2-14}
              & \multirow{2}[0]{*}{Others} & PU-Flow \cite{mao2022pu} & \checkmark   & \checkmark   & 0.563 & 0.321 & 1.315 & 0.820  & 5.280  & 8.970  & -     & -     & - \\
              \cline{3-14} &
              & {Meta-PU} \cite{ye2021meta} &
              \checkmark & $\times$ &
              - & - & - & - & - & - & - & - & - \\
              \hline
      \multirow{18}[0]{*}{3D Euclidean Space Upsampling} & \multirow{11}[0]{*}{Refinement-based} & Grad-PU \cite{he2023grad} & \checkmark   & \checkmark   & 0.404 & 3.732 & 1.474 & 0.250  & 2.370  & 1.890  & 0.108 & 2.352 & 2.127 \\
      \cline{3-14}
              &       & SAPCU \cite{zhao2022self} & \checkmark   & $\times$    & -     & -     & -     & 0.465 & 10.572 & 3.421 & 0.510  & 6.739 & 5.442 \\
              \cline{3-14}
              &       & PU-EVA \cite{luo2021pu} & \checkmark   & $\times$    & 0.649 & 8.870  & 2.715 & 0.571 & 5.840  & 3.937 & 0.342 & 8.140  & 4.473 \\
              \cline{3-14}
              &       & PUDM \cite{qu2023conditional}  & \checkmark   & \checkmark   & 0.217 & 2.164 & 1.477 & 0.131 & 1.220  & 1.912 & 0.082 & 1.120  & 2.114 \\
                \cline{3-14}
              &       & PU-Ray \cite{lim2023pu} & \checkmark   & $\times$    & 0.404 & 4.025 & 1.341 & 0.260  & 2.736 & 1.707 & 0.121 & 2.654 & 2.093 \\
              \cline{3-14}
              &       & ND-PUFlow \cite{hu2023noising} & \checkmark   & $\times$    & -     & -     & -     & 0.212 & 0.940  & 6.080  & 0.660  & 0.790  & 4.140 \\
              \cline{3-14}
              &       & PUGeo-Net \cite{qian2020pugeo} & $\times$    & \checkmark   & -     & -     & -     & 0.187 & 0.278 & 1.568 & 0.111 & 0.296 & 1.770 \\
              \cline{3-14}
              &      & iPUNet \cite{wei2023ipunet} & $\times$   & $\times$   & -   &
              -     & -    & -  &-   & -   & -   &-  & -  \\
              \cline{3-14}
              &      & PUSS-AS \cite{zhao2023self} &\checkmark & $\times$& - & - & -
              & 0.251 & 3.491 & 2.163
              & 0.153 & 3.696 & 2.513 \\
              \cline{3-14}
              &       & SPU-PMD \cite{liu2024pu} & $\times$ & $\times$ & 0.544 & 4.926 & 1.861 & 0.318 & 3.317 & 2.508 & - & - & - \\
              \cline{3-14}
              &       & RepKPU \cite{rong2024repkpu} & $\times$ & \checkmark   & 0.327 & 0.268 & 0.938 & 0.248 & 2.880 & 1.906 & 0.107 & 3.345 & 2.068 \\
              \cline{2-14}
              & \multicolumn{1}{c|}{\multirow{7}[0]{*}{Surface-reconstruction-based}} & NePs\cite{feng2022neural}  & \checkmark   & \checkmark   & -     & -     & -     & 0.260  & 3.650  & 1.940  & 0.152 & 4.910  & 2.198 \\
              \cline{3-14}
              &       & APU-LDI \cite{li2024learning}& \checkmark   & $\times$    & 0.371 & 3.197 & 1.111 & 0.232 & 1.675 & 1.338 & 0.092 & 1.504 & 1.544 \\
              \cline{3-14}
              &       & PU-VoxelNet\cite{du2024arbitrary} & \checkmark   & \checkmark   & 0.338 & 2.694 & 1.183 & 0.233 & 1.751 & 2.137 & 0.091 & 1.726 & 2.301 \\
              \cline{3-14}
              &       & ASUR3D \cite{kumbar2023asur3d} & \checkmark   & $\times$    & 0.398 & 3.609 & 1.525 & 0.238 & 2.303 & 1.745 & 0.103 & 2.275 & 2.088 \\
              \cline{3-14}
              &       & TP-NoDe \cite{kumbar2023tp} & $\times$    & \checkmark   & -     & -     & -     & 0.330  & 3.490  & 5.210  & -     & -     & - \\
              \cline{3-14}
              &       & PSCU \cite{cai2023parametric} & \checkmark   & $\times$    & 0.886 & -     & 1.091 & -     & -     & -     & -     & -     & - \\
              \cline{3-14}
              &       & PU-SDF \cite{pan2024pu} & \checkmark & \checkmark & - & - & - & - & - & - & - & - & - \\
              \hline
        \end{tabular}}
        \vspace{-0.2cm}
      \label{upsamplingtable}
    \end{table*}
\subsection{3D Euclidean Space Upsampling}
Instead of manipulating feature vectors in a latent space, 3D Euclidean space upsampling techniques enhance point cloud resolution by directly adding more points in the Euclidean space, representing a fundamentally different approach. Given the nature of these methods, our focus will be more on the overall workflow rather than the specifics of individual modules.\par
These methods can generally be divided into two main categories: refinement-based and surface-reconstruction-based.
\subsubsection{\textbf{Refinement-based}}
Refinement-based methods are utilized to enhance the density of point clouds in the Euclidean space by augmenting existing points. These methods, however, often generate points with imprecise locations, necessitating further refinement to achieve a high-resolution point cloud.\par
\textbf{Differential geometry formulation.} One pioneering approach, PUgeo-Net by Qian et al. \cite{qian2020pugeo}, integrates differential geometry with deep learning. PUgeo-Net aims to learn a local parameterization for each point, employing an augmented Jacobian matrix to achieve a linear transformation, as shown in Fig. \ref{classical}f. This transformation facilitates the movement of sampled points from a 2D parametric domain to a 3D tangent space, and subsequently onto a 3D curved surface through the estimation of normal and coordinate offsets. On  the contrary, MFU, another method detailed by Qian et al. \cite{qian2021deep}, is parameterization-free and instead uses Taylor’s expansion to decompose the algebraic representation of an upsampled point into a linear approximation and higher-order errors. This technique predicts an interpolation matrix that upsamples both the point set and extracted features, followed by a refinement process based on these features. Similarly, PU-EVA \cite{luo2021pu} almost replicates the formulation used in \cite{qian2021deep}, introducing an Edge-Vector based Approximation upsampling unit. In this unit, edge vectors are calculated and mapped into high-dimensional feature spaces to produce a matrix similar to that in MFU for point set upsampling. Feature-based refinement follows this upsampling process, further refining the point set.\par
\textbf{Interpolating and updating.} Considering the natural uniformity of voxel grid center points, Zhao et al. \cite{zhao2022self} proposed SAPCU. It utilizes voxel-based point sampling and self-supervised learning-based projection to accomplish the entire upsampling process. Seed points are sampled from voxel volumes, after which two self-supervised implicit neural functions are employed to estimate directions and distances towards underlying surface to project seed points. Due to the failure of this method in scenarios such as narrow surface gaps, Zhao et al. \cite{zhao2023self} proposed PUSS-AS, which incorporates projection rectification step to improve the performance of previous method. As a trade-off, despite the implementation of acceleration strategies, the inference latency of this method remains notably protracted.\par
Following a similar pipeline, Grad-PU \cite{he2023grad} first utilizes midpoint interpolation for upsampling, subsequently followed by a refinement phase to enhance the interpolated point cloud. Different from other methods that rely on point-to-surface distances, Grad-PU uses the point-to-point (P2P) distance, which represents the Euclidean distance between a point and its nearest neighbor point in the ground-truth dense point cloud, rendering Grad-PU as a supervised method. \par
\textbf{Noising sampling and denoising.} TP-NoDe \cite{kumbar2023tp} takes score-based denoising \cite{luo2021score} as the backend. It first employs density-aware \textit{k}-nearest neighbour (DA-\textit{k}NN) to extract local neighbour points in a topology-aware way. The extracted points are perturbed and concatenated back to where they are extracted. ND-PUFlow \cite{hu2023noising} applies a noising-denoising pipeline as well. In detail, a higher dimension Gaussian noise is injected into the input sparse point cloud first, and surface projection continuous normalizing flows (SP-CNFs) module built on flow-based model is designed to denoise the noisy yet denser point cloud.\par
\textbf{Others.} As aforementioned methods follow similar pipelines, the following methods introduce new ideas borrowed from computer graphics or popular generative models.\par
Qu et al. \cite{qu2023conditional} also introduced denoising duffusion probalistic model to tackle the issue of point cloud upsampling. It contains two similar networks, i.e., the conditional network (C-Net) and the noise network (N-Net). C-Net takes the sparse point cloud as input and output, while N-Net deals with interpolated one. Conditions are generated from C-Net and interact with N-Net via a transfer module. Besides encoding time step \textit{t}, rate prompt is also encoded into N-Net to enhance its performance when arbitrary ratios are selected.\par
Since N-rotational symmetric (N-RoSy) fields are commonly used in computer graphics, from a disparate viewpoint, Wei et al. \cite{wei2023ipunet} introduced cross field, also called 4-RoSy field, for point cloud upsampling. It performs as a way of local parameterization since a cross field has 4 status respectively corresponding to $\textit{k}\pi/{2} (\textit{k}=1,...,4)$. To achieve feature awareness, cross fields aligned with sharp features are learned in a self-supervised mode to express the local topology structures such as edges. \par
Derived from sphere tracing \cite{hart1996sphere}, PU-Ray \cite{lim2023pu} adopts a ray-marching process, which is a standard technique in computer graphics. Each query point reaches the target surface by multiple steps, each of which is dynamically adjusted by the closest distance from the point to the surface. In fact, PU-Ray obtains the distance via an implicit surface, i.e., distance functions implemented by MLP. To enhance sphere tracing, another MLP is trained to estimate a mere offset towards target surface in the end.\par
Liu et al. \cite{liu2024spu} considered point cloud upsampling as generating dense point clouds from the viewpoint of deformable topology and proposed a self-supervised network SPU-PMD. SPU-PMD is formulated by a series of coarse mesh interpolator and mesh deformers. At each stage, the mesh interpolator first produces the initial dense point clouds via mesh interpolation. Meanwhile, the deformer infers the morphing by estimating the movements of mesh nodes and reconstructs the topology structure. By associating mesh deformation with feature expansion, this module can progressively refine point clouds’ surface uniformity and entire structure. \par
Inspired by KPConv \cite{thomas2019kpconv}, Rong et al. \cite{rong2024repkpu} presented RepKPU. In this work, a novel paradigm named as Kernel-to-Displacement generation was proposed, and point cloud upsampling was reformulated as the deformation of kernel points. 
Moreover, KP-Queries was introduced, which is a set of kernel points with predefined positions and learned features, serving as the initial state of upsampling. RepKPoints interacts with KP-Queries via cross-attention mechanisms, and subsequently KP-Queries are converted to displacement features, followed by an MLP to predict the displacements of KP-Queries to obtain the generated points. \par
\subsubsection{\textbf{Surface-reconstruction-based}}
From a different perspective, point cloud upsampling can be accomplished by sampling points on the underlying surface. Surface-reconstruction-based methods embark from the idea that completes surface reconstruction on sparse point cloud first, and then obtains more points based on the reconstructed explicit/implicit surface.\par
\textbf{Parametric-surface-based.} They usually first formulate an explicit parametric function, and then fit the underlying surface.\par
Cai et al. \cite{cai2023parametric} introduced a parametric-surface-constrained upsampler called PSCU, formulating an explicit polynomial surface function. Specifically, a bicubic function is formulated to model the local underlying surface, and an additional rotation matrix is multiplied onto the local coordinates to describe surface in the global frame. Thus, PSCU predicts the coefficient of bicubic function and a 6D representation \cite{zhou2019continuity} that can be decoded into a rotation matrix to ensure the properties of {$SO(3)$}. A peculiarity of PSCU is that points are reconstructed under constrained by the parametric surface, and features are only used to predict a relative displacement.\par
\textbf{Implicit-surface-based.} These methods rely on implicit representations of surface, such as occupancy networks and neural fields.\par
ASUR3D \cite{kumbar2023asur3d} tries to employ an implicit occupancy representation. The occupancy representation converts surface representation into a classification problem, in which a point that is closer to surface always gets a higher occupancy score. After establishing the voxel grid with occupancy scores, the output point cloud is acquired by applying the marching cubes algorithm.
Noticing the contiguity of surface, Feng et al. \cite{feng2022neural} proposed a point representation named neural points, introducing a discrete-to-continuous strategy. 
Adopting implicit neural field as well, Li et al. \cite{li2024learning} introduced APU-LDI. With a pretrained local distance indicator indicating surface implicitly, a global implicit field is learned to represent the entire point cloud. During inference time, a query point would directly result in the generation of a new point.\par
Signed distance function (SDF) is also a commonly-used way to represent implicit surface. Pan et al. \cite{pan2024pu} proposed a network called PU-SDF, which consists of a local feature-based signed distance function network (LSDF-network) and a 3D-grid query points generation module (GPGM). GPGM can generate uniform and unlimited query points in sparse voxel space with arbitrary resolution, and the LSDF-network utilizing the SDF-network proposed in GenSDF \cite{chou2022gensdf} can perform point cloud upsampling at arbitrary rates based on 3D-grid features and query points.\par
Remarkably, the main difference between refinement-based approaches and surface-reconstruction-based approaches is that the former ones rely on the intrinsic perception of the underlying surface, but the latter ones directly reconstruct the surface.
\subsection{Summary}
The primary characteristics of current upsampling methods are summarized as follows:
\begin{enumerate}
    \item \textbf{Embedding space upsampling methods}. They use end-to-end networks to expand feature vectors, converting a point set into a high-resolution point cloud. Although convenient for training and inference, these methods often tie the upsampling ratio to the network structure, limiting them to fixed or incremental ratio upsampling. 
    \item \textbf{3D Euclidean space upsampling methods}. On the contrary, they primarily conduct several processing steps and some steps might not be incorporated in neural network. However, by separating the upsampling ratio from the network architecture, they facilitate upsampling at arbitrary ratios.
    \item \textbf{Synthetic benchmarks are still dominating}. Current upsampling methods tend to focus heavily on synthetic data, which is generally easier to process than real-world raw data. The construction of real-world benchmarks as well as a more comprehensive evaluation of existing methods on them are urgent.
    \item \textbf{Unsupervised upsampling methods yield competitive performance}. Most upsampling networks are trained in a fully supervised manner, requiring high-resolution point clouds as labels. Encouragingly, many self-supervised methods, such as SPU-Net \cite{liu2022spu} and PUSS-AS \cite{zhao2023self} that do not rely on additional labels, have demonstrated potential to achieve comparable performance  to supervised methods.
\end{enumerate}
\section{Challenges and Opportunities}
Through the comprehensive review of existing deep learning methods for 3D point cloud enhancement, we have witnessed their significant progress. Nonetheless, there are still several challenging problems in this field that deserve further research attention.
\begin{enumerate}
    \item \textbf{A unified framework for enhancement.} At present, there is a lack of unified frameworks for handling various kinds of nuisances simultaneously. However, raw point clouds are usually contaminated by a mixture of degrading factors such as noise, sparsity and occlusion. Thus, designing a unified framework for point cloud enhancement is urgent for real-world 3D point cloud processing,
    \item \textbf{Point cloud enhancement in extreme cases.} Currently methods generally consider common degrading cases. In some extreme cases, such as sensing with low-cost sensors and in high-dynamic scenes, the captured point clouds may be ultra noisy, sparse and incomplete. Dealing with such data is a challenging direction.
    \item \textbf{Large-scale real-world benchmark evaluation.} Existing quantitative evaluation benchmarks are typically synthetic ones, such as ShapeNet and ModelNet. In addition, the comparison on on real-world datasets tends to be only qualitative. Thus, it will be more convincing to construct large-scale real-world benchmarks for quantitative evaluation.
    \item \textbf{More attention on unsupervised \& weakly-supervised methods.} Data annotation for 3D point cloud enhancement is non-trivial, because the location and the number of so-called ground-truth points are generally confusing. As such, developing methods with less supervision signal is important this research realm. From our survey, we find that the research toward this direction is still at a very early stage. 
    \item \textbf{Down-stream task evaluation for assessing point cloud enhancement performance.} Point cloud enhancement usually acts as a pre-processing step for downstream tasks, such as point cloud registration, 3D object detection, and 3D semantic segmentation. Unfortunately, most current evaluation methods only focus on point cloud qualities, rather than the performance of downstream tasks. Hence, performing down-stream task evaluation for assessing point cloud enhancement  performance is necessary~\cite{lv2022intrinsic}.
    \item \textbf{AIG3D meets point cloud enhancement.} Large-scale pretraining have witnessed great success in artificial intelligence generated content (AIGC)~\cite{cao2023comprehensive} and artificial intelligence generated 3D (AIG3D)~\cite{raj2023dreambooth3d}. As point cloud enhancement can also be served as a point generation task, AIG3D is therefore closely related. Combining AIG3D with point cloud enhancement is an interesting direction in the era of large-scale neural models.
    \item \textbf{The leverage of 2D priors.} Multi-modal data are now increasingly popular in 3D vision applications. Images or videos are often accessible when capturing point clouds. As 2D images can also provide rich geometric priors, learning to resample point clouds with 2D priors could be a new solution.
    \item \textbf{Enhancement for 4D point clouds.} Most existing methods concentrate on processing a single frame point cloud. Nevertheless, 4D point clouds (point cloud sequences) are common data in autonomous driving, action recognition, and etc. Therefore, the leverage of temporal information for 4D point clouds enhancement remains an open issue.
\end{enumerate}\par

Point cloud enhancement has a wide range of applications in the fields of robotics, autonomous driving, augmented reality, and even embodied artificial intelligence. It is because the raw 3D point cloud generated by low-cost ranging sensor are low-quality, which hinders geometric and semantic understanding of the environment \cite{chen2024rppera, gu2023dense}. For example, in the background of autonomous driving, raw LiDAR point cloud is sparse so that objects far from LiDAR sensor are difficult to be detected. It could increase the risk of intelligent vehicle accident. In the applications of augmented reality, point cloud quality is main bottleneck of virtual-real synthesis and virtual scene rendering. In the context of robotics, a complete and high resolution 3D point cloud map greatly enlarges both the accuracy and success rate of robot localization and navigation. Thus, 3D point cloud enhancement is an essential technique in the community of 3D computer vision. Therefore, it is worthwhile to deal with above challenges of point cloud enhancement.


\section{Conclusions}
This paper provides a comprehensive overview of deep-learning-based methods for 3D point cloud enhancement, typically including point cloud denoising, completion, and upsampling. We have presented a comprehensive taxonomy and performance comparison of considered methods. The traits, merits, and demerits of these methods have been summarized. Finally, several potential future directions in this field have been discussed.

\footnotesize
\bibliographystyle{IEEEtran}
\bibliography{completion, denoise, upsample, into}
\end{document}